\documentclass[9.5pt,journal,compsoc]{IEEEtran}

\usepackage[pagebackref=true,hidelinks,breaklinks=true,bookmarks=false,linkcolor={black},citecolor={black},urlcolor={black}]{hyperref}
\usepackage{graphicx}
\usepackage{xcolor}
\usepackage{enumitem}
\usepackage{amsmath}
\usepackage{amssymb}
\usepackage{multirow}
\usepackage{amsthm}
\usepackage{mathrsfs}
\usepackage{booktabs}
\usepackage{bbding}
\usepackage{bbm}
\usepackage{subfigure}
\usepackage{ragged2e}
\usepackage{hyperref}
\usepackage{svg}
\svgpath{{fig/}} %
\usepackage{amsmath}
\usepackage{url}

\renewcommand{\paragraph}[1]{\vspace{2mm}\noindent \textbf{#1}}

\ifCLASSOPTIONcompsoc
  \usepackage[nocompress]{cite}
\else
  \usepackage{cite}
\fi

\begin{document}

\title{NOPE-SAC: Neural One-Plane RANSAC for Sparse-View Planar 3D Reconstruction}

\author{Bin~Tan,
        Nan~Xue,
        Tianfu~Wu,
        Gui-Song~Xia%
\IEEEcompsocitemizethanks{\IEEEcompsocthanksitem B. Tan, N. Xue, and G.-S. Xia are with the School of Computer Science, Wuhan University, Wuhan, China, 430072.
\IEEEcompsocthanksitem B. Tan and N. Xue are also with Ant Group, Hangzhou, China, 310013.
\IEEEcompsocthanksitem T. Wu is with the Department of Electrical and Computer Engineering, North Carolina State University, Raleigh, NC, USA, 27606.%
}
\thanks{(Corresponding author: Nan Xue.)}
}

\IEEEtitleabstractindextext{%
\justify
\begin{abstract}
This paper studies the challenging two-view 3D reconstruction problem in a rigorous sparse-view configuration, which is suffering from insufficient correspondences in the input image pairs for camera pose estimation. We present a novel {\em Neural One-PlanE RANSAC framework} (termed NOPE-SAC in short) that exerts excellent capability of neural networks to learn one-plane pose hypotheses from 3D plane correspondences. Building on the top of a Siamese network for plane detection, our NOPE-SAC first generates putative plane correspondences with a coarse initial pose. It then feeds the learned 3D plane correspondences into shared MLPs to estimate the one-plane camera pose hypotheses, which are subsequently reweighed in a RANSAC manner to obtain the final camera pose. Because the neural one-plane pose minimizes the number of plane correspondences for adaptive pose hypotheses generation, it enables stable pose voting and reliable pose refinement with a few of plane correspondences for the sparse-view inputs.
In the experiments, we demonstrate that our NOPE-SAC significantly improves the camera pose estimation for the two-view inputs with severe viewpoint changes, setting several new state-of-the-art performances on two challenging benchmarks, {\em i.e.}, MatterPort3D and ScanNet, for sparse-view 3D reconstruction. The source code is released at \textbf{\url{https://github.com/IceTTTb/NopeSAC}} for reproducible research.
\end{abstract}

}

\maketitle
\IEEEdisplaynontitleabstractindextext

\IEEEpeerreviewmaketitle

\IEEEraisesectionheading{\section{Introduction}\label{sec:introduction}}

\begin{figure}[!t]
\begin{center}
\subfigure[SuperGlue Keypoint Correspondences\cite{SuperGlue}]{
\includegraphics[width=0.87\linewidth]{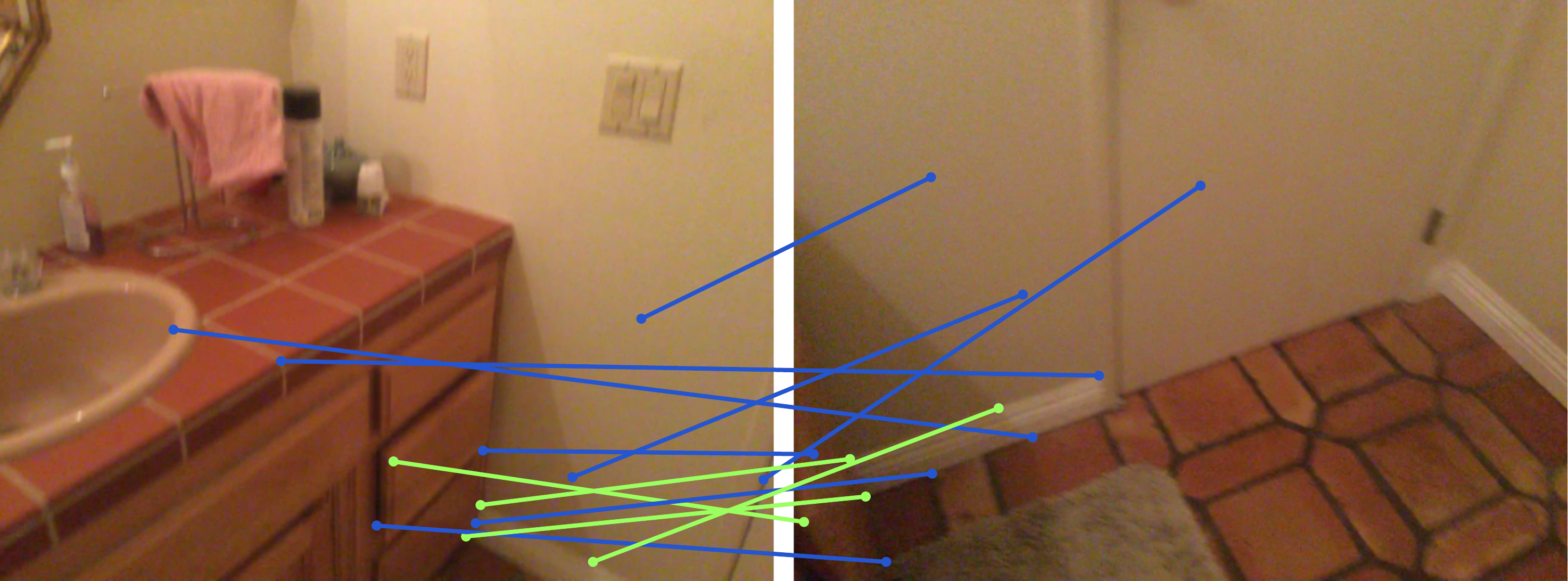}
\label{fig:teaser_a}}
\subfigure[Plane Correspondences]{
\includegraphics[width=0.87\linewidth]{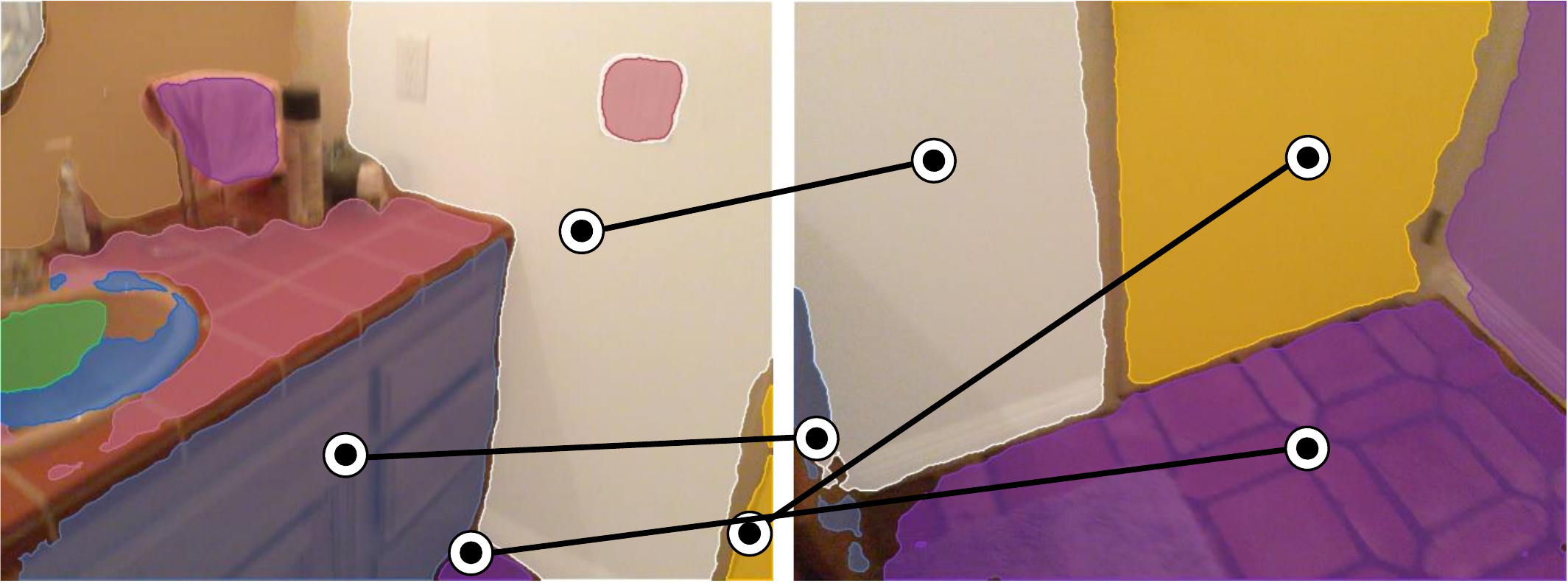}
\label{fig:teaser_b}}
\subfigure[Camera Poses and Planar Reconstruction]{
\includegraphics[width=0.87\linewidth]{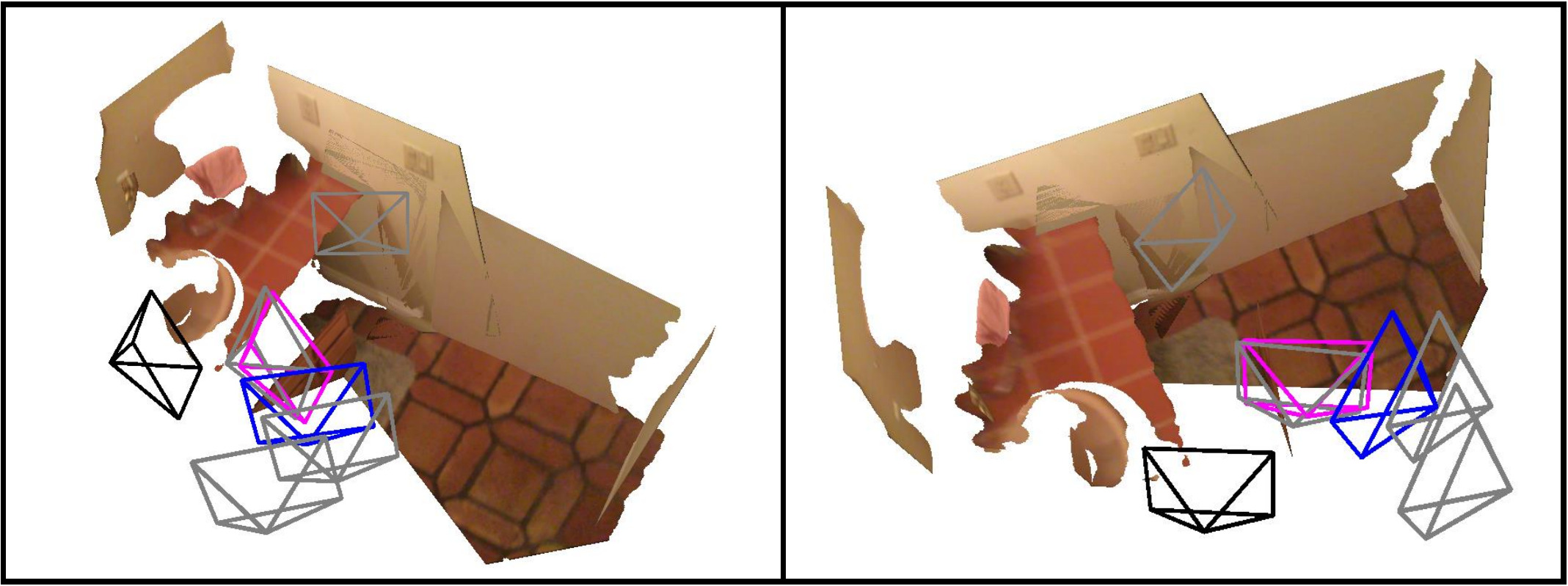}
\label{fig:teaser_c}}
\caption{An illustration of the indoor sparse-view 3D reconstruction. (a) shows the keypoint correspondences of SuperGlue~\cite{SuperGlue} including only 4 inliers (\textbf{Green} lines) and 8 outliers (\textbf{Blue} lines). (b) and (c) present the plane correspondences and reconstructed 3D planar scene of our NOPE-SAC from two sparse-view images. The \textbf{Black} frustum shows the ground truth camera of the second image. The \textbf{Pink}, \textbf{Gray}, and \textbf{Blue} frustums show the initial, the one-plane hypothetical, and the final cameras of the first image estimated by our NOPE-SAC.
}
\vspace{-6mm}
\label{fig:teaser}
\end{center}
\end{figure}

\IEEEPARstart{T}{wo-view} 3D reconstruction is a fundamental and longstanding problem in computer vision. It usually involves the recovery of the relative camera pose and scene geometry by establishing correspondences between features~\cite{MVG}. While this classic formulation has been successfully employed in conventional Structure-from-Motion systems using keypoints~\cite{CrandallOSH13,MonoSLAM,orbslam,COLMAP}, it faces challenges when dealing with severe viewpoint changes and low-texture appearance in indoor scenes, as illustrated in Fig.~\ref{fig:teaser_a}. In this paper, we aim to address this challenging configuration for two-view pose estimation and 3D scene reconstruction, which is commonly referred to as \emph{sparse-view} 3D reconstruction in the literature\cite{Ass3D,jin2021planar,planeformer}.

The \emph{sparse-view images} often come from large viewpoint changes during capturing, leading to a low overlap rate between these images, as shown in Fig.~\ref{fig:teaser_a} for instance. When employing the state-of-the-art SuperGlue~\cite{SuperGlue} for keypoint correspondences in the sparse-view image pair, the well-known 5-point algorithm~\cite{fivePointSolver} becomes infeasible to estimate the camera pose due to the presence of only $4$ inliers among the $12$ putative correspondences. However, upon careful examination of the input sparse-view images, particularly focusing on some planar elements such as the {\em wall}, {\em floor}, and {\em door}, we can intuitively perceive that these two images can be aligned in a unified 3D space. Pioneering works~\cite{jin2021planar,planeformer} have validated this observation in learning-based reconstruction systems by leveraging 3D plane correspondences.
They tackle the challenging relative pose estimation problem by voting among a set of pose hypotheses, according to the geometric correctness of plane correspondences.

The voting approaches presented in~\cite{jin2021planar,planeformer} share a similar concept with the typical keypoint-based RANSAC framework~\cite{RANSAC} for pose estimation. However, it differs in that they utilize precomputed pose hypotheses instead of computing them from inlier plane correspondences. To ensure that the precomputed pose hypotheses can handle complex pose distributions, the number of hypotheses would be large (\emph{{e.g.,} 1024 in SparsePlanes~\cite{jin2021planar}}). This leads to a contradiction between the abundance of pose hypotheses and the scarcity of plane correspondences. For instance, in Fig.~\ref{fig:teaser_b}, there are only 4 plane correspondences available for hypotheses voting. As a result, it becomes unstable and unreliable to distinguish the best hypothesis among the 1024 possibilities. Consequently, a time-consuming continuous optimization with the assistance of keypoint correspondences is required in~\cite{jin2021planar} for accurate sparse-view 3D reconstruction. Despite achieving impressive reconstruction results, they ignored a key problem of pose hypotheses generation from the perspective of RANSAC~\cite{RANSAC}, resulting in inflexibility and high computational cost for sparse-view 3D reconstruction.

In this paper, we pursue to formulate the camera pose estimation with pure plane correspondences in the RANSAC framework to get rid of the fixed pose hypotheses.
In the sparse-view setting, explicitly modeling the relationship between plane correspondences and the relative camera pose often leads to ill-posed minimal problems~\cite{PlaneOdometry}. To address this challenge, we leverage neural networks to encode the parameters of plane correspondences into a pose embedding space, enabling us to generate pose hypotheses through learned representations. 
We minimize the number of plane correspondences to 1 for the generation of each pose hypothesis, which we refer to as the {\em one-plane pose}. This formulation forms the basis of our proposed framework called {\em Neural One-PlanE RANSAC} (NOPE-SAC). By generating one-plane poses, we significantly reduce the number of pose hypotheses to be the same as the number of plane correspondences. 
As a result, the quantity contradiction between the pose hypotheses and their supported inlier plane correspondences is greatly alleviated. This allows us to fully exploit the information of plane correspondences and recover accurate camera poses from sparse-view images without relying on post-processing steps such as the continuous optimization described in~\cite{jin2021planar}.

We build our NOPE-SAC together with the plane detection and matching modules to establish a complete planar reconstruction system. Specifically, given two sparse-view images as input, our NOPE-SAC first detects 3D planes within each image and establishes plane correspondences by solving an optimal transport problem as in~\cite{SuperGlue}. To initiate the pose estimation process in RANSAC with an initial coarse camera pose, we utilize a convolutional neural network (ConvNet) to learn a pose embedding, which is then decoded into the initial pose. The parameters of each plane correspondence, along with the initial pose embedding, are combined to create new pose embeddings for generating one-plane pose hypotheses. To address the flexibility issue of pose initialization from ConvNets, we introduce an arbitrary pose initialization module (AIM), which follows an auto-encoder structure to encode poses obtained from other methods into pose embeddings. With a coarse initial camera pose and $N$ potential plane correspondences, our NOPE-SAC generates $N$ one-plane pose hypotheses. These hypotheses are then voted based on the geometric cost of the plane correspondences. Finally, the refined pose is estimated by fusing all hypotheses together according to their voting scores. The holistic planar reconstruction is achieved as the final output of our system. Figure~\ref{fig:teaser} provides an illustrative example of our NOPE-SAC with sparse-view input images. In this example, we consider four plane correspondences and an initial pose (pink frustum) predicted by the siamese network. By utilizing these correspondences, our NOPE-SAC generates four one-plane pose hypotheses (gray frustums). Subsequently, the plane correspondences are employed to vote and fuse all pose hypotheses (including the initial pose), enabling us to achieve a holistic 3D reconstruction of the scene.

In the experiments, we evaluate our NOPE-SAC on two indoor benchmark datasets, {\em i.e.}, Matterport3D~\cite{mp3d} with sparse-view splits created by SparsePlanes~\cite{jin2021planar} and ScanNet~\cite{dai2017scannet} with a more challenging split created by ourselves (see Sec.~\ref{subsec:exp_dataset} for details). On both benchmarks, our NOPE-SAC achieves state-of-the-art performance in terms of pose estimation accuracy and planar 3D reconstruction precision. Compared to the prior arts (\emph{e.g.,} PlaneFormers~\cite{planeformer}), our NOPE-SAC pushed the accuracy of pose estimation on the Matterport3D dataset to $73.2\%$ and $89.0\%$ for translation and rotation ($6.4\%$ and $5.2\%$ absolute improvements), the accuracy of pose estimation on the ScanNet dataset to $82.0\%$ and $82.6\%$ for translation and rotation ($6.7\%$ and $9.4\%$ absolute improvements), and the Average Precision (AP) of 3D plane reconstruction to $43.29\%$ on the Matterport3D and $39.39\%$ on the ScanNet ($5.76\%$ and $4.75\%$ absolute improvements). The comprehensive ablation studies further verified the design rationales of the proposed NOPE-SAC.

In summary, we present a novel approach, {\em i.e.}, NOPE-SAC, to address the challenging problem of sparse-view planar 3D reconstruction in a RANSAC framework, which fully takes the advantage of end-to-end deep neural networks. Benefiting from the accurate camera pose estimated from our one-plane pose hypotheses, we achieve superior 3D reconstruction results without incurring any offline optimization procedures. Our method sets several new state-of-the-art results on both the Matterport3D~\cite{mp3d} and the ScanNet~\cite{dai2017scannet} datasets for both pose estimation and holistic planar reconstruction.

\section{Related Work}
\subsection{Single-View 3D Reconstruction}

One relevant task to the sparse-view 3D reconstruction is to recover 3D scenes from single images. As one of the most widely used solution, pixel-wise depth estimation from single views has been extensively studied~\cite{MonocularDepthV2, robustDepth, ImproveMDE,VND,MegaDepth}.
Benefiting from the advances of deep learning and the richness of the training data~\cite{omnidata,midas}, we have witnessed their significant improvements in depth estimation accuracy and generalization ability.

However, single-view depth estimation alone has limitations in generating structured 3D point clouds. It can introduce structural distortions when applied to scenes with well-defined structures, such as indoor scenes. To overcome this problem, some researchers~\cite{planeAE,planercnn,planenet,planeTR} have proposed to predict structured 3D planes from a single image directly. For example, Liu~\MakeLowercase{\textit{et~al.}}~\cite{planercnn} applied a two-stage instance segmentation framework to jointly detect plane instance masks and estimate 3D plane parameters for the single view planar reconstruction. Although these insightful methods work well for single-view indoor plane reconstruction, they can not recover the holistic indoor scene because of the limited field of view in every single-view image. 
In this paper, we build upon the advantages of single-view planar 3D reconstruction and go further to the challenging sparse two-view configurations. We demonstrate that the estimated 3D planes from single-view images are favorable to both camera pose estimation and holistic planar 3D reconstruction when dealing with sparse two-view images.

\subsection{Two-View Camera Pose Estimation}
Two-view 3D reconstruction is the most fundamental task in computer vision, which is formulated to solve the relative camera poses between the input images and estimate the scene geometry from the camera motions. The problem of camera pose estimation is the core of this task. 

A common solution is to estimate camera poses from keypoint correspondences~\cite{SIFT,ORB,SURF,BRISK,ASIFT} relying on a typical 5-point solver~\cite{fivePointSolver} within a RANSAC~\cite{RANSAC} framework. 
Following this paradigm, there have been tremendous efforts to improve the performance of keypoint detection and description~\cite{superpoint, R2D2,ContextDesc,PixelPerfect} and \textcolor{black}{matching by neural networks~\cite{SuperGlue,COTR,LoFTR,matchformer, ASpanFormer}.} As our study mainly focuses on the sparse-view configuration for two-view 3D reconstruction, such a common solution becomes infeasible due to the low overlap rate between the sparse-view inputs, which hampers the establishment of reliable feature correspondences.

Recently, some approaches have leveraged neural networks to directly estimate camera poses from feature cost volumes built up on dense pixel correspondences~\cite{deepsfm,posenet,MonoIndoor,RPNet}. Besides, Wei~\MakeLowercase{\textit{et~al.}}~\cite{deepsfm} combined the traditional and learning-based methods by estimating dense pixel correspondences with a neural network and solving camera poses with the 5-point solver~\cite{fivePointSolver}. While these methods have exhibited impressive performances, they heavily rely on sufficient image overlap to extract motion cues from correspondences. Consequently, they face challenges when confronted with sparse-view indoor images that have limited overlap rates.

Most recently, there has been significant research dedicated to addressing the problem of pose estimation from sparse-view indoor images~\cite{Ass3D,jin2021planar,planeformer}. Due to the challenge of obtaining accurate camera poses, these approaches utilize a large number of precomputed pose templates clustered from the ground-truth camera poses. The pose estimation problem is then formulated as a label voting task, where classification likelihoods and plane correspondences are used to determine the scores of pose templates.
However, scoring a large number of pose templates with a few plane correspondences can potentially impact estimation accuracy. As a result, these approaches have to apply an extra optimization scheme for pose refinement. 
In contrast, we show that our NOPE-SAC can work well in such a voting pipeline to accurately estimate camera poses, thanks to the generation and fusion of our novel one-plane pose hypotheses in end-to-end learning.

\begin{figure*}
\centering
\includegraphics[width=0.95\linewidth]{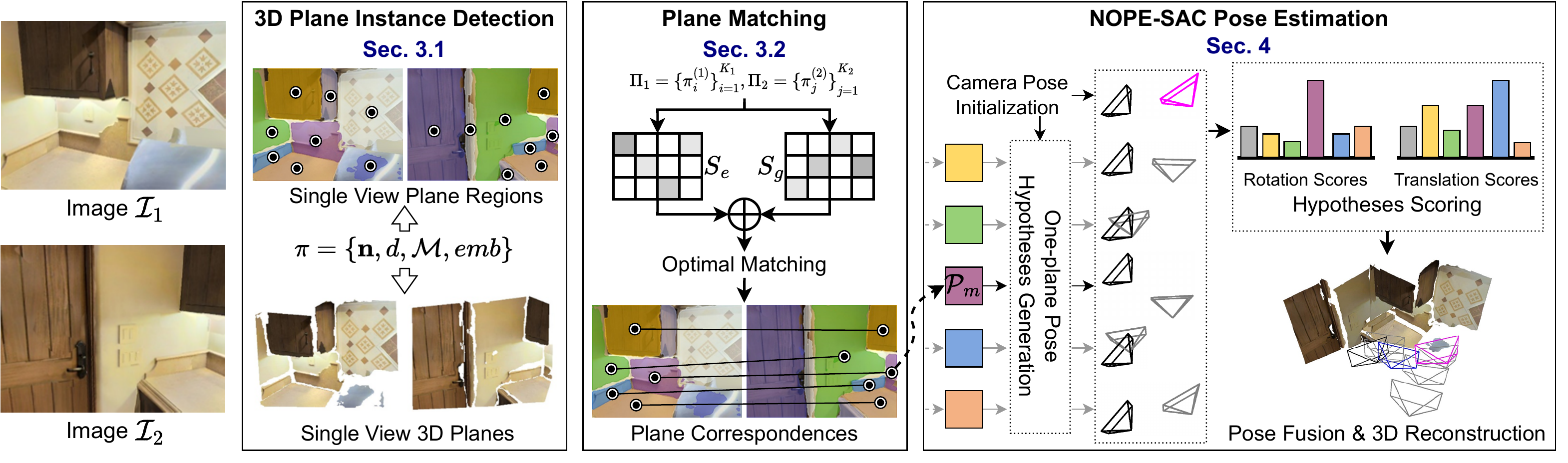}
\vspace{-10pt}
\caption{Overview of the proposed NOPE-SAC. Our network first detects 3D planes (Sec.~\ref{sec:plane_detection}) and estimates plane correspondences (Sec.~\ref{sec:plane_matching}) from the input sparse-view images. Then, the final relative camera pose of the first image (\textbf{Blue} frustum) is voted from a coarse initial pose (\textbf{Pink} frustum) and a few one-plane poses (\textbf{Gray} frustum) according to the geometric cost of plane correspondences (Sec.~\ref{sec:NopeSAC-pose}). At last, the planar 3D reconstruction is achieved as the final output. The \textbf{Black} frustum shows the camera of the second image. 
}
\label{fig:network}
\end{figure*}

\section{Planar Correspondences Preparation}
The overview of our proposed NOPE-SAC for planar 3D reconstruction is illustrated in Fig.~\ref{fig:network}. It detects single-view 3D planes from the input sparse-view images and establishes the plane correspondences between views for camera pose estimation and planar 3D reconstruction.
We focus on 3D plane instance detection and matching in this section and leave the key components for pose estimation in Sec.~\ref{sec:NopeSAC-pose}.
Using the estimated relative camera pose, we align the 3D plane correspondences between viewpoints to achieve the final scene reconstruction.

\subsection{3D Plane Instance Detection}
\label{sec:plane_detection}
Similar to~\cite{jin2021planar}, we define a plane instance as $\pi = \left\{\mathbf{n}, d, \mathcal{M}, \textcolor{black}{\textit{emb}} \right\}$, where $\mathbf{n} \in \mathbb{R}^3$ and \textcolor{black}{$d \in \mathbb{R^{+}}$} are the plane normal and the offset from the plane to the camera center, $\mathcal{M} \in \mathbb{R}^{H \times W}$ is the plane segmentation mask ($H$ and $W$ are the image height and width) and $\textit{emb} \in \mathbb{R}^{256}$ is a plane appearance embedding. 

We use PlaneTR~\cite{planeTR} as the 3D plane detection module for each input view with two main modifications: 
(1) the line segment branch in PlaneTR~\cite{planeTR} is excluded to keep the simplicity; (2) the backbone network of PlaneTR~\cite{planeTR} is replaced to ResNet-50~\cite{resnet} for efficiency.

During our computation, we leverage the output feature of the Transformer decoder from~\cite{planeTR} as our plane appearance embedding. Atop this embedding, we predict the normal and offset for each plane using a linear layer. In addition, we decode the plane mask by concurrently utilizing the backbone features and the plane appearance embedding, a strategy akin to the one presented in~\cite{cheng2021maskformer}.

Our 3D plane detection module employs the following loss function:
\begin{equation}
\mathcal{L}_{\text{plane}} = \mathcal{L}_{\text{cls}} + 20\mathcal{L}_{\text{mask}} + \mathcal{L}_{\text{dice}} + \mathcal{L}_{\text{parm}} + 0.5\mathcal{L}_{\text{center}},
\end{equation}
where $\mathcal{L}_{cls}$ represents the classification loss,
$\mathcal{L}_{mask}$ denotes the mask loss,
$\mathcal{L}_{dice}$ signifies the dice loss as described in~\cite{dice},
$\mathcal{L}_{parm}$ stands for the plane parameter loss, and
$\mathcal{L}_{center}$ indicates the plane center loss.
These loss functions have been previously utilized in studies such as~\cite{planeTR, cheng2021maskformer}. For a comprehensive understanding, readers can refer to~\cite{planeTR, cheng2021maskformer} for details.

\subsection{Plane Matching}
\label{sec:plane_matching}
Denoted by $\Pi_1 ={ \{ \pi_i^{(1)} \} }_{i=1}^{K_1}$ and $\Pi_2 ={ \{ \pi_j^{(2)} \} }_{j=1}^{K_2}$ the plane sets from two view images, we  find an optimal partial assignment $A \in [0, 1]^{K_1 \times K_2}$ from $\Pi_1$ to $\Pi_2$ for plane matching, where 
$K_1$ and $K_2$ are the number of plane instances in two images. The optimal partial assignment $A$ is solved as an Optimal Transport (OT) problem based on the scoring matrix $S \in \mathbb{R}^{K_1 \times K_2}$.

\paragraph{Scoring Matrix.}
Given the plane sets $\Pi_1$ and $\Pi_2$, we compute two distinct affinity matrices: one based on plane appearance and another on geometry. These matrices are then linearly added to create the final scoring matrix.
To compute the affinity of plane appearance, we first apply an Attentional Graph Neural Network (AGNN) proposed in~\cite{SuperGlue} to obtain two encoded sequences ${E_{1}} \in \mathbb{R}^{K_1 \times 256}$ and ${E_{2}} \in \mathbb{R}^{K_2 \times 256}$ by:
\begin{equation}
    \textcolor{black}{ \{E_{1}, E_{2} \} = \text{AGNN}(\{ \textit{emb}_i^{(1)} \}_{i=1}^{K_1}, \{ \textit{emb}_i^{(2)} \}_{i=1}^{K_1}),}
\end{equation}
where $\{ \textit{emb}_i^{(1)} \}_{i=1}^{K_1}$ and $\{ \textit{emb}_i^{(2)} \}_{i=1}^{K_1}$ are the appearance embeddings of planes in two images. 
In the AGNN, both the \textit{self-edge} and \textit{cross-edge} layers are set to a count of 9, consistent with the configuration in SuperGlue~\cite{SuperGlue}. We chose to omit the positional encoding scheme from~\cite{SuperGlue} when dealing with planes. This is due to the view-dependent definition of the center for 3D planes, which renders it less suitable in our specific context.
Then, the appearance affinity matrix is computed by:

\begin{equation}
    S_e={E_{1}} {E_{2}}^T.
\end{equation}

The affinity based on geometry is characterized by the parameters between two 3D planes. Given an initial relative camera pose, represented by \( R \) and \( \mathbf{t} \) (for a detailed explanation, refer to Sec.~\ref{subsec:pose_initial}), we first transition the 3D plane parameters of \( \Pi_1 \) to \( {\widetilde{\Pi}}_1 \). This represents the transformed version of \( \Pi_1 \), adjusted to the coordinate frame of \( \Pi_2 \). Subsequently, we derive the geometric affinity matrix as:
\begin{equation}
\label{eq:geo_score}
S_g(i,j)= -\lambda_1 \text{acos}(\tilde{\mathbf{n}}_i^{(1)}, {\mathbf{n}}_j^{(2)})-\lambda_2 |\tilde{d}_i^{(1)}-d_j^{(2)}|,
\end{equation}
where $\tilde{\mathbf{n}}_i^{(1)}$ and $\tilde{d}_i^{(1)}$ are the $i^{th}$ plane normal and offset in ${\widetilde{\Pi}}_1$. $\mathbf{n}_j^{(2)}$ and $d_j^{(2)}$ are the $j^{th}$ plane parameters in $\Pi_2$. $\lambda_1$ and $\lambda_2$ are set to 0.125 and 0.25 to balance the magnitude of two terms. 

Finally, the scoring matrix $S$ can be calculated as:
\begin{equation}
\label{eq:full_score}
S=S_e + S_g
\end{equation}

\paragraph{Optimal Matching.}
Given a scoring matrix $S$, we apply a differentiable Sinkhorn Algorithm~\cite{sinkhorn1967concerning,computeOptTrans} \textcolor{black}{implemented by~\cite{SuperGlue}} to achieve the soft assignment matrix $A$. For a pair of planes $\{ \pi_i^{(1)}, \pi_j^{(2)} \}$, they can be regarded as matched planes if $A(i,j)$ is the maximal score both in the $i^{th}$ row and the $j^{th}$ column of $A$, and $A(i,j)$ is larger than a fixed matching threshold (0.2 in this paper).
We supervise the plane matching module with the loss defined in Eqn.~\eqref{eq:ot-loss}. More precisely, let $\bar{A} \in \mathbb{R}^{(K_1+1) \times (K_2+1)}$ to be the soft assignment matrix augmented with dustbins and the expected soft assignment matrix $A=\bar{A}_{1:K_1,1:K_2}$, the plane matching loss can be calculated as:
\begin{equation}\label{eq:ot-loss}
\small 
    \mathcal{L}_{\text{match}} = -\sum_{(i,j)\in \mathcal{A}} \log\bar{A}_{i,j} -\sum_{i \in \mathcal{I}}\text{log}\bar{A}_{i, K_2+1}
    - \sum_{j \in \mathcal{J}}\text{log}\bar{A}_{K_1+1,j} ,
\end{equation}
where $\mathcal{A}=\{(i,j)\}$ are the indices of ground truth matches, $\mathcal{I}$ and $\mathcal{J}$ are the indices of unmatched planes in two images.

\paragraph{Correspondences Preparation for Pose Estimation.}
Following the computation of optimal matching, we extract a set of $M$ plane correspondences, denoted by $\mathbb{P} = \{\mathcal{P}_m = (\mathcal{P}_m^{(1)},\mathcal{P}_m^{(2)})\}_{m=1}^M$. Here, the superscript $^{(i)}$ signifies the viewpoint index. For every matched plane, $\mathcal{P}_m^{(i)}$, we define its normal as $\mathbf{n}_{m}^{(i)}\in\mathbb{R}^3$ and its offset as $d_m^{(i)}\in\mathbb{R}^+$. With the plane correspondences encapsulated in $\mathbb{P}$, our objective is to estimate the relative camera pose, as discussed in Sec.~\ref{sec:NopeSAC-pose}.

\section{NOPE-SAC Pose Estimation}\label{sec:NopeSAC-pose}
As depicted in Fig.~\ref{fig:network}, we begin with an initial camera pose, represented by $\xi_0 = (R_{0}, \mathbf{t}_0)$. For a given pair of matched planes $\mathcal{P}_m = (\mathcal{P}_m^{(1)}, \mathcal{P}_m^{(2)})$ from images $\mathcal{I}_1$ and $\mathcal{I}_2$, our proposed NOPE-SAC is designed to learn the one-plane pose $\xi_m = (R_m, \mathbf{t}_m)$, which exists within the embedding space defined by $\xi_0$ and $\mathcal{P}_m$. For all $M$ potential plane matches, this process yields $M$ distinct one-plane poses. These individual poses are then assessed based on the geometric cost in the planar consensus set. Subsequently, we integrate these poses to derive a refined camera pose.

\subsection{Camera Pose Initialization} \label{subsec:pose_initial}
\begin{figure}[t]
\begin{center}
\includegraphics[width=0.9\linewidth]{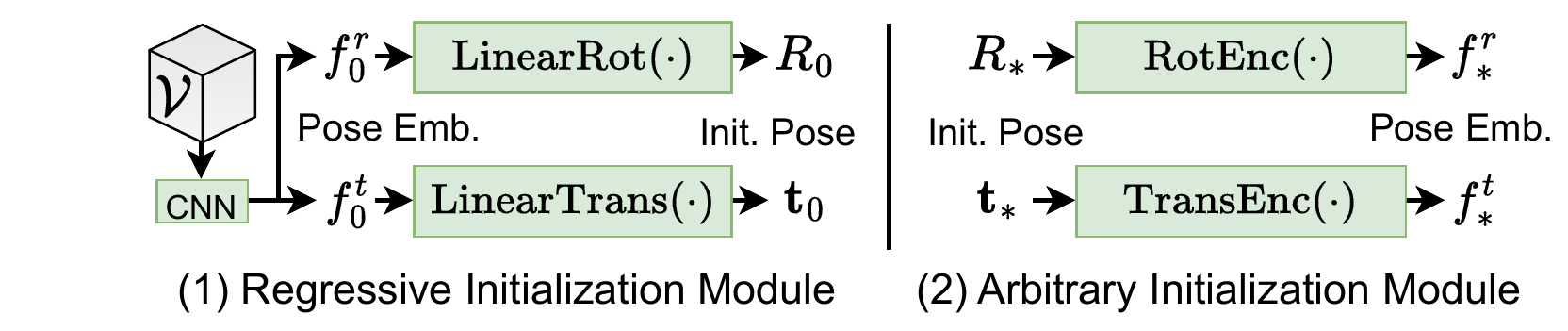}
\end{center}
\vspace{-0.5cm}
\caption{An architecture of camera pose initialization.}
\label{fig:pose_init}
\end{figure}

In consensus sampling pipelines for camera pose estimation, an initial coarse estimation is required. While the keypoint-based RANSAC and its variants typically achieve this by randomly sampling a subset of keypoint correspondences, deriving a coarse estimation from plane correspondences using an explicit mathematical model presents a greater challenge. As a result, we leverage a convolutional neural network to learn the initial pose $\xi_0$ from the feature volume of images $\mathcal{I}_1$ and $\mathcal{I}_2$. This is facilitated by our Regressive Initialization Module (RIM) depicted in Fig.~\ref{fig:pose_init} (left).

\paragraph{\textcolor{black}{Regressive Initialization Module (RIM).}} We take ResNet-50~\cite{resnet} as the backbone network for the initial pose regression. Given two input images $\mathcal{I}_1$ and $\mathcal{I}_2$ with a spatial size of $H\times W$, we take their output backbone features $\mathcal{F}_1$ and $\mathcal{F}_2$ to build a 4D affinity volume $\mathcal{V} \in \mathbb{R}^{(h\times w)\times (h\times w)}$ by: 
\begin{equation}
    \textcolor{black}{\mathcal{V}(\mathbf{p},\mathbf{p}') = \frac{\exp(\mathcal{F}_2^T(\mathbf{p}')\mathcal{F}_1(\mathbf{p}))}{\sum_{\mathbf{p}'} \exp(\mathcal{F}_2^T(\mathbf{p}')\mathcal{F}_1(\mathbf{p}))},}
\end{equation}
for any pair of pixels $\mathbf{p}$ and $\mathbf{p}'$ in the coordinates of $\mathcal{F}_1$ and $\mathcal{F}_2$, where $h = H/s$, $w = W/s$, and $s=32$ is the stride of backbone network. The computation of $\mathcal{V}$ follows a similar approach to~\cite{jin2021planar}. However, we opt to directly regress the camera pose rather than learning the classification likelihood based on a set of discrete pose templates.

To predict the initial camera pose $\xi_0 = (R_0, \mathbf{t}_0) \in \text{SE}(3)$, the affinity volume is first rearranged to $\mathcal{V} \in \mathbb{R}^{hw\times h \times w}$. Subsequently, it is transformed using 6 convolutional layers (including BatchNorm and LeakyReLU) followed by a linear layer. 
This yields two pose embeddings: $f_{0}^{r}$ and $f_{0}^{t} \in \mathbb{R}^{256}$, corresponding to the rotation $R_0$ and translation $\mathbf{t}_0$. Notably, the stride for every even-numbered convolution operation is set to 2 to downsample the feature size.
With $f_{0}^{r}$ and $f_{0}^{t}$, it is straightforward to predict the rotation and translation by two linear layers $\text{LinearRot}(\cdot)$ and $\text{LinearTrans}(\cdot)$ as:
\begin{equation}\label{eq:linear-pose-decoding}
    \textcolor{black}{{R}_{0} = \text{LinearRot}(f_0^{r}), \mathbf{t}_0 = \text{LinearTrans}(f_0^{t}),}
\end{equation}
where rotation ${R}_0 \in \mathbb{R}^4$ is represented as the unit quaternion that satisfies $\left\|{R}_0\right\|_2 = 1$.
The weights of $\text{LinearRot}(\cdot)$ and $\text{LinearTrans}(\cdot)$ are optimized by the MSE (Mean-Square Error) loss as:
\begin{equation}\label{eq:cam-loss}
    \mathcal{L}_{\text{cam}}^{\text{init}} = \|{R}_0-{R}^{gt}\|_2^2 + \|\mathbf{t}_0-\mathbf{t}^{gt}\|_2^2,
\end{equation}
where ${R}^{gt}$ and $\mathbf{t}^{gt}$ are the ground truth of the rotation (in unit quaternion) and translation between the input images.

Besides predicting the initial camera poses, the pose embeddings \(f_0^{r}\) and \(f_0^{t}\) are instrumental in generating one-plane pose hypotheses, as detailed in Sec.~\ref{subsec:pose_refine} within our NOPE-SAC framework. A challenge emerges when using initial poses that do not originate from our Regressive Initialization Module; this can compromise the flexibility in generating the desired \(f_0^{r}\) and \(f_0^{t}\) embeddings. To address this, we introduce an Arbitrary Initialization Module (AIM). This module reconstructs any given camera pose through auto-encoding, as depicted in Fig.~\ref{fig:pose_init}~(right).

\paragraph{Arbitrary Initialization Module (AIM).}
\textcolor{black}{We featurize any given rotation $R_{*}\in\mathbb{R}^4$ and translation $\mathbf{t}_{*}\in\mathbb{R}^3$ in an auto-encoding manner by:
\begin{equation}
\label{eq:feat_initial}
\begin{split}
    f_*^r = \text{RotEnc}({R}_*),\, f_*^t = \text{TransEnc}(\mathbf{t}_*), \\
    \tilde{{R}}_* = \text{LinearRot}(f_*^r),\, \tilde{\mathbf{t}}_* = \text{LinearTrans}(f_*^t),
\end{split}
\end{equation}
where $\text{RotEnc}(\cdot)$ and $\text{TransEnc}(\cdot)$ are the encoders for rotation and translation implemented by MLPs, $\text{LinearRot}(\cdot)$ and $\text{LinearTrans}(\cdot)$ are the shared layers defined in Eqn.~\eqref{eq:linear-pose-decoding}. The camera pose is reconstructed by minimizing the following loss function:
\begin{equation}
    \mathcal{L}_{\text{rec}}= \left\| \tilde{{R}}_{*} - {R}_{*} \right\|_2^2 + \left\| \tilde{\mathbf{t}}_{*} - \mathbf{t}_{*} \right\|_2^2.
\end{equation}
For the training of AIM, we randomly sample the rotations and translations from the uniform distribution. Please move to Sec.~\ref{sec:implementation} for the implementation details.
}

\begin{table}[t!]
\caption{Detailed architecture of MLPs used in NOPE-SAC Pose Estimation Module. For each MLP, the corresponding equation, notation, the number of linear layers, the channel dimensions of inputs and outputs, and the activation functions are listed.}
\vspace{-2mm}
\centering
\resizebox{0.8 \linewidth}{!}{ 
    \begin{tabular}{c|c|c|cc|c}
    \toprule
        \multicolumn{1}{c|}{\multirow{2}{*}{Equation}} & \multicolumn{1}{c|}{\multirow{2}{*}{Notation}} & \multicolumn{1}{c|}{\multirow{2}{*}{$\#$~Layers}} &\multicolumn{2}{c|}{$\#$~Channels} & \multicolumn{1}{c}{\multirow{2}{*}{Activation}} \\ 
               & &        & In   & Out                   &            \\\midrule
        \text{Eqn.~\eqref{eq:feat_initial}} & \text{RotEnc / TransEnc} & 6 & 256 & 256 & ReLU \\\midrule
        \multicolumn{1}{c|}{\multirow{5}{*}{\text{Eqn.~\eqref{eq:corrs_enc}}}} & \multicolumn{1}{c|}{\multirow{5}{*}{$\mathcal{G}$}} & 1 & 8 & 1024 & ReLU \\
        & & 4 & 1024 & 1024 & ReLU \\
        & & 1 & 1024 & 1024 & -    \\
        & & 2 & 1024 & 1024 & ReLU \\
        & & 1 & 1024 & 1024 & -     \\\midrule
        \multicolumn{1}{c|}{\multirow{3}{*}{\text{Eqn.~\eqref{eq:corrsEmb2poseFeat}}}} & \multicolumn{1}{c|}{\multirow{3}{*}{$\mathcal{E}_r$}} & 1 & 1024 & 512 & ReLU \\
        & & 4 & 512 & 512 & ReLU    \\
        & & 1 & 512 & 256 & -    \\ \midrule
        \multicolumn{1}{c|}{\multirow{6}{*}{\text{Eqn.~\eqref{eq:corrsEmb2poseFeat}}}} & \multicolumn{1}{c|}{\multirow{6}{*}{$\mathcal{E}_t$}} & 1 & 1280 & 1024 & ReLU \\
        & & 1 & 1024 & 1024 & ReLU \\
        & & 1 & 1024 & 1024 & - \\
        & & 1 & 1024 & 512  & ReLU \\
        & & 4 & 512 & 512   & ReLU \\
        & & 1 & 512 & 256   & - \\ \midrule
        \multicolumn{1}{c|}{\multirow{2}{*}{\text{Eqn.~\eqref{eq:one_plane_emb}}}} & \multicolumn{1}{c|}{\multirow{2}{*}{$\mathcal{D}_r$ / $\mathcal{D}_t$}} & 2 & 512 & 512 & ReLU \\
        & & 1 & 512 & 256  & ReLU \\
        \bottomrule
    \end{tabular}
}
\label{tab:MLPs}
\end{table}

\subsection{Camera Pose Refinement}
\label{subsec:pose_refine}
Given the initial camera pose $\xi_0 = (R_0, \mathbf{t}_0)$ and its corresponding pose embeddings \(f_0^{r}\) and \(f_0^{t}\) (obtained either from RIM or AIM), as well as \(M\) 3D plane correspondences 
$\mathbb{P} = \{\mathcal{P}_m=(\mathcal{P}_m^{(1)}, \mathcal{P}_m^{(2)})\}_{m=1}^M$, we introduce the core of NOPE-SAC, which encompasses three main steps: (1) One-plane Pose Hypotheses Generation, (2) Hypotheses Scoring, and (3) Final Pose Estimation.

\begin{figure}[t]
\begin{center}
\includegraphics[width=0.95\linewidth]{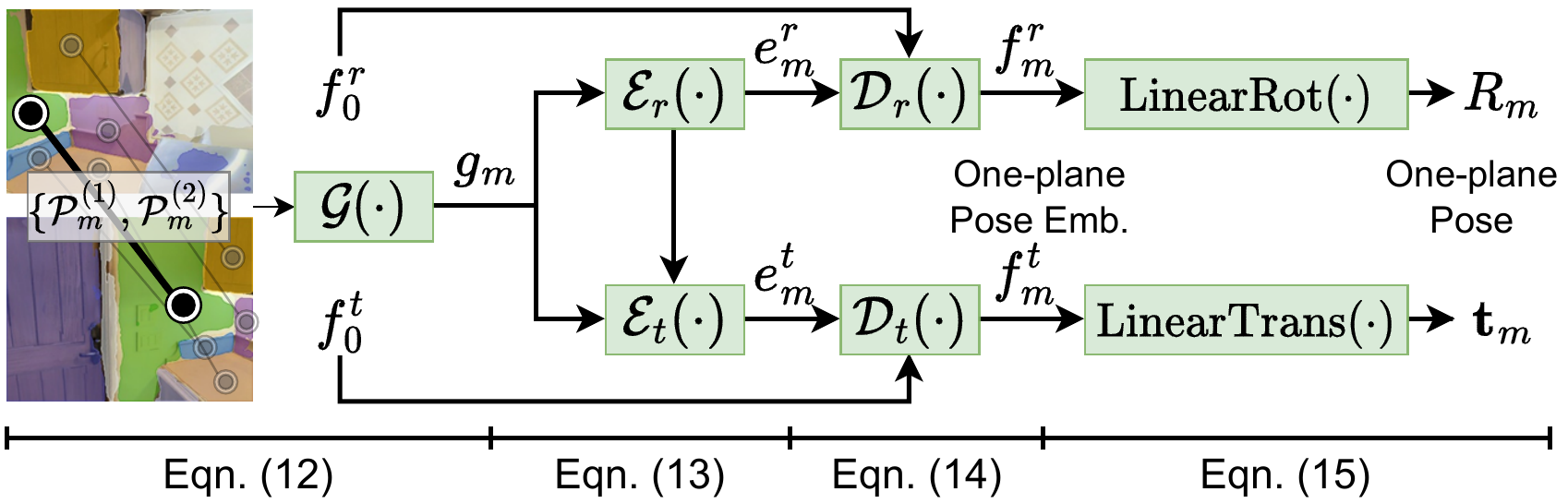}
\end{center}
\vspace{-0.5cm}
\caption{An architecture of the one-plane pose hypotheses generation.}
\label{fig:one_plane_pose}
\end{figure}

\paragraph{One-plane Pose Hypotheses Generation.} As shown in Fig.~\ref{fig:one_plane_pose}, for the $m$-th 3D plane correspondence $\mathcal{P}_m^{(1)} = (\mathbf{n}_m^{(1)}, {d}_{m}^{(1)})$ in $\mathcal{I}_1$ and $\mathcal{P}_m^{(2)} = (\mathbf{n}_m^{(2)}, {d}_{m}^{(2)})$ in $\mathcal{I}_2$, we leverage an MLP layer $\mathcal{G}(\cdot): \mathbb{R}^8 \mapsto \mathbb{R}^{1024}$ to embed the correspondence by:
\begin{equation}\label{eq:corrs_enc}
    g_m = \mathcal{G}(\mathbf{n}_{m,0}^{(1)} \oplus {d}_{m,0}^{(1)} \oplus \mathbf{n}_m^{(2)} \oplus d_m^{(2)}),
\end{equation}
where $\mathbf{n}_{m,0}^{(1)}$ and $d_{m,0}^{(1)}$ are the warped normal of $\mathbf{n}_m^{(1)}$ and offset of $d_m^{(1)}$ by the relative camera pose $(R_0,\mathbf{t}_0)$, $\oplus$ is the concatenation operation. Here, the warped plane parameters by the initial camera pose could be regarded as a kind of normalization to facilitate the learning of plane correspondence embedding in neural networks.

After achieving the correspondence feature $g_m$, we further transform it to the rotation feature $e_{m}^r \in \mathbb{R}^{256}$ and the translation feature $e_m^t\in\mathbb{R}^{256}$ by two MLPs $\mathcal{E}_r (\cdot): \mathbb{R}^{1024} \mapsto \mathbb{R}^{256}$ and $\mathcal{E}_t (\cdot): \mathbb{R}^{1280} \mapsto \mathbb{R}^{256}$ as:
\begin{equation}
\label{eq:corrsEmb2poseFeat}
\begin{split}
e_m^r &= \mathcal{E}_r(g_m), \\
e_m^t &= \mathcal{E}_t(g_m \oplus e_m^r).
\end{split}
\end{equation}

Then, we concatenate $e_m^r$ and $e_m^t$ with the embeddings $f_0^{r}$ and $f_0^{t}$ of the initial camera pose and transform the concatenated features by another two MLP layers $\mathcal{D}_r(\cdot): \mathbb{R}^{512} \mapsto \mathbb{R}^{256}$ and $\mathcal{D}_t(\cdot): \mathbb{R}^{512} \mapsto \mathbb{R}^{256}$ to yield the one-plane pose embeddings $f_m^r$ and $f_m^t$ as:
\begin{equation} \label{eq:one_plane_emb}
\begin{split}
f_m^r &= \mathcal{D}_r(f_0^{r} \oplus e_m^r), \\
f_m^t &= \mathcal{D}_t(f_0^{t} \oplus e_m^t).
\end{split}
\end{equation}

Finally, to obtain the one-plane pose $\xi_m=(R_m, \textbf{t}_m)$ for the $m$-th plane correspondence, we leverage the linear layers $\text{LinearRot}(\cdot)$ and $\text{LinearTrans}(\cdot)$ defined in Eqn.~\eqref{eq:linear-pose-decoding} as the shared headnets by:
\begin{equation}
\begin{split}
    {R}_{m} &= \text{LinearRot}(f_m^r), \\
    \mathbf{t}_{m} &= \text{LinearTrans}(f_m^t),
\end{split}
\end{equation}
where ${R}_{m}$ and $\mathbf{t}_{m}$ are the predicted quaternion of rotation and the translation.

\paragraph{Hypotheses Scoring.} Next to the pose hypotheses generation, \textcolor{black}{we score the predicted one-plane rotations and translations by computing their geometric costs. Specifically, for the $m\text{-th}$ pose hypothesis $\xi_m=(R_{m}, \mathbf{t}_{m})$, its rotation cost vector $c_m^{r} \in \mathbb{R}^M$ and translation cost vector $c_m^{t} \in \mathbb{R}^M$ throughout all $M$ plane correspondences are calculated by:}
\begin{equation}\label{eq:cost_pose}
    \begin{split}
        c_m^{r} &= \left(\| \mathbf{n}_{i,m}^{(1)} - \mathbf{n}_i^{(2)}\|_2\right)_{i=1}^M\\
        c_m^{t} &= \left(\| d_{i,m}^{(1)} \mathbf{n}_{i,m}^{(1)} - d_i^{(2)}\mathbf{n}_i^{(2)}\|_2\right)_{i=1}^M,
    \end{split}
\end{equation}
where $\mathbf{n}_{i,m}^{(1)}$ and $d_{i,m}^{(1)}$ are the warped normal and offset of the plane $\mathcal{P}_i^{(1)}$ by the $m$-th pose hypothesis $\xi_m$. 

After computing and concatenating the cost vectors of all $M+1$ pose hypotheses (one initial pose and $M$ one-plane poses), there will be two cost matrices $C_r\in\mathbb{R}^{(M+1)\times M}$ and $C_t\in\mathbb{R}^{(M+1)\times M}$. These two cost matrices are then fed into two parallel Hypotheses Scoring Layers to yield the rotation score vector $\mathbf{w}_r = (w_0^{r},\ldots,w_M^{r})$ and the translation score vector $\mathbf{w}_t = (w_0^{t},\ldots,w_M^{t})$, respectively. Here, the sum of each score vector is equal to $1$, and we use the Softmax operation in our implementation. In detail, each Hypotheses Scoring Layer consists of an MLP layer (including three linear layers with ReLU activations and 64 channels per layer) and one linear layer for score prediction.

To train the scoring layer, we design a scoring loss to dynamically supervise the learning of pose scores. Let $i$ and $j$ be the indices of rotation and translation in all pose hypotheses $\{ R_{m} \}_{m=0}^M$ and $\{ \mathbf{t}_{m} \}_{m=0}^M$ that are closest to the ground truth camera poses. Then, the heuristic scoring loss can be calculated as:
\begin{align}\label{eq:score-loss}
    \mathcal{L}_{\text{score}} = \| 1-  w_i^r\|_1 + 2\| 1-  w_j^t\|_1 + \frac{10}{M} \sum_{m=1}^M c_m^{t}(m),
\end{align}
where $w_i^r$ and $w_j^t$ are the predicted rotation and translation scores, \textcolor{black}{and $c_m^{t}(m)$ is the $m$-th element in $c_m^{t}$ which means the cost of plane correspondence $\mathcal{P}_m=(\mathcal{P}_m^{(1)}, \mathcal{P}_m^{(2)})$ with the $m$-th pose hypothesis $\xi_m$.}

\paragraph{Final Pose Estimation.} After the hypotheses scoring, we present a \emph{Soft} fusion strategy as the default approach to obtain the final pose. 
In detail, we use the predicted scores $\mathbf{w}_r$, $\mathbf{w}_t$ to obtain the embedding of the final refined pose by:
\begin{equation}\label{eq:feat_m}
    f_{\rm ref}^r = \sum_{m=0}^{M} w_m^r f_m^{r},\, 
    f_{\rm ref}^t = \sum_{m=0}^{M} w_m^t f_m^{t},
\end{equation}
and leverage the layers $\text{LinearRot}(\cdot)$ and $\text{LinearTrans}(\cdot)$ defined in Eqn.~\ref{eq:linear-pose-decoding} again to predict the final refined pose $\xi_{\rm ref} = ({R}_{\rm ref},\mathbf{t}_{\rm ref})$. 

Apart from the \emph{Soft} fusion scheme, there are three alternatives to get the final refined pose from pose hypotheses, which are summarized as follows:
\begin{enumerate}%
    \item[-] \textit{Avg}: It treats all pose hypotheses equally and sets all scores used in Eqn.~\eqref{eq:feat_m} to be $\frac{1}{M+1}$.
    \item[-] \textit{Min-Cost}: It selects the minimal-cost rotation/translation hypothesis according to Eqn.~\eqref{eq:cost_pose} and discard all the remaining hypotheses.
    \item[-] \textit{Max-Score}: It takes the maximal-score rotation/translation hypothesis according to the hypotheses scoring layer as the final pose prediction.
\end{enumerate}
If not explicitly stated, we use the $\textit{Soft}$ fusion strategy for all experiments. We analyze the effectiveness of these fusion strategies in the ablation study. 

We supervise the final pose prediction with the MSE loss. Because there are two initial pose predictions $\xi_0^{\rm reg}$ (\textcolor{black}{from the Regressive Initialization Module}) and $\xi_0^{\rm rec}$ (from the Arbitrary Initialization Module), by taking the predicted plane correspondences $\mathbb{P}_{\rm pred}$ and the ground-truth $\mathbb{P}_{gt}$ into account during training, there will be four predictions of the refined camera pose for the \emph{Soft} fusion scheme. We compute the loss functions $\mathcal{L}_{\rm cam}^{\rm soft}$ and $\mathcal{L}_{\rm score}$ for the refinement module similar to Eqn.~\eqref{eq:cam-loss} and Eqn.~\eqref{eq:score-loss}, but the four predictions are all taken in the computation. To avoid unstable training, the \textit{Avg} fusion strategy is also involved to compute a loss function $\mathcal{L}_{\rm cam}^{\rm avg}$. Finally, the loss function of the refinement module is denoted by: 
\begin{equation}
    \mathcal{L}_{\rm cam}^{\rm ref} = \mathcal{L}_{\rm cam}^{\rm soft}  + \mathcal{L}_{\rm cam}^{\rm avg} + 0.01 \mathcal{L}_{\rm score}.
\end{equation}
The total loss for our NOPE-SAC pose estimation is achieved by linearly adding the losses $\mathcal{L}_{\text{cam}}^{\text{init}}$, $\mathcal{L}_{\text{rec}}$ and $\mathcal{L}_{\text{cam}}^{\text{ref}}$ together as:
\begin{equation}
    \mathcal{L}_{\text{pose}} = \mathcal{L}_{\text{cam}}^{\text{init}} + \mathcal{L}_{\text{rec}} + \mathcal{L}_{\text{cam}}^{\text{ref}}
\end{equation}

\section{Experiments and Analysis}

\begin{figure*}[t!]
\centering
\centering
    \subfigure[Matterport3D~\cite{mp3d} train]{
            \includegraphics[height=0.15\linewidth]{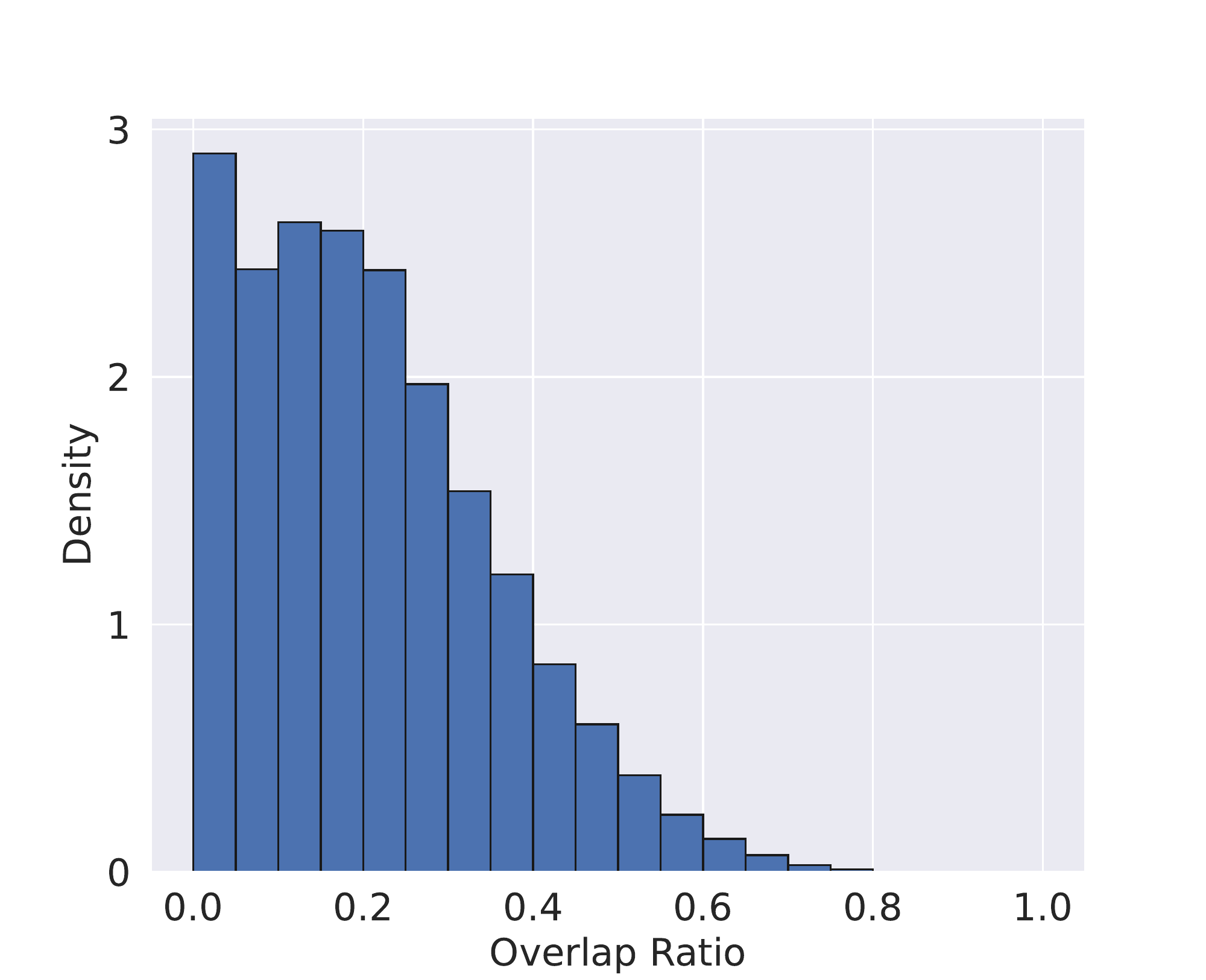}
    }\hfill
    \subfigure[Matterport3D~\cite{mp3d} test]{
            \includegraphics[height=0.15\linewidth]{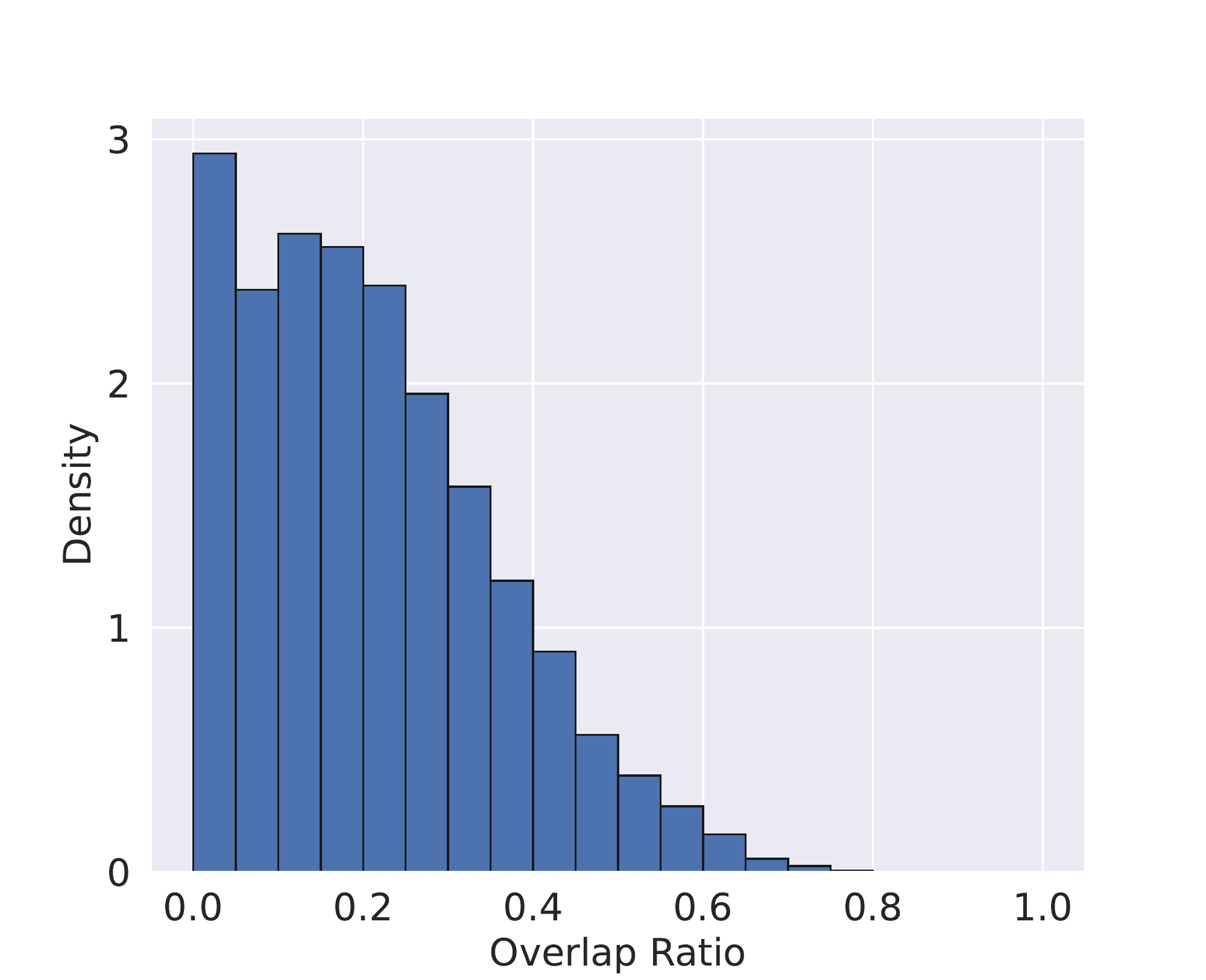}
    }\hfill
    \subfigure[ScanNet~\cite{dai2017scannet} train]{
            \includegraphics[height=0.15\linewidth]{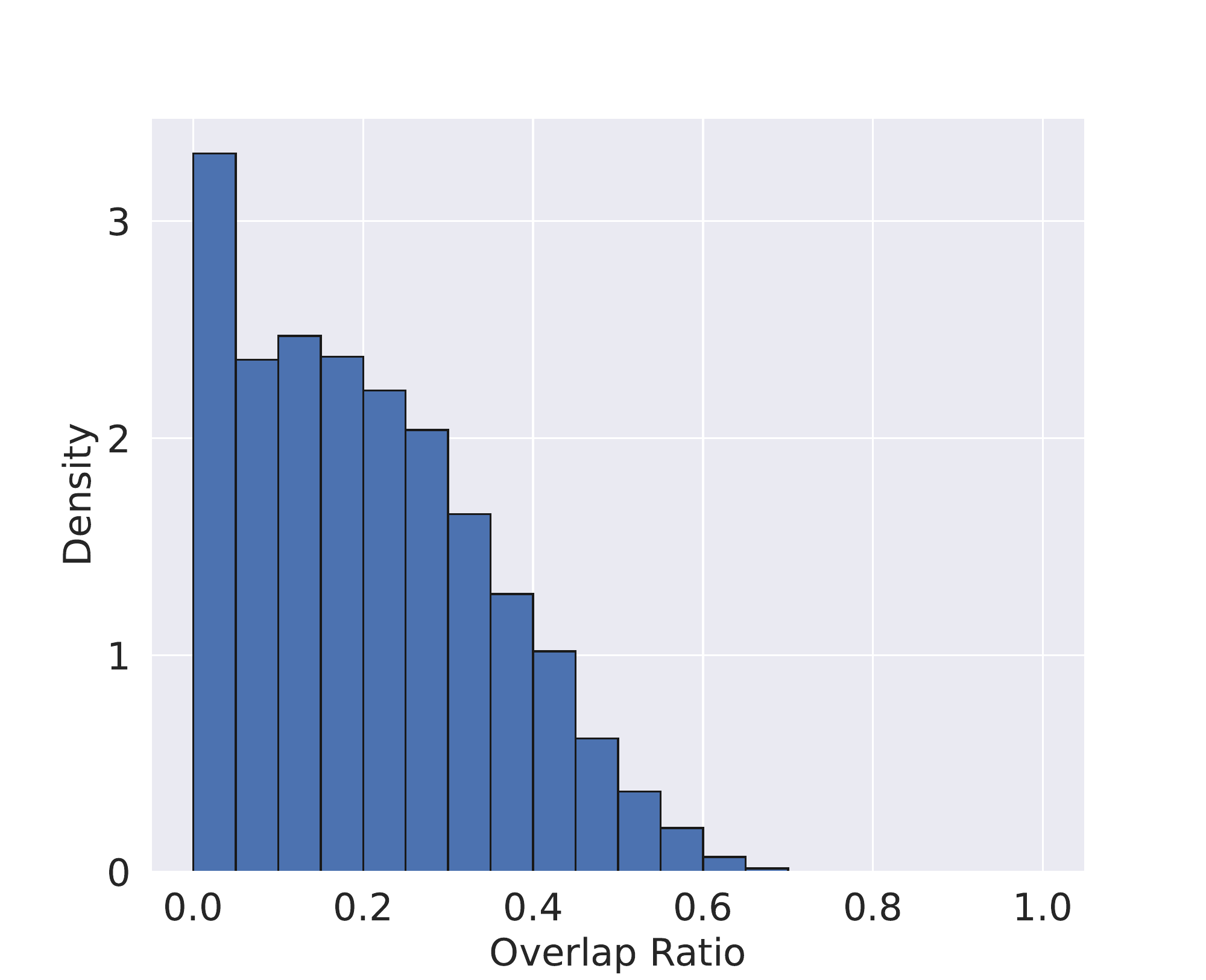}
    }\hfill
    \subfigure[ScanNet~\cite{dai2017scannet} test]{
            \includegraphics[height=0.15\linewidth]{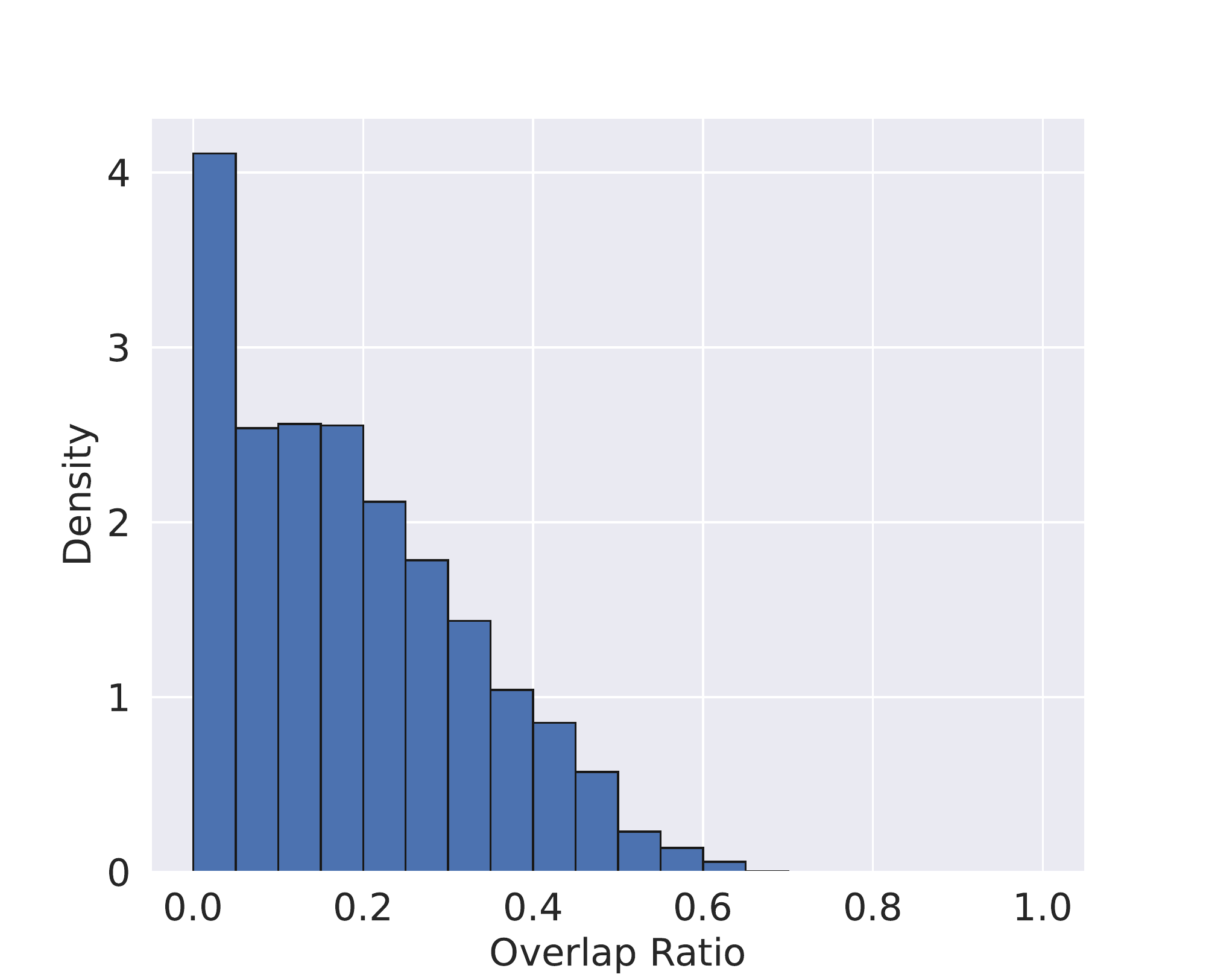}
    }
    
    \subfigure[Rotation train]{
            \includegraphics[height=0.15\linewidth]{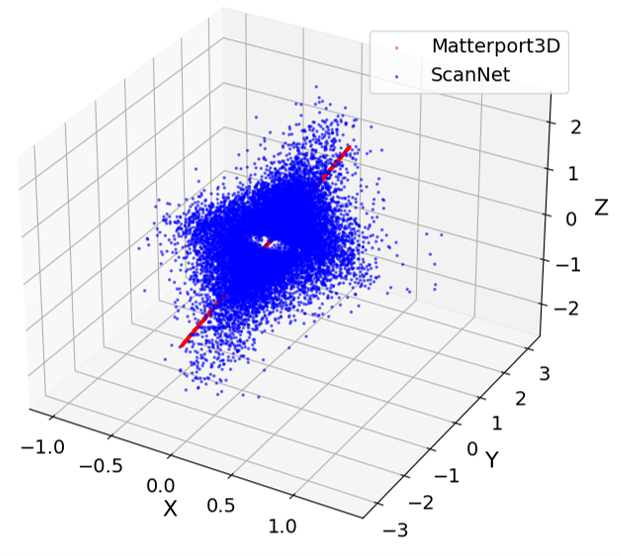}
    }\hfill
    \subfigure[Rotation test]{
            \includegraphics[height=0.15\linewidth]{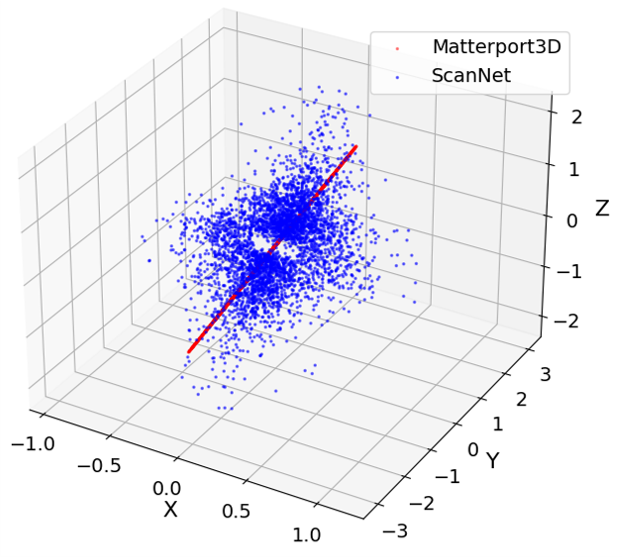}
    }\hfill
    \subfigure[Translation train]{
            \includegraphics[height=0.15\linewidth]{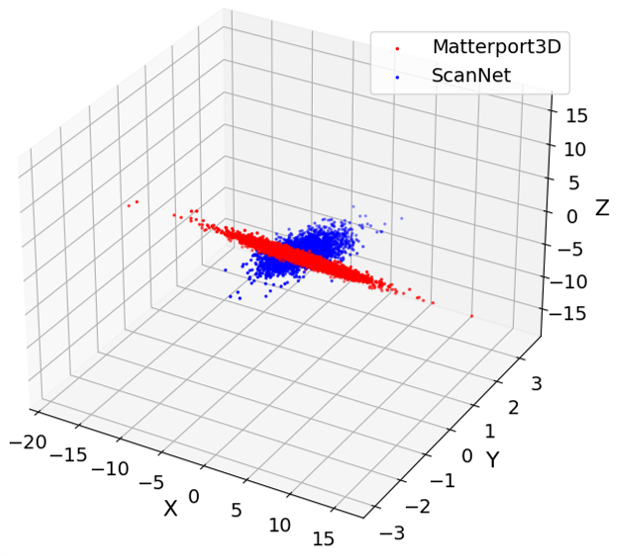}
    }\hfill
    \subfigure[Translation test]{
            \includegraphics[height=0.15\linewidth]{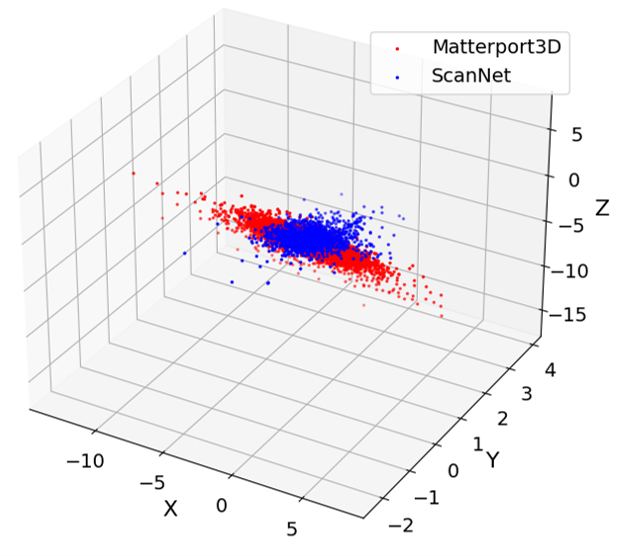}
    }
\vspace{-2mm}
\caption{Dataset analysis of the sparse-view split on the Matterport3D~\cite{mp3d} and the ScanNet~\cite{dai2017scannet} datasets. The top row shows the image overlap ratio on the training and test set of two datasets (from left to right, the average percentages of image overlap are $20.9\%$, $21.0\%$, $20.6\%$ and $18.6\%$). At the bottom, we visualize the rotation and translation distributions on the ScanNet (\textbf{Blue} dot) and the Matterport3D (\textbf{Red} dot) datasets.}
\label{fig:dataset}
\end{figure*}

\subsection{Datasets}
\label{subsec:exp_dataset}
\paragraph{Matterport3D Dataset.}
We use the sparse-view dataset based on Matterport3D~\cite{mp3d} with ground truth camera poses and plane annotations generated by~\cite{jin2021planar}, which contains 31932, 4707, and 7996 image pairs for training, validation, and testing. The size of each image is $480 \times 640$.

\paragraph{ScanNet Dataset.} 
We create a new sparse-view split on the indoor ScanNet~\cite{dai2017scannet} video dataset with plane annotations generated by~\cite{planercnn}. 
The image size is $480 \times 640$. We randomly sample 17237/4051 image pairs from 1210/303 non-overlapping scenes for training/testing. The frame interval within a sampled image pair is at least 20 and 40 frames in the training and test sets, respectively. 

\paragraph{Dataset Analysis.} We first analyze the image overlap on the Matterport3D~\cite{mp3d} and the ScanNet~\cite{dai2017scannet} datasets. As shown in the top row of Fig.~\ref{fig:dataset}, our split on the ScanNet dataset contains more low-overlap image pairs than the split on the Matterport3D dataset created by~\cite{jin2021planar}. Furthermore, we visualize the distributions of rotations (represented as rotation vectors) and translations on the training and test sets of these two datasets. As shown in the bottom row of Fig.~\ref{fig:dataset}, although the translation range on the ScanNet dataset is smaller than that on the Matterport3D dataset, the rotation distribution on the ScanNet dataset is much more complex. 
This observed difference in complexity underscores the importance of rigorous testing and benchmarking. A more challenging rotation distribution, as found in our ScanNet dataset split, necessitates advanced algorithms and robust methodologies to ensure accurate pose estimation.

\subsection{Evaluation Metrics}
\paragraph{Metrics for Plane Matching.} We use precision (P), recall (R) and F-score (F) to evaluate the matching performance. A predicted plane correspondence is regarded as a true positive if it can be matched to a ground truth plane correspondence with mask IoU $\ge$ 0.5 in both two images.

\paragraph{Metrics for Camera Pose Estimation.} We evaluate camera poses with the rotation angle and translation distance errors as used in~\cite{jin2021planar,posenet,deepsfm}. The reported metrics include the mean and median errors and the percentage of errors lower than a threshold. 

\paragraph{Metrics for 3D Plane Reconstruction.} According to SparsePlanes~\cite{jin2021planar}, the matched planes are merged, and all 3D planes from two views are converted to the same coordinate system. Then, the reconstructed full scene is evaluated like a detection problem with the metric of average precision (AP). A reconstructed 3D plane is regarded as a true positive if the following conditions of (1) its mask IoU $\ge$ 0.5, (2) its plane normal angle error in degree is less than $\alpha^{\circ}$, and (3) its plane offset error in meters is less than $\beta$ are all satisfied.

\begin{figure*}[htb!]
\centering
\centering
    \subfigure[Image 1]{
    \begin{minipage}[b]{0.13\linewidth}
        \includegraphics[height=52pt]{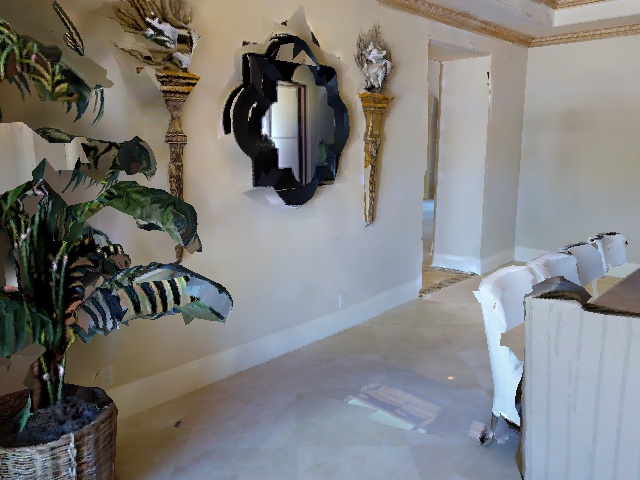}\vspace{2pt}
        \includegraphics[height=52pt]{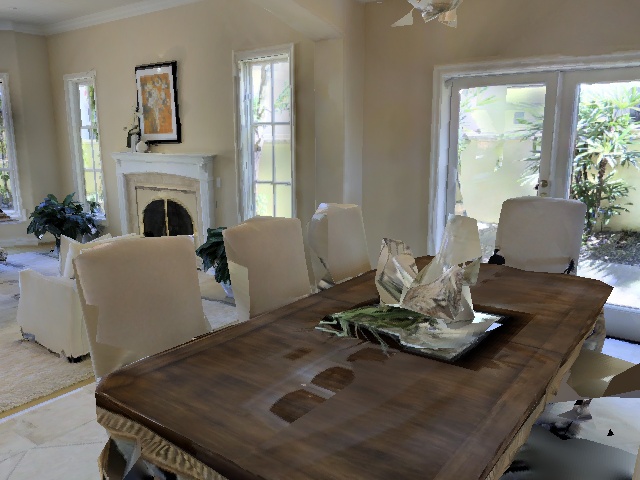}\vspace{2pt}
        \includegraphics[height=52pt]{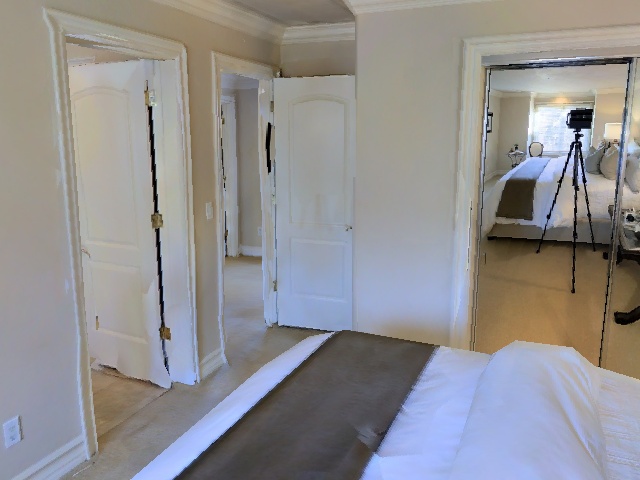}\vspace{2pt}

        \includegraphics[height=52pt]{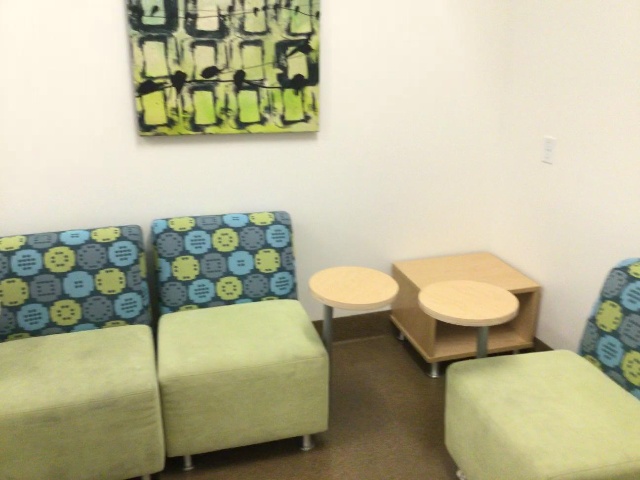}\vspace{2pt}
        \includegraphics[height=52pt]{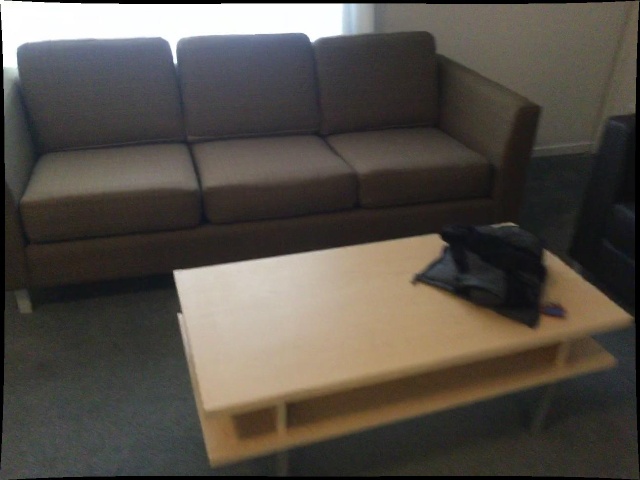}\vspace{2pt}
        \includegraphics[height=52pt]{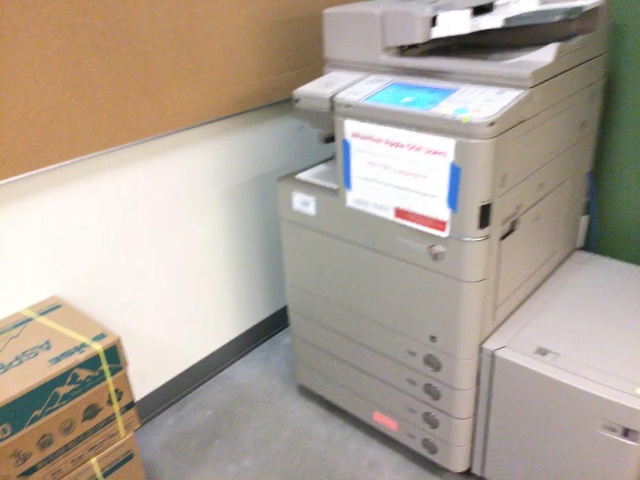}\vspace{2pt}
    \end{minipage}}
    \subfigure[Image 2]{
    \begin{minipage}[b]{0.13\linewidth}
        \includegraphics[height=52pt]{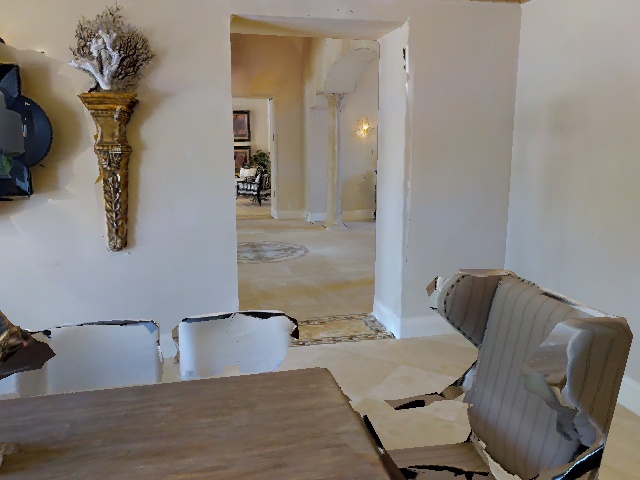}\vspace{2pt}
        \includegraphics[height=52pt]{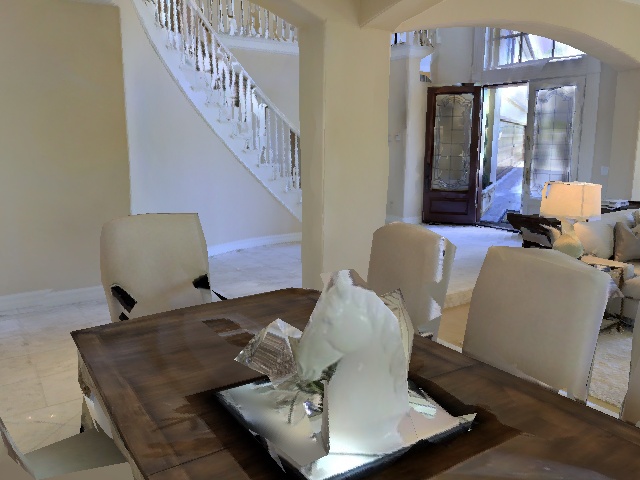}\vspace{2pt}
        \includegraphics[height=52pt]{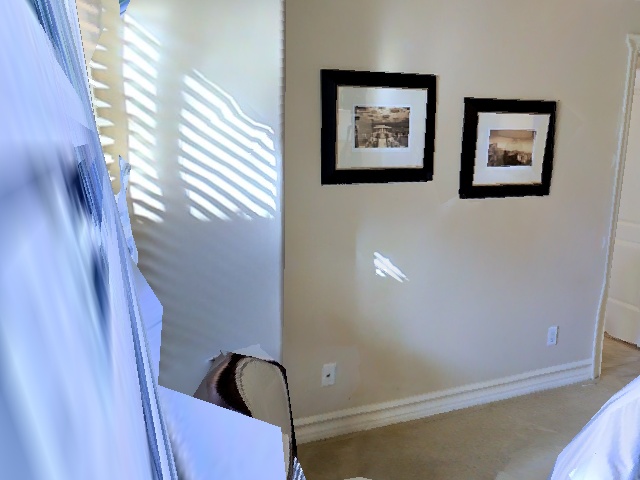}\vspace{2pt}

        \includegraphics[height=52pt]{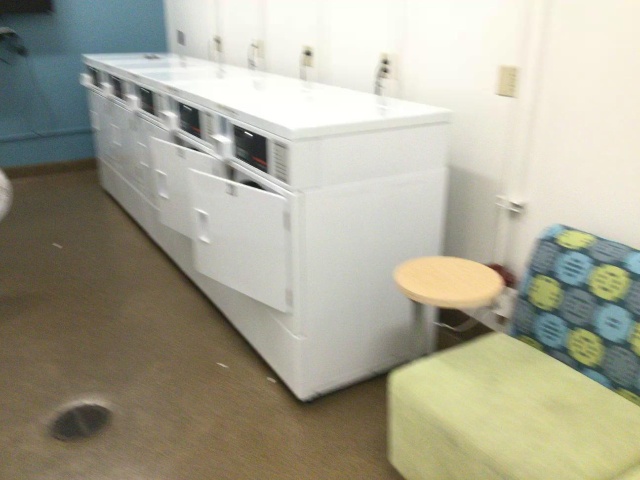}\vspace{2pt}
        \includegraphics[height=52pt]{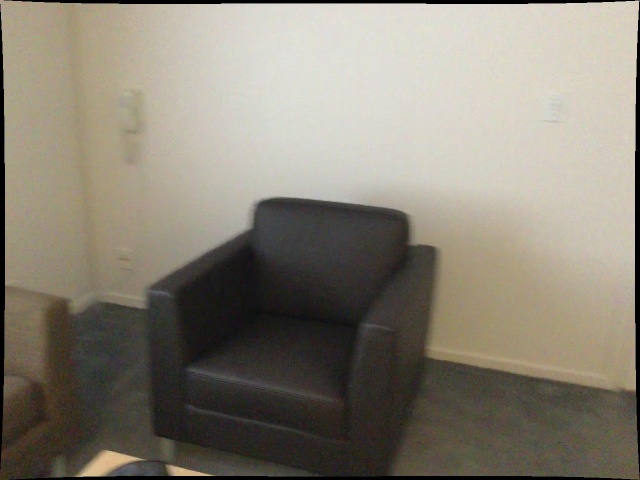}\vspace{2pt}
        \includegraphics[height=52pt]{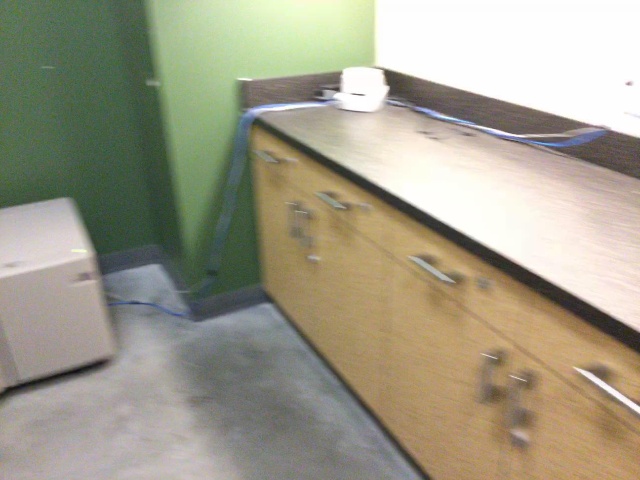}\vspace{2pt}
    \end{minipage}}
    \subfigure[Plane Correspondences]{
    \begin{minipage}[b]{0.275\linewidth}
        \includegraphics[height=52pt]{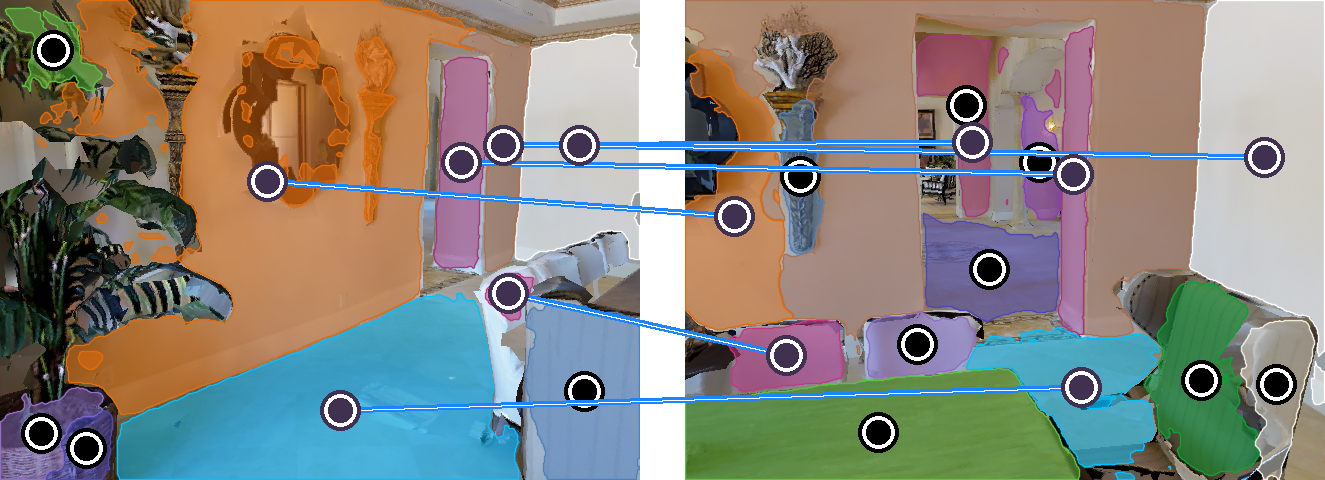}\vspace{2pt}
        \includegraphics[height=52pt]{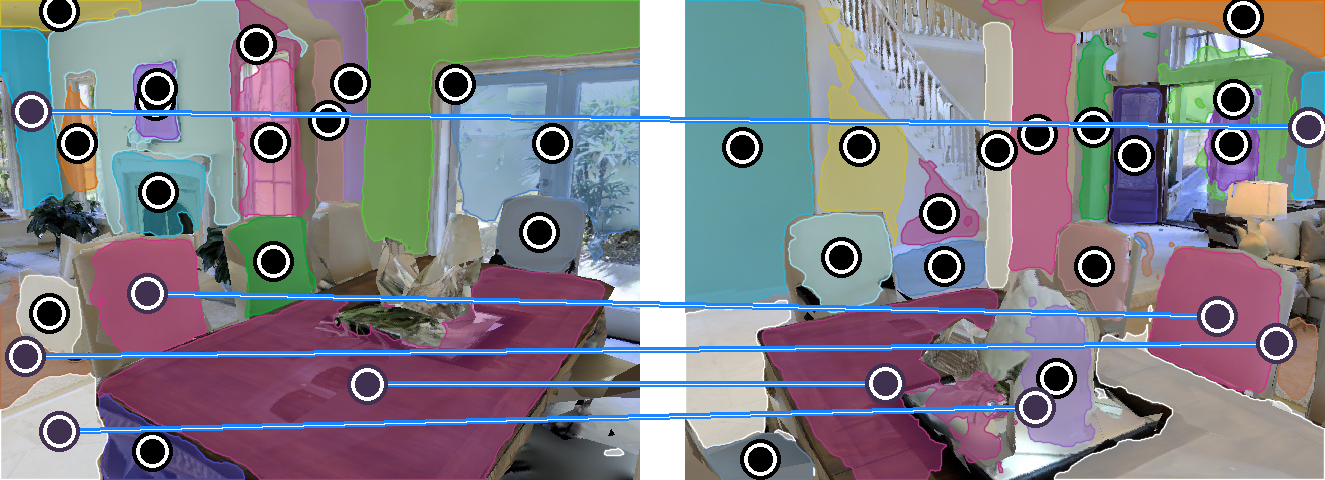}\vspace{2pt}
        \includegraphics[height=52pt]{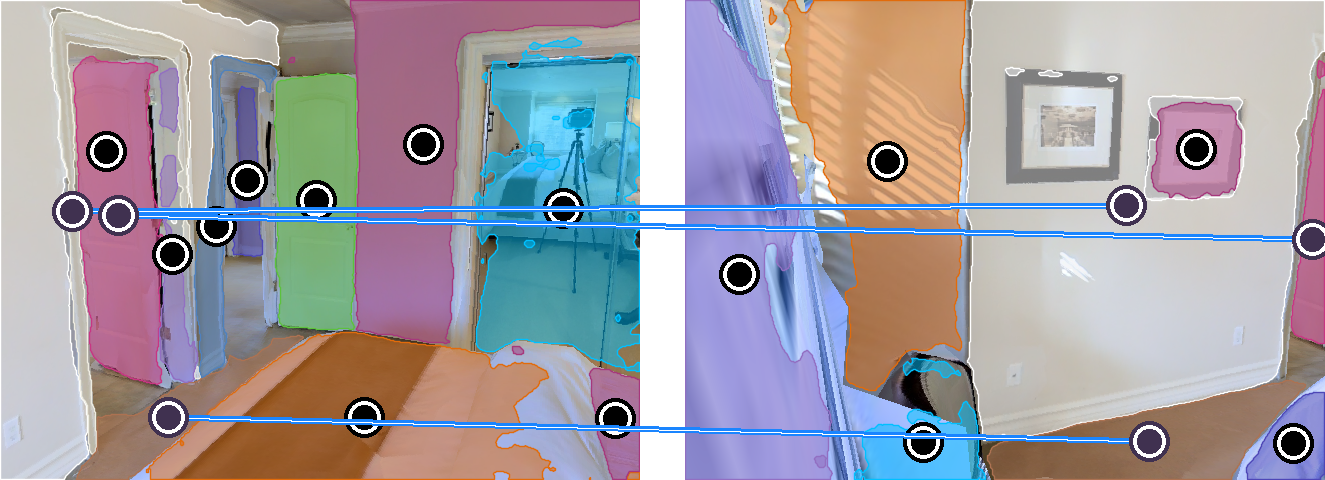}\vspace{2pt}
        
        \includegraphics[height=52pt]{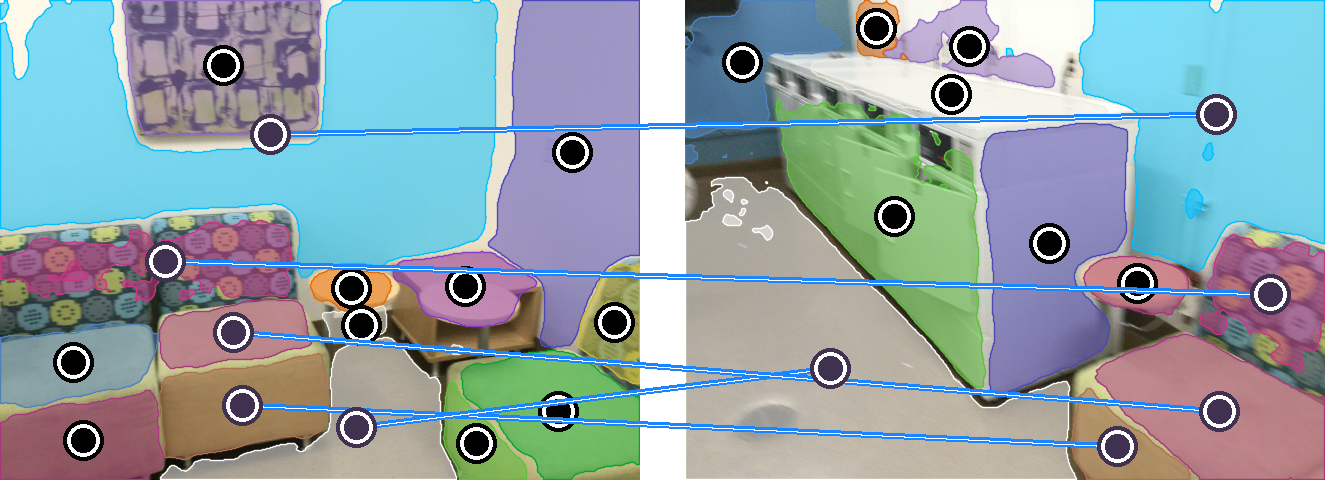}\vspace{2pt}
        \includegraphics[height=52pt]{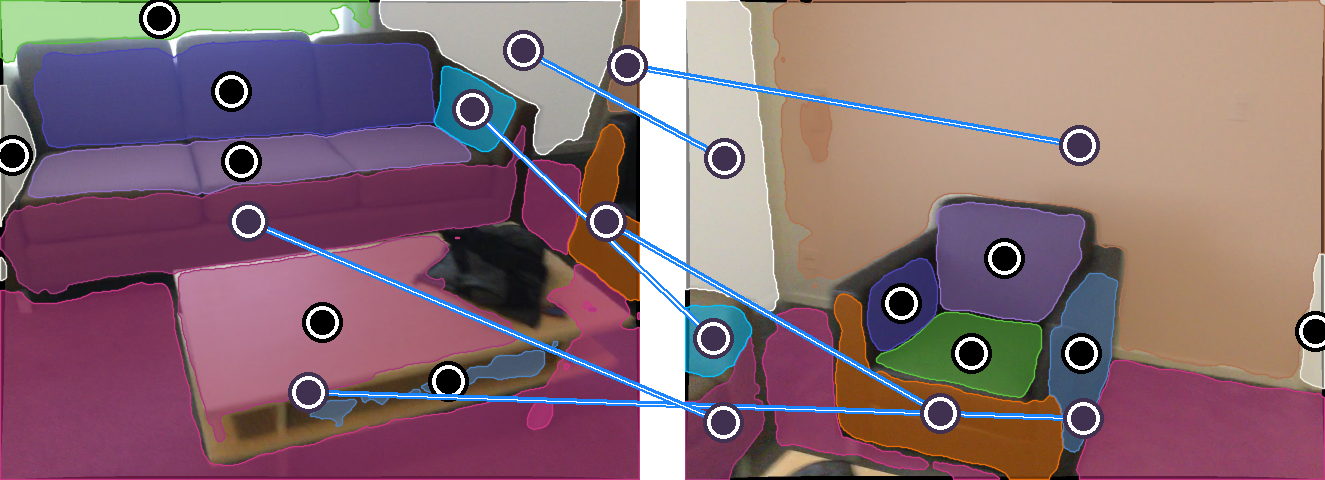}\vspace{2pt}
        \includegraphics[height=52pt]{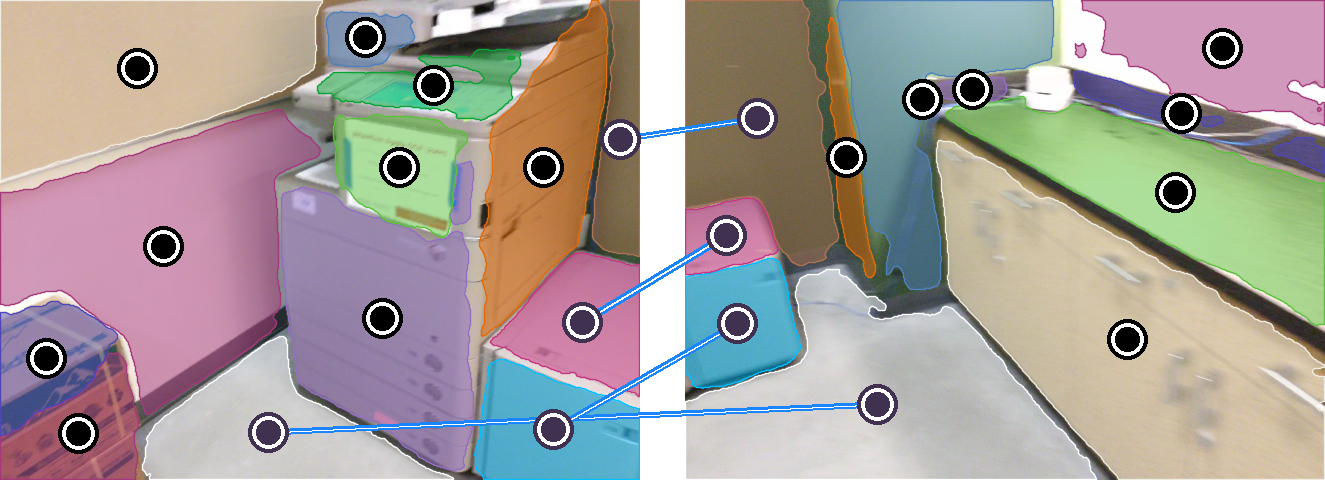}\vspace{2pt}
    \end{minipage}}
    \subfigure[Poses (viewpoint 1)]{
    \begin{minipage}[b]{0.17\linewidth}
        \includegraphics[height=52pt]{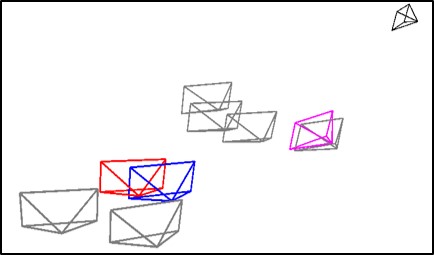}\vspace{2pt}
        \includegraphics[height=52pt]{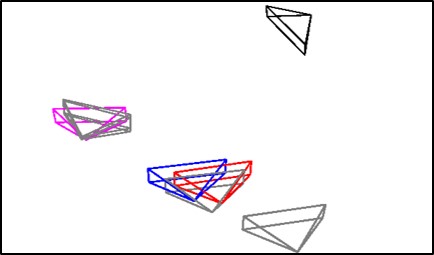}\vspace{2pt}
        \includegraphics[height=52pt]{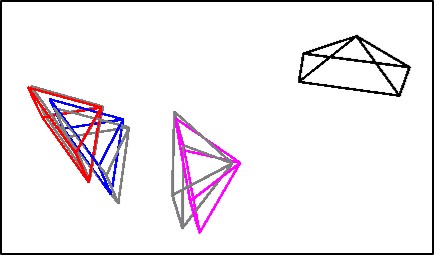}\vspace{2pt}
        
        \includegraphics[height=52pt]{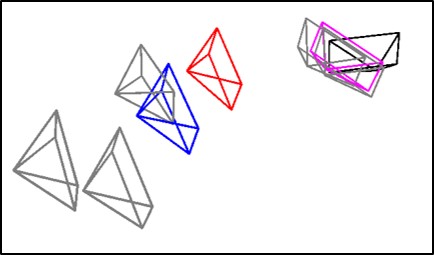}\vspace{2pt}
        \includegraphics[height=52pt]{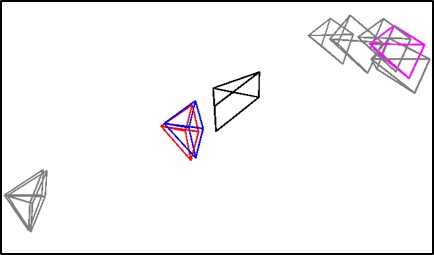}\vspace{2pt}
        \includegraphics[height=52pt]{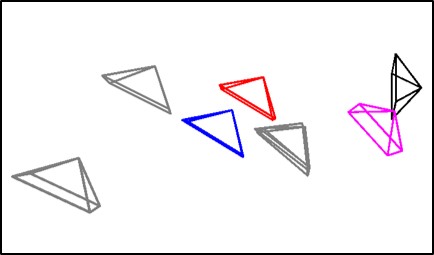}\vspace{2pt}
    \end{minipage}}
    \subfigure[Poses (viewpoint 2)]{
    \begin{minipage}[b]{0.17\linewidth}
        \includegraphics[height=52pt]{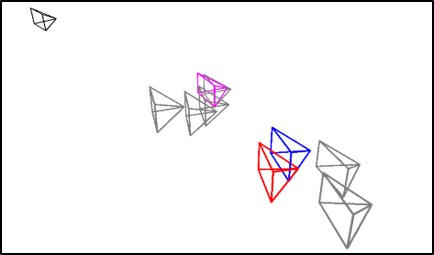}\vspace{2pt}
        \includegraphics[height=52pt]{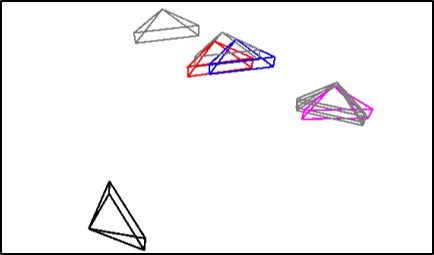}\vspace{2pt}
        \includegraphics[height=52pt]{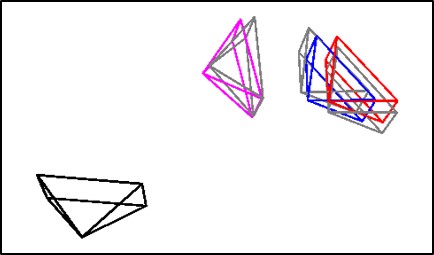}\vspace{2pt}
        
        \includegraphics[height=52pt]{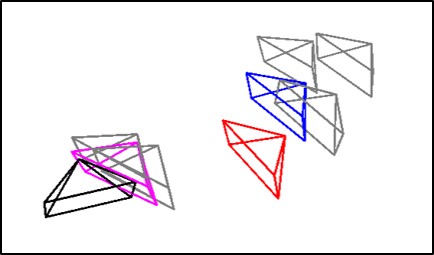}\vspace{2pt}
        \includegraphics[height=52pt]{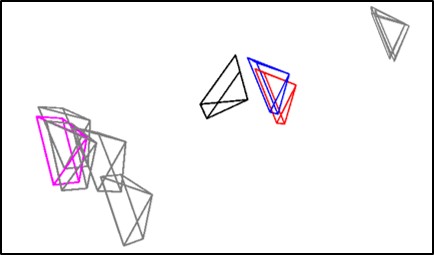}\vspace{2pt}
        \includegraphics[height=52pt]{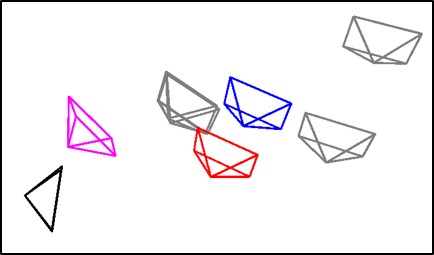}\vspace{2pt}
    \end{minipage}}
\caption{Qualitative results of the refined poses and plane correspondences on the Matterport3D~\cite{mp3d} dataset (first three rows) and the ScanNet~\cite{dai2017scannet} dataset (last three rows). The \textbf{Pink}, \textbf{Blue} and \textbf{Red} frustums show the initial, the refined, and the ground truth cameras of the first image respectively. \textbf{Gray} frustums show the one-plane pose hypotheses of the first image generated from plane correspondences. \textbf{Black} frustums show the camera of the second image.}
\label{fig:poses}
\end{figure*}

\subsection{Implementation Details}\label{sec:implementation}
Our NOPE-SAC is implemented with Detectron2~\cite{wu2019detectron2} and the AdamW optimizer~\cite{AdamW} is used for training with a batch size of 16. 
Since our network consists of multiple neural modules for 3D plane detection and matching, camera pose initialization and pose refinement with one-plane pose hypotheses, we train the entire network in three stages on the Matterport3D~\cite{mp3d} training set. In the first stage, only the 3D plane detection module is optimized for 12k iterations with a fixed learning rate of $10^{-4}$. Next, in the second training stage, we introduce the plane matching module and the camera pose initialization modules (including RIM and AIM), and train them for 50k iterations. The learning rate in this stage is initialized to $10^{-4}$ and divided by 10 at 34k and 44k iterations.
In the last stage, we incorporate the camera pose refinement module and continue the training for an additional 14k iterations. The learning rate in this stage is initialized to $10^{-4}$ for the first 6k iterations and then decayed to $10^{-5}$ for the remaining 8k iterations.

On the ScanNet~\cite{dai2017scannet} dataset, we finetune the model trained on the Matterport3D dataset with two additional stages. In the first stage, we train the 3D plane detection module, the plane matching module, and the camera pose initialization modules (including RIM and AIM) for 20K iterations. The initial learning rate is set to $10^{-4}$ and divided by 10 after 4.4K iterations. In the second stage, we add the camera pose refinement module and train the entire network for 15K iterations. The initial learning rate is set to $10^{-4}$ and divided by 10 after 2.2K iterations.

When training the Arbitrary Initialization Module (AIM), we generate the input translation $\textbf{t}_*\in \mathbb{R}^3$ and rotation ${R}_*\in \mathbb{R}^4$ as used in Eqn.~\eqref{eq:feat_initial} from a uniform distribution. Specifically, we first represent the input rotation as a rotation vector $\textbf{v}_* \in \mathbb{R}^3$. Then, we randomly sample each axis of $\textbf{v}_*$ and $\textbf{t}_*$ from the uniform distribution $U(-2.5, 2.5)$. At last, we convert the sampled rotation vector $\textbf{v}_*$ to a unit quaternion ${R}_*$.

\begin{table*}[t!]
\centering
\caption{Comparison of camera poses on the Matterport3D~\cite{mp3d} dataset and the ScanNet~\cite{dai2017scannet} dataset.} 
\resizebox{0.9\linewidth}{!}{ 
    \begin{tabular}{l|ccccc|ccccc}
    \toprule
        \multicolumn{1}{c|}{\multirow{2}{*}{Method}} & \multicolumn{5}{c|}{Translation} & \multicolumn{5}{c}{Rotation} \\ 
                                & Med.~$\downarrow$ & Mean~$\downarrow$ & ($\le$1m)~$\uparrow$ & ($\le$0.5m)~$\uparrow$& ($\le$0.2m)~$\uparrow$ & Med.~$\downarrow$ & Mean~$\downarrow$ & ($\le 30^{\circ}$)~$\uparrow$ & ($\le 15^{\circ}$)~$\uparrow$ & ($\le 10^{\circ}$)~$\uparrow$\\
        \midrule
        \multicolumn{10}{c}{Matterport3D~\cite{mp3d} dataset} \\\midrule
        SuperGlue~\cite{SuperGlue} & - & - & - & - & - & 3.88 & 24.17 & 77.8\% & 71.0\% & 65.7\%\\
        \textcolor{black}{LoFTR~\cite{LoFTR}} & - & - & - & - & - & 5.85 & 33.13 & 67.0\% & 61.0\% & 57.0\% \\\midrule
        SparsePlanes~\cite{jin2021planar} & 0.63& 1.15 & 66.6\% & 40.4\% & 11.9\% & 7.33 & 22.78& 83.4\% & 72.9\% & 61.2\% \\
        PlaneFormers~\cite{planeformer} & 0.66 & 1.19 & 66.8\% & 36.7\% & 8.7\% & 5.96 & 22.20 & 83.8\% & 77.6\% & 68.0\% \\\midrule
        SparsePlanes-TR~\cite{jin2021planar,planeTR} & 0.61 & 1.13 & 67.3\% & 41.7\% & 12.2\% & 6.87 & 22.17 & 83.8\% & 74.5\% & 63.3\% \\
        PlaneFormers-TR~\cite{planeformer,planeTR} & 0.64 & 1.17 & 67.9\% & 38.7\% & 8.9\% & 5.28 & 21.90 & 83.9\% & 79.0\% & 70.8\% \\\midrule
        NOPE-SAC-Cls (ours) & 0.66  &  1.20  &  65.6\%  &   37.9\%  &  9.8\%  & 2.98   &   19.68  &  84.9\%   &  83.1\%  &  80.2\% \\
        NOPE-SAC-Reg (ours) & \textbf{0.52}  &  \textbf{0.94}  &  \textbf{73.2\%}  &   \textbf{48.3\%}  &   \textbf{16.2\%} & \textbf{2.77}   &   \textbf{14.37}  &  \textbf{89.0\%}   &  \textbf{86.9\%}  &  \textbf{84.0\%} \\
        \midrule
        \multicolumn{10}{c}{ScanNet~\cite{dai2017scannet} dataset} \\
        \midrule
        SuperGlue~\cite{SuperGlue} & - & - & - & - & - & 10.90 & 31.00 & 67.8\% & 56.0\% & 48.4\%\\
        \textcolor{black}{LoFTR~\cite{LoFTR}} & - & - & - & - & - & \textbf{5.49} & 27.13 & 71.0\% & 63.3\% & 58.4\%\\\midrule
        SparsePlanes~\cite{jin2021planar}  & 0.56& 0.81& 73.7\% & 44.6\% & 10.7\% & 15.46& 33.38& 70.5\% & 48.7\% & 28.0\% \\
        PlaneFormers~\cite{planeformer} & 0.55 & 0.81 & 75.3\% & 45.5\% & 11.3\% & 14.34 & 32.08 & 73.2\% & 52.1\% & 32.3\% \\\midrule
        SparsePlanes-TR~\cite{jin2021planar,planeTR} &  0.57 &  0.82 &  73.4\% &  43.6\% & 10.1\% & 14.57 & 32.36 & 72.8\% & 51.2\% & 30.1\% \\
        PlaneFormers-TR~\cite{planeformer,planeTR} & 0.53 & 0.79 & 76.2\% & 47.0\% & 11.4\% & 13.81 & 31.58 & 74.5\% & 54.1\% & 33.6\% \\\midrule
        NOPE-SAC-Cls (ours) & 0.49  & 0.76  & 77.5\%  & 50.9\%   &  14.1\% & 9.01  & 27.84  &  77.9\%  &  69.5\%  & 55.1\% \\
        NOPE-SAC-Reg (ours) & \textbf{0.41}  & \textbf{0.65}  & \textbf{82.0\%}  &  \textbf{59.1\%}  &  \textbf{21.2\%} & 8.27  & \textbf{22.12}  &  \textbf{82.6\%}  &  \textbf{73.2\%}  &  \textbf{59.5\%} \\ 
        \bottomrule
    \end{tabular}
}
\label{tab:cam_pose}
\end{table*}

\subsection{Baseline Configurations}\label{subsec:baselines}
We compare our method with the state-of-the-art learning solutions, including SparsePlanes~\cite{jin2021planar} and PlaneFormers~\cite{planeformer} for plane matching, pose estimation, and planar 3D reconstruction, as well as two keypoint-based solutions SuperGlue~\cite{SuperGlue} and LoFTR~\cite{LoFTR}.
For the SparsePlanes~\cite{jin2021planar} and PlaneFormers~\cite{planeformer}, we additionally set two variant baselines to compare the performance gain with the same 3D plane detection module. 

\paragraph{NOPE-SAC Models.} 
For our method, there are two models, {NOPE-SAC-Reg} and {NOPE-SAC-Cls}. The {NOPE-SAC-Reg} is the main version that predicts the initial pose in a regressive way as introduced in Sec.~\ref{subsec:pose_initial} while {NOPE-SAC-Cls} directly uses the Top-1 classification pose predicted by SparsePlanes~\cite{jin2021planar} as the initial pose. Both versions are evaluated in experiments to demonstrate the effectiveness and flexibility of our design.

\paragraph{SparsePlanes and SparsePlanes-TR}.
SparsePlanes~\cite{jin2021planar} is the first sparse-view planar 3D reconstruction approach. It involves detecting 3D planes in each view and predicting the likelihood of precomputed pose templates using deep neural networks. 
Then, a discrete optimization problem is solved to determine the optimal plane correspondences and camera pose template by taking the geometric and appearance consistency between 3D planes across views, as well as the classification likelihoods of pose templates into account. 
Subsequently, a continuous optimization problem is formulated to refine the camera pose with the assistance of SIFT~\cite{SIFT} keypoints. 
Due to the different 3D plane detection modules used in SparsePlanes~\cite{jin2021planar} and NOPE-SAC, we create a variant baseline called SparsePlanes-TR to ensure fair comparisons. In SparsePlanes-TR, we replace the plane detection module with PlaneTR~\cite{planeTR} as our network used.
As the official implementation of SparsePlanes~\cite{jin2021planar} does not provide results on the ScanNet~\cite{dai2017scannet} dataset, we train these baselines on the ScanNet~\cite{dai2017scannet} dataset by ourselves.

\paragraph{PlaneFormers and PlaneFormers-TR}. PlaneFormers~\cite{planeformer} is the prior art built upon on SparsePlanes~\cite{jin2021planar}. It selects Top-9 classification poses from~\cite{jin2021planar} as hypotheses. For each pose hypothesis, PlaneFormers jointly estimates the pose score and plane matching cost using a neural network. The pose with the best score is selected and refined with an extra estimated pose residual. The plane correspondences are achieved by conducting an offline Hungarian algorithm on the matching cost. Like SparsePlanes~\cite{jin2021planar}, we implemented the results of PlaneFormers on the ScanNet~\cite{dai2017scannet} dataset by ourselves. In PlaneFormers-TR, we use the planes detected by PlaneTR~\cite{planeTR}, the same as what our NOPE-SAC used to ensure fairness of comparisons.

\paragraph{SuperGlue and LoFTR}. 
These are two strong baselines for keypoints or local feature matching with neural networks. The camera poses are calculated from matched point correspondences with `5-point solver~\cite{fivePointSolver} + RANSAC~\cite{RANSAC}'. 
Since SuperGlue~\cite{SuperGlue} and LoFTR~\cite{LoFTR} lack scale information in translations, we do not report their translation errors. 
When SuperGlue/LoFTR fails to estimate camera poses, we set their rotation results as identity matrices like~\cite{jin2021planar}.
For the results of plane reconstruction, we use the ground truth scales in translations and match planes detected by PlaneTR~\cite{planeTR} with our plane matching module.

\begin{table}[]
\centering
\caption{\textcolor{black}{Comparison of plane matching results.}}
\vspace{-2mm}
\resizebox{0.9\linewidth}{!}{
    \begin{tabular}{l|ccc|ccc}
    \toprule
        \multicolumn{1}{c|}{\multirow{2}{*}{Methods}} & \multicolumn{3}{c|}{Matterport3D~\cite{mp3d}} & \multicolumn{3}{c}{ScanNet~\cite{dai2017scannet}}\\ 
                                & P~$\uparrow$ & R~$\uparrow$ & F~$\uparrow$ & P~$\uparrow$ & R~$\uparrow$ & F~$\uparrow$\\
        \midrule
        SparsePlanes-TR~\cite{jin2021planar,planeTR} & 44.4 & 49.0 & 46.6 & 37.5 & 45.7 & 41.2\\
        PlaneFormers-TR~\cite{planeformer, planeTR} & \textbf{55.0} & 46.2 & 50.2 & \textbf{49.1} & 40.4 & 44.3\\
        \midrule
        NOPE-SAC-Cls (ours) & 49.2 & 50.8 & 50.0& 43.4 & 46.9 & 45.1 \\ 
        NOPE-SAC-Reg (ours) & 49.9& \textbf{51.5} & \textbf{50.7} & 44.3& \textbf{48.0} & \textbf{46.0} \\
        \bottomrule
    \end{tabular}
}
\label{tab:matching_compare}
\end{table}

\subsection{Main Results}
In this section, we position our methodology alongside several leading-edge techniques across various dimensions. We first commence with a quantitative assessment of diverse plane matching networks. Subsequently, our attention shifts to the primary objectives: camera pose estimation and 3D planar reconstruction, for which we provide both quantitative and qualitative results. Finally, we present data on the model size and inference time to show the efficiency of our approach.
\subsubsection{Comparison of Plane Matching}
\label{subsec:exp_plane_matching}
We compare our method with SparsePlanes-TR~\cite{jin2021planar,planeTR} and PlaneFormers-TR~\cite{planeformer, planeTR}, in which we use the same single-view 3D plane detection results for fair comparisons. As shown in Tab.~\ref{tab:matching_compare}, both two versions of our method significantly outperform SparsePlanes-TR which uses an offline Hungarian algorithm. When compared to PlaneFormers-TR which utilizes neural networks for matching, our approach achieves competitive F-score results and demonstrates superior performance in terms of matching recall.

\begin{table}[t!]
\centering
\caption{Detailed camera pose comparison with SparsePlanes-TR~\cite{jin2021planar,planeTR}. `Cls. Top-1' means the Top-1 classification pose of SparsePlanes~\cite{jin2021planar}. `Con.' means the continuous optimization proposed by SparsePlanes.}
\vspace{-2mm}
\resizebox{0.9\linewidth}{!}{ 
    \begin{tabular}{l|ccc|ccc}
    \toprule
        \multicolumn{1}{c|}{\multirow{2}{*}{Method}} & \multicolumn{3}{c|}{Translation} & \multicolumn{3}{c}{Rotation} \\ 
                                & Med.~$\downarrow$ & Mean~$\downarrow$ & ($\le$0.5m)~$\uparrow$& Med.~$\downarrow$ & Mean~$\downarrow$ & ($\le 15^{\circ}$)~$\uparrow$ \\
        \midrule
        \multicolumn{7}{c}{Matterport3D~\cite{mp3d} dataset} \\
        \midrule
        Cls. Top-1                   & 0.90 &  1.40 & 21.1\% & 7.65 & 24.57 & 71.7\%  \\
        \cite{jin2021planar, planeTR} w/o Con. & 0.88 &  1.35 & 21.9\% & 7.17 & 22.36 & 74.6\%  \\
        \cite{jin2021planar,planeTR} & \textbf{0.61} & \textbf{1.13} & \textbf{41.7\%} & 6.87 & 22.17 & 74.5\%  \\
        NOPE-SAC-Cls                 & 0.66 &  1.20 & 37.9\% & \textbf{2.98} & \textbf{19.68} & \textbf{83.1\%}  \\
        \midrule
        \multicolumn{7}{c}{ScanNet~\cite{dai2017scannet} dataset} \\
        \midrule
        Cls. Top-1                   & 0.56 & 0.83 & 43.9\% & 15.21 & 33.16 & 49.5\%  \\
        \cite{jin2021planar,planeTR} w/o Con.& 0.55 & 0.83 & 44.2\% & 14.55 & 32.36 & 51.3\%  \\
        \cite{jin2021planar,planeTR} & 0.57 & 0.82 & 43.6\% & 14.57 & 32.36 & 51.2\%  \\
        NOPE-SAC-Cls                 & \textbf{0.49} & \textbf{0.76} & \textbf{50.9\%} & \textbf{9.01}  & \textbf{27.84} & \textbf{69.5\%}  \\
        \bottomrule
    \end{tabular}
}
\label{tab:cam_pose_detail}
\end{table}

\begin{table*}[!t]
\centering
\caption{Average Precision (AP) of 3D plane reconstruction conditioned with mask IoU, normal angle error, and offset distance error. The threshold of mask IoU is fixed to 0.5. `All' means we consider all three conditions. `-Offset' and `-Normal' mean we ignore the offset and the normal conditions respectively. `Con.' means the continuous optimization proposed by SparsePlanes~\cite{jin2021planar}.}
\vspace{-2mm}
\resizebox{0.92\linewidth}{!}{ 
    \begin{tabular}{l|ccc|ccc|ccc}
    \toprule
        \multicolumn{1}{c|}{\multirow{2}{*}{Method}} & \multicolumn{3}{c|}{Offset$\le$1m,~Normal$\le 30^{\circ}$} & \multicolumn{3}{c|}{Offset$\le$0.5m,~Normal$\le 15^{\circ}$} & \multicolumn{3}{c}{Offset$\le$0.2m,~Normal$\le 5^{\circ}$} \\
        & All & -Offset & -Normal & All & -Offset & -Normal & All & -Offset & -Normal \\\midrule
        \multicolumn{10}{c}{Matterport3D~\cite{mp3d} dataset} \\\midrule
        SuperGlue-TR~\cite{SuperGlue, planeTR}  & 39.51 & 44.11 & 44.08 & 28.98 & 37.53 & 34.40 & 11.29 & 21.87 & 17.27  \\
        \textcolor{black}{LoFTR-TR~\cite{LoFTR, planeTR}}  & 35.71 & 40.74 & 41.36 & 25.84 & 34.04 & 31.48 & 10.74 & 20.51 & 16.05  \\\midrule
        SparsePlanes~\cite{jin2021planar} w/o Con. & 35.87 & 42.13 & 38.80 & 23.36 & 35.34 & 27.48 & 8.07 & 17.28 & 12.99 \\ 
        SparsePlanes~\cite{jin2021planar} & 36.02 & 42.01 & 39.04 & 23.53 & 35.25 & 27.64 & 6.76 & 17.18 & 11.52\\
        PlaneFormers~\cite{planeformer} & 37.62 & 43.19 & 40.36 & 26.10 & 36.88 & 29.99 & 9.44 & 18.82 & 14.78 \\\midrule
        SparsePlanes-TR~\cite{jin2021planar,planeTR} w/o Con. & 39.91 & 46.50 & 42.53 & 27.37 & 40.79 & 31.03 & 9.99 & 22.80 & 14.64 \\
        SparsePlanes-TR~\cite{jin2021planar,planeTR} & 40.35 & 46.39 & 43.03 & 27.81 & 40.65 & 31.38 & 9.02 & 22.80 & 13.66 \\
        PlaneFormers-TR~\cite{planeformer,planeTR} & 41.87 & 47.50 & 44.43 & 30.78 & 42.82 & 34.03 & 12.45 & 25.98 & 17.34 \\\midrule
        NOPE-SAC-Cls Init. (ours) & 38.94 & 46.60 & 41.96 & 26.17 & 40.48 & 29.89 & 9.89 & 22.55 & 14.29 \\
        NOPE-SAC-Cls Ref. (ours) & 41.92 & 48.18 & 44.01 & 31.36 & 44.24 & 34.01 & 13.59 & 30.05 & 17.45 \\\midrule
        NOPE-SAC-Reg Init. (ours) &  40.07 & 46.03 & 43.59 & 26.78 & 36.76 & 31.95 & 10.09 & 19.09 & 15.55 \\
        NOPE-SAC-Reg Ref. (ours) &  \textbf{43.29} & \textbf{49.00} & \textbf{45.32} & \textbf{32.61} & \textbf{44.94} & \textbf{35.36} & \textbf{14.25} & \textbf{30.39} & \textbf{18.37} \\
        \midrule
        
        \multicolumn{10}{c}{ScanNet~\cite{dai2017scannet} dataset} \\\midrule
        SuperGlue-TR~\cite{SuperGlue, planeTR} & 33.20 & 33.77 & 43.40 & 22.78 & 24.94 & 36.72 & 4.35 & 6.19 & 19.33\\
        \textcolor{black}{LoFTR-TR~\cite{LoFTR, planeTR}} & 34.61 & 35.30 & 43.55 & 25.52 & 27.79 & 37.32 & 5.91 & 8.22 & 20.62  \\\midrule
        SparsePlanes~\cite{jin2021planar} w/o Con. & 33.20& 34.12& 40.74& 22.89& 25.62& 33.67& 3.03& 4.52& 17.17 \\ 
        SparsePlanes~\cite{jin2021planar}  & 33.08& 34.12& 40.51& 21.69& 25.59& 32.20& 2.52&  4.50& 14.85\\
        PlaneFormers~\cite{planeformer} & 34.64& 35.47& 41.37& 24.48& 27.19& 34.69& 3.93& 5.52& 18.58\\\midrule
        SparsePlanes-TR~\cite{jin2021planar,planeTR} w/o Con. & 35.56& 36.51& 42.14& 26.01& 29.61& 35.12& 3.96& 6.10& 18.59\\  
        SparsePlanes-TR~\cite{jin2021planar,planeTR}  & 35.32& 36.50& 41.92& 24.71& 29.55& 33.50& 3.21& 6.07& 15.32\\
        PlaneFormers-TR~\cite{planeformer,planeTR} & 36.82& 37.87& 43.01& 27.41& 30.72& 36.31& 4.83& 7.02& 19.94\\\midrule
        NOPE-SAC-Cls Init. (ours)& 35.41 & 36.68 & 42.44 & 25.21 & 28.78 & 34.96 & 3.84 & 5.74 & 18.66\\
        NOPE-SAC-Cls Ref. (ours)& 38.23 & 39.36 & 43.27 & 30.25 & 34.15 & 36.93 & 6.23 & 9.57 & 20.56  \\\midrule
        NOPE-SAC-Reg Init. (ours)& 36.39 & 37.35 & 43.15 & 25.59 & 28.54 & 36.06 & 4.59 & 6.41 & 19.92 \\
        NOPE-SAC-Reg Ref. (ours)& \textbf{39.39} & \textbf{40.30} & \textbf{43.88} & \textbf{31.21} & \textbf{34.89} & \textbf{37.88} & \textbf{6.74} & \textbf{10.10} & \textbf{21.41} \\
        \bottomrule
    \end{tabular}
}
\label{tab:3DPlane}
\end{table*}

\subsubsection{Comparison of Camera Pose Estimation}
\label{subsec:exp_pose_compare}

\paragraph{Quantitative Results.} We first compare the performance of rotation estimation. As shown in Tab.~\ref{tab:cam_pose}, both our NOPE-SAC-Reg and NOPE-SAC-Cls significantly outperform all baselines on both datasets, particularly when the rotation threshold is small, \emph{e.g.,} $84.0\% $ (NOPE-SAC-Reg) v.s. $63.3\%$ (SparsePlanes-TR) with the threshold of $10^{\circ}$ on the Matterport3D~\cite{mp3d} dataset. Then, we evaluate the translation results. When compared to SparsePlanes-TR~\cite{jin2021planar,planeTR} and PlaneFormers-TR~\cite{planeformer,planeTR}, which use the a-priori classification poses, our NOPE-SAC-Cls performs slightly worse on the Matterport3D~\cite{mp3d} dataset but outperforms them on the more challenging ScanNet~\cite{dai2017scannet} dataset. 

In a further comparison with SparsePlanes-TR~\cite{jin2021planar,planeTR} as shown in Tab.~\ref{tab:cam_pose_detail}, we find that the translation improvement of SparsePlanes-TR largely depends on its keypoint-based continuous optimization. 
While SparsePlanes-TR performs well on the MatterPort3D~\cite{mp3d} dataset, its continuous optimization performance significantly deteriorates on the ScanNet~\cite{dai2017scannet} dataset. This is mainly due to the difficulty of achieving accurate keypoint matches on the ScanNet dataset, as illustrated in Fig.~\ref{fig:teaser_a}. In contrast, our NOPE-SAC-Cls consistently improves the initial poses using only the plane correspondences on both datasets.

\paragraph{Qualitative Results.} Fig.~\ref{fig:poses} shows the pose estimation results of our NOPE-SAC-Reg from two different viewpoints on the Matterport3D~\cite{mp3d} and the ScanNet~\cite{dai2017scannet} datasets. As described in Sec.~\ref{sec:NopeSAC-pose}, we compute one-plane pose hypotheses (\textbf{Gray} frustums) from estimated plane correspondences (the third column). Despite the outlier poses caused by incorrect correspondences and inaccurate plane parameters, our NOPE-SAC effectively achieved the final refined pose (\textbf{Blue} frustum) from all one-plane pose hypotheses (\textbf{Gray} frustums) and the initial pose (\textbf{Pink} frustum).

\begin{figure*}[t!]
\centering
\centering
    \subfigure[Image 1]{
    \begin{minipage}[b]{0.135\linewidth}
        \includegraphics[height=55pt]{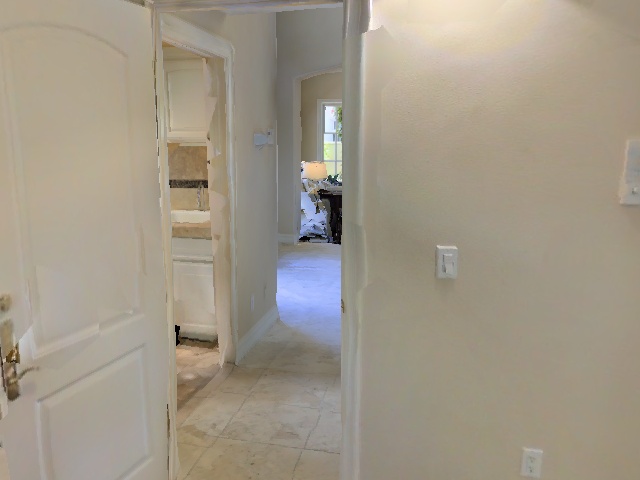}\vspace{2pt}
        \includegraphics[height=55pt]{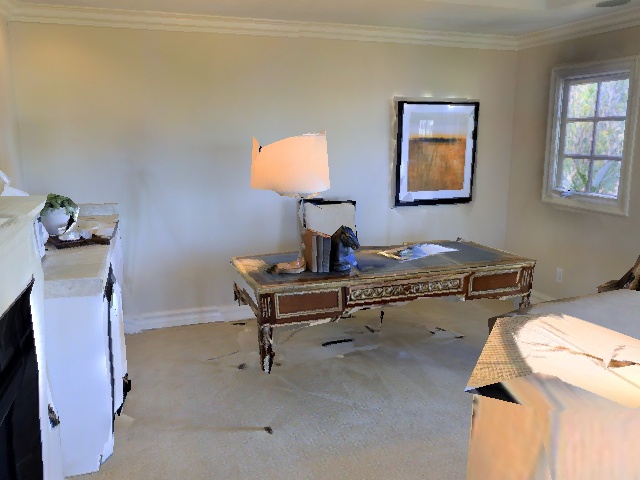}\vspace{2pt}
        \includegraphics[height=55pt]{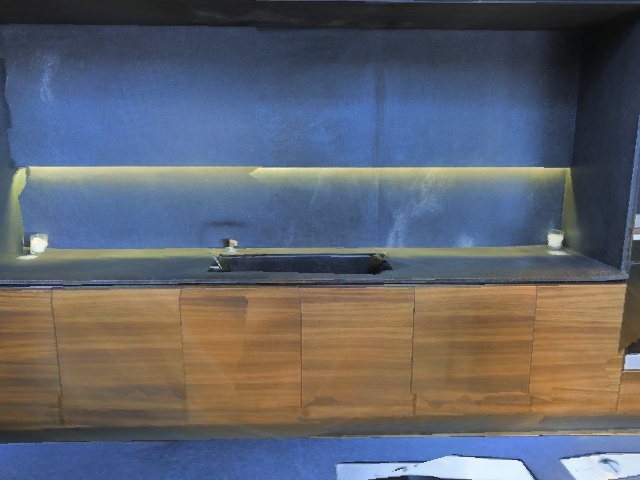}\vspace{2pt}
        \includegraphics[height=55pt]{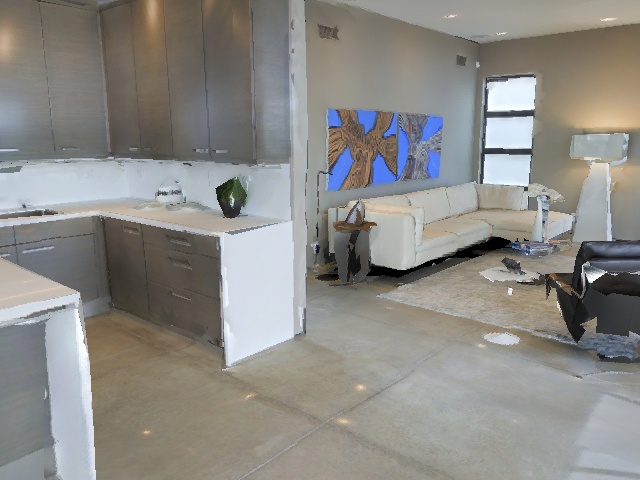}\vspace{2pt}
        \includegraphics[height=55pt]{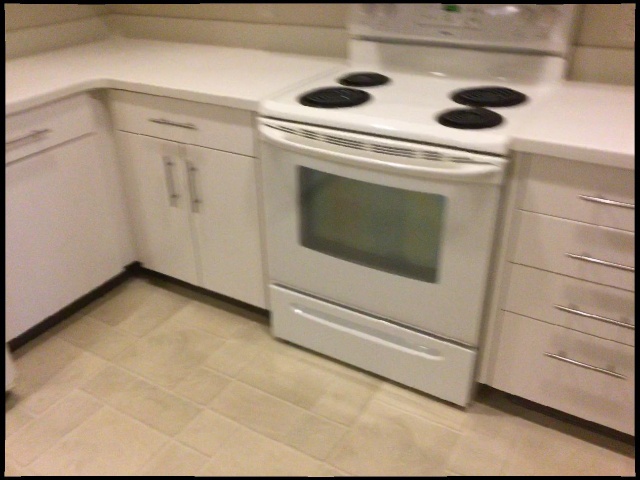}\vspace{2pt}
        \includegraphics[height=55pt]{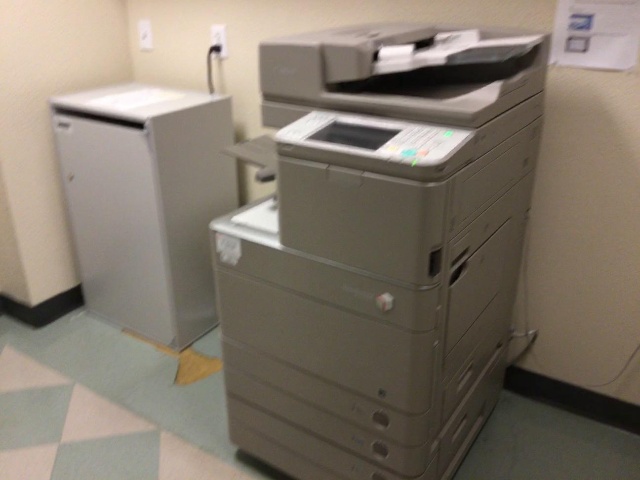}\vspace{2pt}
        \includegraphics[height=55pt]{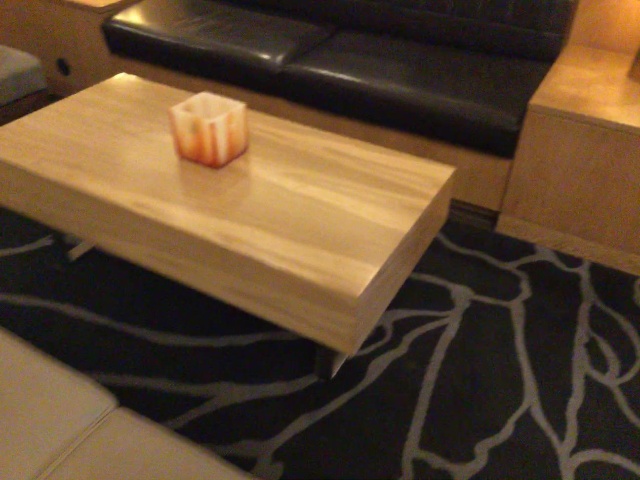}\vspace{2pt} 
    \end{minipage}}
    \subfigure[Image 2]{
    \begin{minipage}[b]{0.135\linewidth}
        \includegraphics[height=55pt]{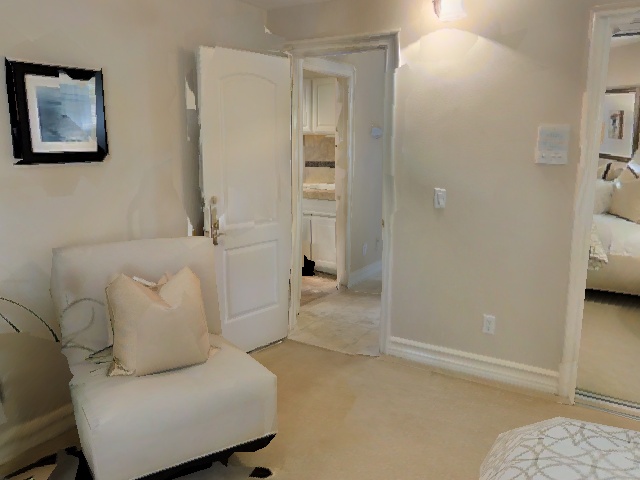}\vspace{2pt}
        \includegraphics[height=55pt]{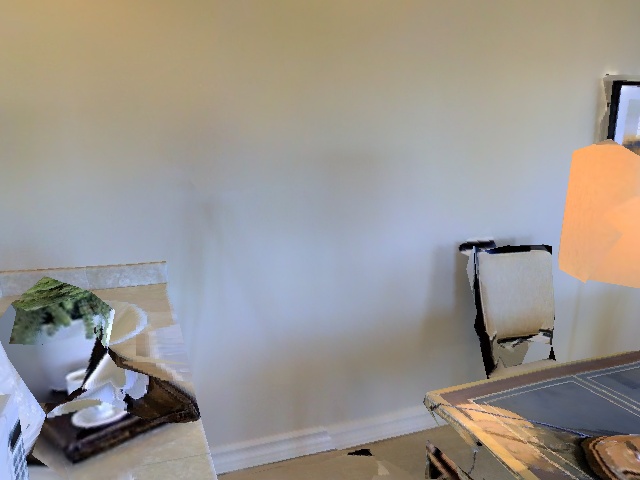}\vspace{2pt}
        \includegraphics[height=55pt]{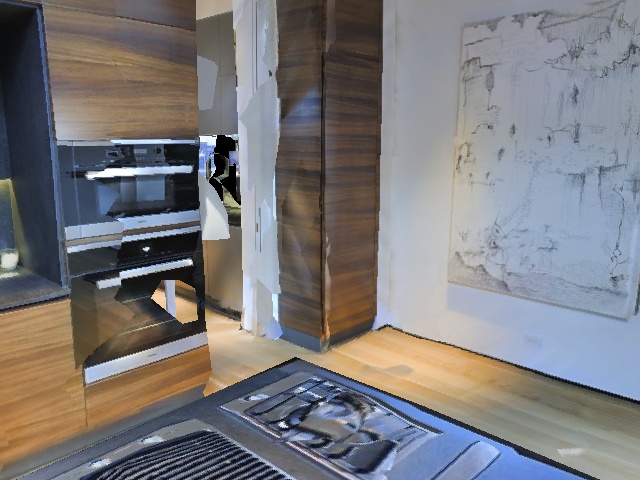}\vspace{2pt}
        \includegraphics[height=55pt]{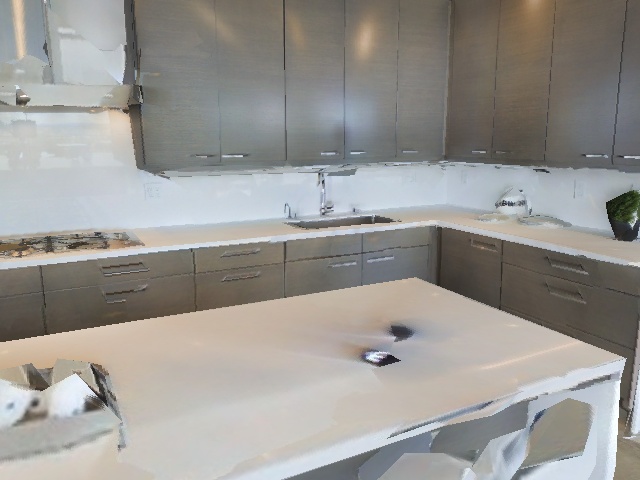}\vspace{2pt}
        \includegraphics[height=55pt]{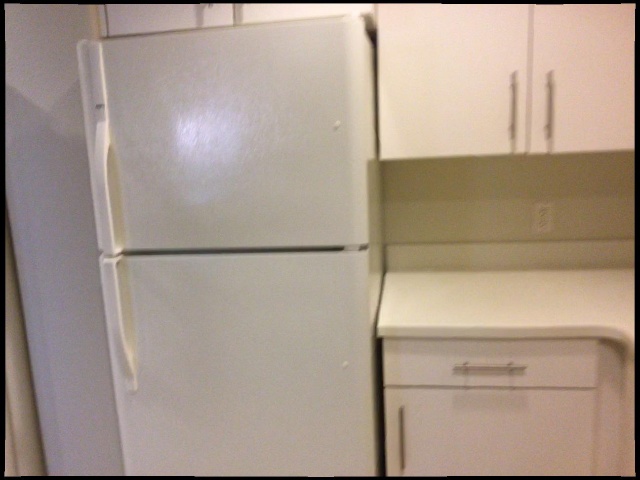}\vspace{2pt}
        \includegraphics[height=55pt]{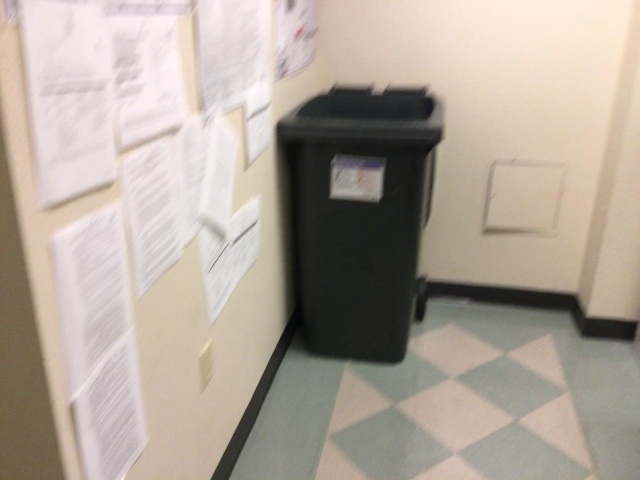}\vspace{2pt}
        \includegraphics[height=55pt]{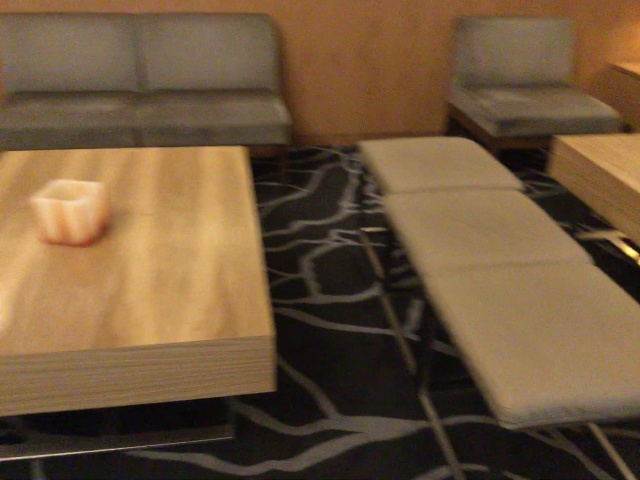}\vspace{2pt}
    \end{minipage}}
    \subfigure[SparsePlanes-TR]{
    \begin{minipage}[b]{0.165\linewidth}
        \includegraphics[height=55pt]{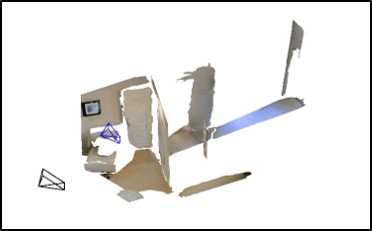}\vspace{2pt}
        \includegraphics[height=55pt]{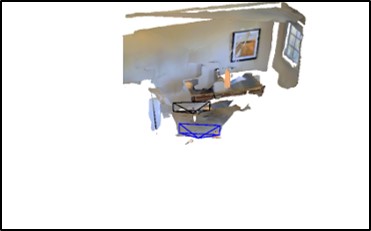}\vspace{2pt}
        \includegraphics[height=55pt]{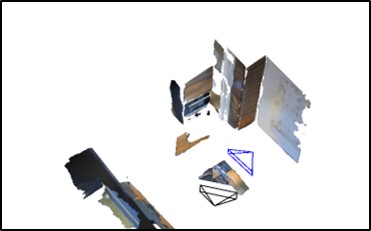}\vspace{2pt}
        \includegraphics[height=55pt]{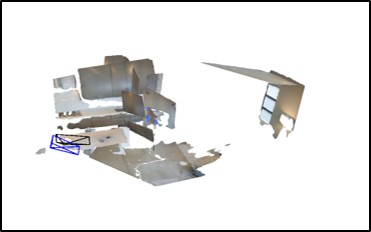}\vspace{2pt}
        \includegraphics[height=55pt]{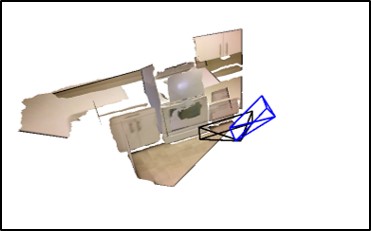}\vspace{2pt}
        \includegraphics[height=55pt]{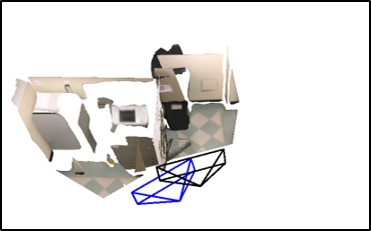}\vspace{2pt}
        \includegraphics[height=55pt]{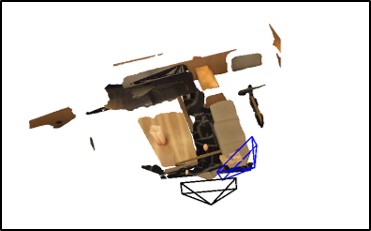}\vspace{2pt}
    \end{minipage}}
    \subfigure[PlaneFormers-TR]{
    \begin{minipage}[b]{0.165\linewidth}
        \includegraphics[height=55pt]{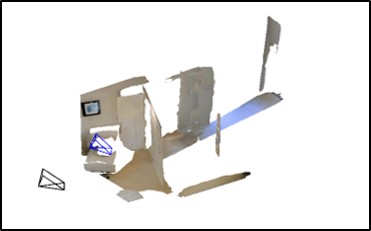}\vspace{2pt}
        \includegraphics[height=55pt]{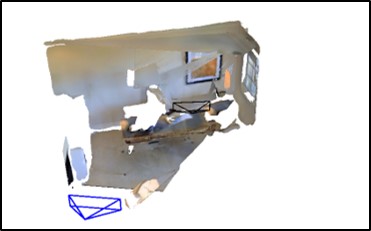}\vspace{2pt}
        \includegraphics[height=55pt]{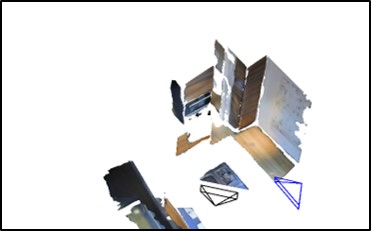}\vspace{2pt}
        \includegraphics[height=55pt]{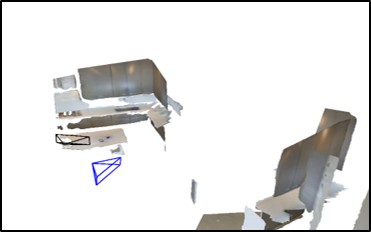}\vspace{2pt}
        \includegraphics[height=55pt]{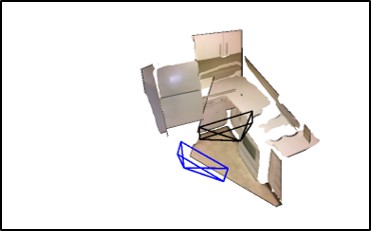}\vspace{2pt}
        \includegraphics[height=55pt]{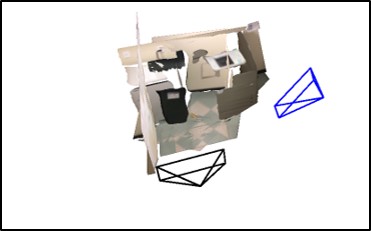}\vspace{2pt}
        \includegraphics[height=55pt]{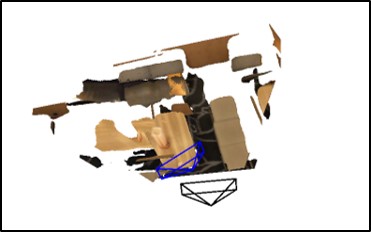}\vspace{2pt}
    \end{minipage}}
    \subfigure[Ours]{
    \begin{minipage}[b]{0.165\linewidth}
        \includegraphics[height=55pt]{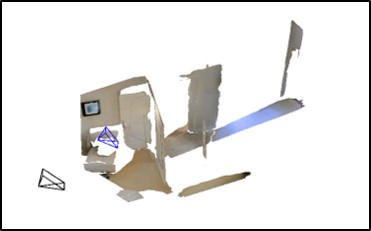}\vspace{2pt}
        \includegraphics[height=55pt]{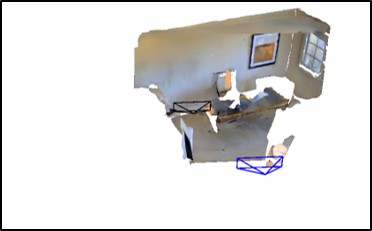}\vspace{2pt}
        \includegraphics[height=55pt]{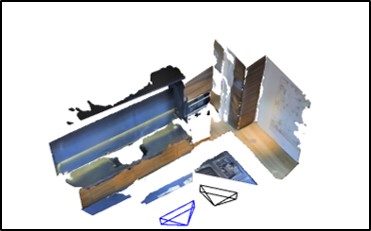}\vspace{2pt}
        \includegraphics[height=55pt]{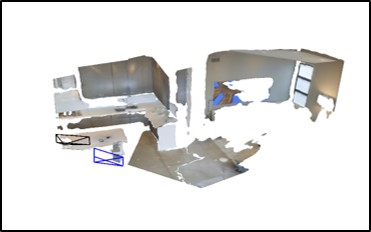}\vspace{2pt}
        \includegraphics[height=55pt]{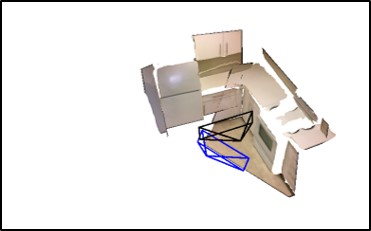}\vspace{2pt}
        \includegraphics[height=55pt]{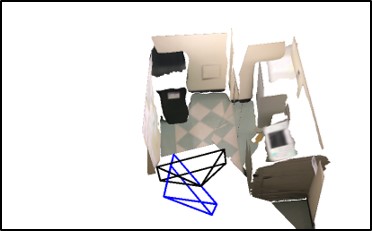}\vspace{2pt}
        \includegraphics[height=55pt]{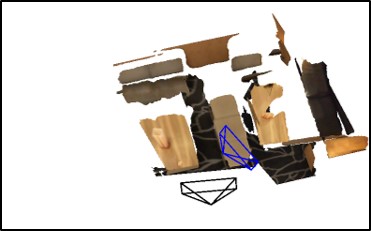}\vspace{2pt}
    \end{minipage}}
    \subfigure[Ground Truth]{
    \begin{minipage}[b]{0.165\linewidth}
        \includegraphics[height=55pt]{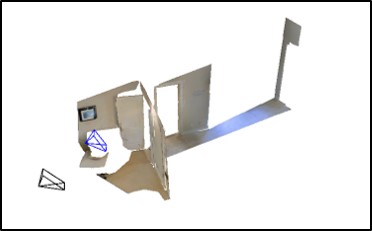}\vspace{2pt}
        \includegraphics[height=55pt]{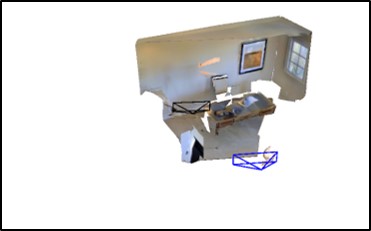}\vspace{2pt}
        \includegraphics[height=55pt]{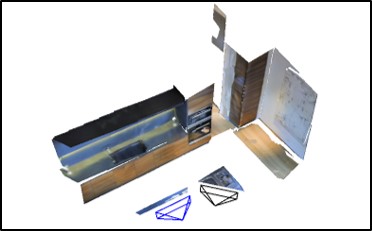}\vspace{2pt}
        \includegraphics[height=55pt]{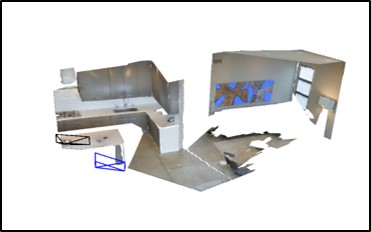}\vspace{2pt}
        \includegraphics[height=55pt]{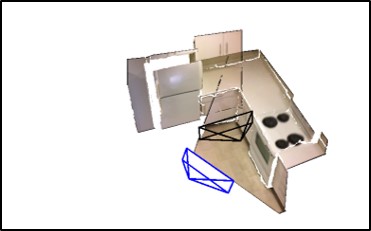}\vspace{2pt}
        \includegraphics[height=55pt]{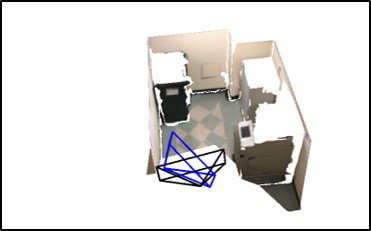}\vspace{2pt}
        \includegraphics[height=55pt]{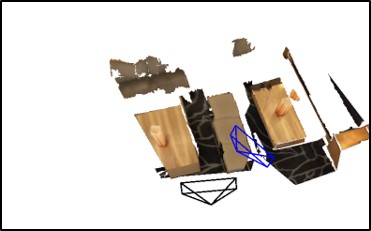}\vspace{2pt}
    \end{minipage}}
\vspace{-2mm}
\caption{Comparison of 3D plane reconstruction results on the Matterport3D~\cite{mp3d} dataset (first four rows) and the ScanNet~\cite{dai2017scannet} dataset (last three rows). \textbf{Blue} and \textbf{Black} frustums show cameras of the first and the second images respectively.}
\label{fig:plane_compare}
\vspace{-1mm}
\end{figure*}

\subsubsection{Comparison of 3D Planar Reconstruction}

\paragraph{Quantitative Results.} We evaluate our method with initial (Init.) and refined (Ref.) poses, and compare with baselines under various plane offset and normal error thresholds from loose to strict. As shown in Tab.~\ref{tab:3DPlane}, Our NOPE-SAC-Reg achieves state-of-the-art performance especially under the strictest settings with `Offset$\le$0.2m' and `Normal$\le 5^{\circ}$' on both Matterport3D~\cite{mp3d} and ScanNet\cite{dai2017scannet} datasets. When compared with SparsePlanes-TR~\cite{jin2021planar,planeTR} which also uses classification poses, our NOPE-SAC-Cls effectively improves the reconstruction performance (\emph{e.g.,} from 25.21 to 30.25 with offset$\le$0.5m and normal$\le10^{\circ}$ on the ScanNet dataset). Those performance gains confirmed the superiority of our proposed NOPE-SAC.

\paragraph{Qualitative Results.} 
Fig.~\ref{fig:plane_compare} visualizes the 3D plane reconstruction results of different approaches on the Matterport3D~\cite{mp3d} and the ScanNet~\cite{dai2017scannet} datasets. As it is shown, our method successfully reconstructs the scenes from sparse views even when the image overlap is very small (\emph{e.g.,} the third row in Fig.~\ref{fig:plane_compare}) and the viewpoint change is very large (\emph{e.g.,} the fourth row in Fig.~\ref{fig:plane_compare}).

\subsubsection{Comparison of Model Size and Inference Time}
In Tab.~\ref{tab:model_size}, we compare the number of model parameters and inference time of our NOPE-SAC with SparsePlanes~\cite{jin2021planar} and PlaneFormers~\cite{planeformer}. Our NOPE-SAC shows its superiority in both model size and inference time.

\subsection{Ablation for Plane Matching}
In this section, we verify the core compositions of our plane matching module including the usage of AGNN and score matrix in Eqn.~\ref{eq:full_score}. As shown in Tab.~\ref{tab:matching_ab}, each component contributes to the improvement of matching, and the best matching results are achieved with the full setting.

\begin{table}
\centering
\caption{\textcolor{black}{Comparsion of model size and inference time.}}
\vspace{-2mm}
\resizebox{0.9 \linewidth}{!}{
    \begin{tabular}{l|c|ccc}
    \toprule
        \multicolumn{1}{c|}{\multirow{2}{*}{Methods}} & \multicolumn{1}{c|}{\multirow{2}{*}{$\#$Parameters (M)}} & \multicolumn{3}{c}{\multirow{1}{*}{Inference Time (s/image)}} \\ 
         & & Detection & Matching + Pose & Total\\
        \midrule
        SparsePlanes~\cite{jin2021planar} & 161.2 & 0.11 & 6.30 & 6.41 \\
        PlaneFormers~\cite{planeformer} & 230.2 & 0.11 & 0.33 & 0.44\\
        NOPE-SAC (Ours) & \textbf{75.3} & 0.13 & 0.08 & \textbf{0.21}\\
        \bottomrule
    \end{tabular}
}
\label{tab:model_size}
\end{table}

\begin{table}
\centering
\caption{\textcolor{black}{Ablation studies of the plane matching module.}}
\vspace{-2mm}
\resizebox{0.9 \linewidth}{!}{
    \begin{tabular}{ccc|ccc|ccc}
    \toprule
        \multicolumn{3}{c|}{Settings} & \multicolumn{3}{c|}{Matterport3D~\cite{mp3d}} & \multicolumn{3}{c}{ScanNet~\cite{dai2017scannet}} \\ 
        AGNN& $S_e$  &  $S_g$  & P~$\uparrow$ & R~$\uparrow$ & F~$\uparrow$& P~$\uparrow$ & R~$\uparrow$ & F~$\uparrow$  \\
        \midrule
        &             & $\checkmark$ & 28.6&  22.2&  25.0& 25.6 & 23.5 & 24.5\\
        $\checkmark$&$\checkmark$ &              & 37.1 & 50.0 & 42.6 & 30.4 & 43.7& 35.8\\
                    &$\checkmark$ & $\checkmark$ & 47.4&  50.3&  48.8 & 41.9&45.6 &43.7 \\
        $\checkmark$&$\checkmark$ & $\checkmark$ & \textbf{49.9} & \textbf{51.5} & \textbf{50.7} & \textbf{44.3} & \textbf{48.0} & \textbf{46.0}\\
        \bottomrule
    \end{tabular}
}
\label{tab:matching_ab}
\end{table}

\begin{table*}[t!]
\centering
\caption{Ablation studies for NOPE-SAC camera pose refinement.}%
\vspace{-2mm}
\resizebox{0.92\linewidth}{!}{ 
    \begin{tabular}{l|ccccc|ccccc}
    \toprule
        \multicolumn{1}{c|}{\multirow{2}{*}{Method}} & \multicolumn{5}{c|}{Translation} & \multicolumn{5}{c}{Rotation} \\ 
                                & Med.~$\downarrow$ & Mean~$\downarrow$ & ($\le$1m)~$\uparrow$ & ($\le$0.5m)~$\uparrow$& ($\le$0.2m)~$\uparrow$ & Med.~$\downarrow$ & Mean~$\downarrow$ & ($\le 30^{\circ}$)~$\uparrow$ & ($\le 15^{\circ}$)~$\uparrow$ & ($\le 10^{\circ}$)~$\uparrow$\\
        \midrule
        \multicolumn{10}{c}{Matterport3D~\cite{mp3d} dataset} \\\midrule
        Initial Pose    &  0.69 &  1.08 & 65.0\% & 37.0\% & 10.1\% &  11.16 & 21.49 & 81.3\% & 60.5\% & 46.5\% \\ \midrule
        Nume-Ref-I  & 0.52 & 1.06 & 69.2\% & \textbf{48.5\%} & \textbf{19.1\%} & 7.17& 21.96 & 79.8\% & 67.4\% & 58.3\% \\
        Nume-Ref-II & 0.68& 1.07& 65.5\%& 38.2\%& 10.7\%& 8.38& 19.97& 82.2\% & 65.7\% & 54.9\% \\
        Homo-Ref & 0.61 & 1.05 & 68.5\% & 42.9\% & 14.9\%& 5.03& 17.47& 85.1\% & 74.5\% & 66.3\% \\
        PlaneFormers-TR~\cite{planeformer,planeTR} & 0.66 & 1.06 & 66.8\% & 37.5\% & 9.8\% & 8.96 & 20.01 & 83.0\% & 65.9\% & 53.5\% \\
        NOPE-SAC (ours) & \textbf{0.52}  &  \textbf{0.94}  &  \textbf{73.2\%}  &   48.3\%  &   16.2\% & \textbf{2.77}  &  \textbf{14.37}  &  \textbf{89.0\%}   &  \textbf{86.9\%}  &  \textbf{84.0\%} \\
        \midrule
        
        \multicolumn{10}{c}{ScanNet~\cite{dai2017scannet} dataset} \\
        \midrule
        Initial Pose      & 0.48 & 0.72 & 77.7\% & 51.9\% & 16.5\% &  14.68 &  26.75 & 73.7\% & 51.0\% & 34.4\% \\ \midrule
        Nume-Ref-I          & 0.54 & 0.80 & 73.2\% & 47.2\% & 13.7\% & 16.57& 29.47& 68.7\% & 47.0\% & 32.3\% \\
        Nume-Ref-II   & 0.48& 0.72& 77.7\% & 51.8\% & 16.5\% & 14.64& 26.72& 73.7\% & 51.1\% & 34.6\% \\
        Homo-Ref           & 0.49 & 0.73 & 77.5\% & 51.3\% & 16.5\% & 14.03& 26.27& 74.2\% & 52.7\% & 36.8\% \\
        PlaneFormers-TR~\cite{planeformer,planeTR} & 0.48 & 0.72 & 78.3\% & 52.2\% & 14.2\% & 14.30 & 26.51 & 74.5\% & 52.2\% & 33.7\% \\
        NOPE-SAC (ours) & \textbf{0.41}  & \textbf{0.65}  & \textbf{82.0\%}  &  \textbf{59.1\%}  &  \textbf{21.2\%} & \textbf{8.27}  & \textbf{22.12}  &  \textbf{82.6\%}  &  \textbf{73.2\%}  &  \textbf{59.5\%} \\ 
        \bottomrule
    \end{tabular}
}
\label{tab:cam_pose_ab_traditional}
\end{table*}

\subsection{Ablation for NOPE-SAC Pose Estimation}
\label{subsec:experiment_pose}
This section presents a series of ablation studies for our NOPE-SAC pose estimation as it plays the most important role in the end task. 
By default, we use the initial pose achieved by our pose regression module.

\paragraph{NOPE-SAC VS. Other Pose Refinement Methods.} Here, we first compare our NOPE-SAC with three traditional methods and one learning-based method for pose refinement, including:
\begin{enumerate}[noitemsep,topsep=2pt,parsep=2pt,partopsep=0pt]
    \item[(1)] \textbf{Nume-Ref-I}: We use the numerical optimization like SparsePlanes~\cite{jin2021planar} but only optimizes the initial camera poses $R_{0}$, $t_{0}$ as follows:
    \begin{equation} \label{eq:nume_ref_I}
        \begin{split}
            \mathop{\min}\limits_{R, t} \sum_{\mathcal{P}\in \mathbb{P}} &L(d_{\text{par}}(\mathcal{P}, R, t)) + d_{\text{pix}}(\mathcal{P}, R, t) \\
           &+ d_{\text{cam}}(R,R_{0}),
        \end{split}
    \end{equation}
    where $\mathcal{P}$ is a pair of matched planes in the predicted plane correspondence set $\mathbb{P}$, $L(\cdot)$ is the Huber loss, $d_\text{par}$ calculates Euclidean distance between plane parameters, $d_\text{pix}$ is the reprojection error of matched SIFT~\cite{SIFT} points on plane regions, and $d_\text{cam}$ is a regularization term which restricts the geodesic distance of rotations. 
    \item[(2)] \textbf{Nume-Ref-II}: A variant of Nume-Ref-I which excludes $d_{\text{pix}}$ in Eq.~\ref{eq:nume_ref_I}.
    \item[(3)] \textbf{Homo-Ref}: We use homographies to estimate refined poses from predicted plane correspondences $\mathbb{P}$. Specifically, to each predicted plane correspondence $\mathcal{P} \in \mathbb{P}$, matched SIFT points are found from plane regions and used to calculate the homography matrix $H$. Then, a refined pose hypothesis can be decomposed from $H$. We use the scale of the initial translation and select the refined pose from all hypotheses, which minimizes $\sum_{\mathcal{P} \in \mathbb{P}} d_{\text{par}}(\mathcal{P}, R, t)$.
    \item[(4)] \textbf{PlaneFormers-TR~\cite{planeformer, planeTR}}: We estimate the refined pose by giving planes detected by PlaneTR~\cite{planeTR} and our regressive initial pose as inputs to the PlaneFormers~\cite{planeformer}.
\end{enumerate}

As shown in Tab.~\ref{tab:cam_pose_ab_traditional}, compared to the learning-based PlaneFormers-TR~\cite{planeformer, planeTR}, which directly estimates a residual from plane correspondences for pose refinement, our method performs more effectively on both two datasets with our one-plane pose hypotheses. Thanks to the well-matched keypoints, the traditional Nume-Ref-I achieves the closest results to our NOPE-SAC on the Matterport3D~\cite{mp3d} dataset. However, Nume-Ref-I performs even worse than the initial pose on the ScanNet~\cite{dai2017scannet} dataset due to the difficulty of finding sufficient good keypoint matches. Similarly, Homo-Ref also suffers from unsatisfied keypoint matches on the ScanNet dataset. In contrast, benefiting from directly learning pose refinement with plane parameters in embedding space, our NOPE-SAC avoids the problem of keypoints detection and matching and achieves state-of-the-art performance.

\begin{table}[t!]
\centering
\caption{Ablation studies of the Arbitrary Initialization Module (AIM) and plane warping in NOPE-SAC pose estimation.}
\vspace{-2mm}
\resizebox{0.9\linewidth}{!}{ 
    \begin{tabular}{cc|ccc|ccc}
    \toprule
        \multicolumn{2}{c|}{Settings} & \multicolumn{3}{c|}{Trans.} & \multicolumn{3}{c}{Rot.} \\ 
        AIM & Warp Plane & Med.~$\downarrow$ & Mean~$\downarrow$ & ($\le$0.5m)~$\uparrow$& Med.~$\downarrow$ & Mean~$\downarrow$ & ($\le 15^{\circ}$)~$\uparrow$ \\
        \midrule
        \multicolumn{8}{c}{Matterport3D~\cite{mp3d} dataset} \\
        \midrule
        $\checkmark$&              & 0.61 & 1.00 & 42.3\% & 3.19 & 14.10 & 85.7\% \\
                    & $\checkmark$ & 0.51 & 0.92 & 49.3\% & 2.97 & 14.34 & 86.8\% \\
        $\checkmark$& $\checkmark$ & 0.52 & 0.94 & 48.3\% & 2.77 & 14.37 & 86.9\% \\ 
        \midrule
        \multicolumn{8}{c}{ScanNet~\cite{dai2017scannet} dataset} \\
        \midrule
        $\checkmark$&              & 0.42 & 0.66 & 57.4\% & 8.17 & 22.05 & 73.2\% \\
                    & $\checkmark$ & 0.40 & 0.64 & 59.9\% & 8.21 & 21.95 & 73.0\% \\
        $\checkmark$& $\checkmark$ & 0.41 & 0.65 & 59.1\% & 8.27 & 22.12 & 73.2\% \\ 
        \bottomrule
    \end{tabular}
}
\label{tab:cam_AIM_warpPlane_ab}
\vspace{-2mm}
\end{table}

\begin{table}[t!]
\centering
\caption{Ablation studies of the Arbitrary Initialization Module (AIM) in NOPE-SAC pose estimation. * means that we use the translation scale from our regressive initial pose for these methods.}
\vspace{-2mm}
\resizebox{0.9\linewidth}{!}{ 
    \begin{tabular}{c|ccc|ccc}
    \toprule
        \multicolumn{1}{c|}{\multirow{2}{*}{Methods}} & \multicolumn{3}{c|}{Trans.} & \multicolumn{3}{c}{Rot.} \\ 
         & Med.~$\downarrow$ & Mean~$\downarrow$ & ($\le$0.5m)~$\uparrow$& Med.~$\downarrow$ & Mean~$\downarrow$ & ($\le 15^{\circ}$)~$\uparrow$ \\
        \midrule
        \multicolumn{7}{c}{Matterport3D~\cite{mp3d} dataset} \\
        \midrule
        SuperGlue*~\cite{SuperGlue}& 0.60& 1.15 & 44.0\% & 3.88 & 24.17 & 71.0\% \\
        NOPE-SAC + SuperGlue* & 0.51 & 1.07 & 49.3\% & 2.95 & 19.76 & 82.2\% \\ \midrule
        LoFTR*~\cite{LoFTR}& 0.71 & 1.43 & 40.3\% & 5.85 & 33.13 & 61.0\% \\
        NOPE-SAC + LoFTR* & 0.61 & 1.31 & 44.2\% & 3.42 & 27.03 & 75.1\% \\
        \midrule
        \multicolumn{7}{c}{ScanNet~\cite{dai2017scannet} dataset} \\
        \midrule
        SuperGlue* ~\cite{SuperGlue}&  0.44&  0.69& 55.1\% & 10.90 & 31.00 & 56.0\% \\
        NOPE-SAC + SuperGlue* & 0.43 & 0.68 & 57.2\% & 8.61 & 26.95 & 67.2\% \\ \midrule
        LoFTR*~\cite{LoFTR}& 0.38 & 0.66 & 59.5\% & 5.49 & 27.13 & 63.3\% \\
        NOPE-SAC + LoFTR* & 0.39 & 0.66 & 59.1\% & 7.62 & 25.37 & 70.0\% \\
        \bottomrule
    \end{tabular}
}
\label{tab:AIM_with_keypointPose}
\vspace{-2mm}
\end{table}

\paragraph{Arbitrary Initialization Module.} 
\textcolor{black}{
This part analyzes the necessity of our Arbitrary Initialization Module (AIM).
As introduced in Sec.~\ref{subsec:pose_initial} and Fig.~\ref{fig:pose_init}, the initial pose embeddings can be achieved from either (1) the Regressive Initialization Module (RIM) or (2) the Arbitrary Initialization Module (AIM). 
Thus, we first fix the initial pose as the regressive prediction and compare the refined poses with initial pose embeddings from the above two methods. 
In Tab.~\ref{tab:cam_AIM_warpPlane_ab}, using initial pose embeddings from AIM achieves similar results to embeddings from the regression module, demonstrating our AIM's effectiveness. 
Furthermore, using AIM, our NOPE-SAC is flexible to refine initial poses that do not come from the convolutional Regressive Initialization Module. One typical example is our NOPE-SAC-Cls that uses the Top-1 classification pose predicted by SparsePlanes~\cite{jin2021planar} for pose initialization. Besides, we further select SuperGlue~\cite{SuperGlue} and LoFTR~\cite{LoFTR} for pose initialization and refine them with the pose embeddings achieved from our AIM. Since SuperGlue~\cite{SuperGlue} and LoFTR~\cite{LoFTR} lack scale information in translations, we use the scale from our Regressive Initialization Module for evaluation. As shown in Tab.~\ref{tab:AIM_with_keypointPose}, our NOPE-SAC successfully improves the initial poses in most metrics. We also find that the translation of LoFTR~\cite{LoFTR} degenerates slightly after our NOPE-SAC refinement on the ScanNet~\cite{dai2017scannet} dataset which is mainly caused by the inaccurate plane parameters under a relatively accurate camera pose estimation.}

\paragraph{Plane Warping for Correspondence Embedding.} Here, we discuss the influence of warping plane parameters before calculating correspondence embeddings as described in  Eqn.~\ref{eq:corrs_enc}. As shown in Tab.~\ref{tab:cam_AIM_warpPlane_ab}, warping plane parameters effectively improves the translation results, especially on the Matterport3D~\cite{mp3d} dataset (\emph{e.g.,} from $42.3\%$ to $48.3\%$).

\paragraph{Pose Hypotheses Fusion.} We then discuss the fusion strategies of one-plane pose hypotheses for pose refinement in Sec.~\ref{subsec:pose_refine}, including (1)~\textit{Soft}, (2)~\textit{Avg}, (3)~\textit{Min-Cost}, and (4)~\textit{Max-Score}. As shown in Tab.~\ref{tab:cam_pose_ab}, all strategies, even selecting only one pose hypothesis (\textit{Min-Cost} and \textit{Max-Score}), can improve the initial pose, demonstrating the effectiveness of our NOPE-SAC. 
Specifically, when evaluating translations, \textit{Avg} performs closely to \textit{Soft}, while \textit{Min-Cost} and \textit{Max-Score} degenerate significantly on both two datasets. This indicates that it is necessary to leverage more than one hypothesis to get better translation refinement results. When evaluating rotations, both \textit{Min-Cost} and \textit{Max-Score} perform closely to or slightly better than \textit{Soft}, while \textit{Avg} degenerates significantly. It indicates that rotations can be refined more easily than translations with one rotation hypothesis. 
However, due to the influence of matching outliers and errors in plane parameter estimation, the strategies of \textit{Avg}, \textit{Min-Cost}, and \textit{Max-Score} are not stable for camera pose estimation. In contrast, \textit{Soft} achieves the best overall performance by fusing all pose hypotheses with learned pose scores.

\begin{table}[t!]
\centering
\caption{Ablation studies of NOPE-SAC pose estimation with different one-plane pose fusion strategies}
\vspace{-2mm}
\resizebox{0.9\linewidth}{!}{ 
    \begin{tabular}{c|ccc|ccc}
    \toprule
        \multicolumn{1}{c|}{\multirow{2}{*}{Method}} & \multicolumn{3}{c|}{Trans.} & \multicolumn{3}{c}{Rot.} \\ 
                                & Med.~$\downarrow$ & Mean~$\downarrow$ & ($\le$0.5m)~$\uparrow$& Med.~$\downarrow$ & Mean~$\downarrow$ & ($\le 15^{\circ}$)~$\uparrow$ \\
        \midrule
        \multicolumn{7}{c}{Matterport3D~\cite{mp3d} dataset} \\
        \midrule
        Initial Pose  & 0.69   & 1.08   & 37.0\%    & 11.16    & 21.49    & 60.5\%  \\ \midrule
        Avg           & 0.56 & 0.96  & 45.3\%       & 5.03     & 17.21    & 77.1\%  \\
        Min-Cost         & 0.56 & 1.00  & 45.4\%  & 2.85    & 14.47    & 86.8\%  \\
        Max-Score      & 0.59 & 1.03 & 43.1\%  & 2.84    & 14.38    & \textbf{87.0\%}  \\
        Soft          & \textbf{0.52}    & \textbf{0.94}    & \textbf{48.3\%}     & \textbf{2.77} & \textbf{14.37} &86.9\%  \\ 
        \midrule
        \multicolumn{7}{c}{ScanNet~\cite{dai2017scannet} dataset} \\
        \midrule
        Initial Pose & 0.48  & 0.72 & 51.9\%  & 14.68  & 26.75 & 51.0\%  \\ \midrule
        Avg          & 0.42 & 0.65 & 57.8\%  & 10.28  & 23.33 & 63.2\%  \\
        Min-Cost       & 0.46  & 0.73  & 53.6\%  & 8.75  & 22.68 & 72.1\%  \\
        Max-Score     & 0.46   & 0.72 & 53.0\%  & 8.77  & 22.91  & 72.4\%  \\
        Soft         & \textbf{0.41}    & \textbf{0.65}    & \textbf{59.1\%}     & \textbf{8.27}     & \textbf{22.12}    & \textbf{73.2\%}  \\ 
        \bottomrule
    \end{tabular}
}
\label{tab:cam_pose_ab}
\end{table}

\begin{table}[!t]
\centering
\caption{Influence of plane parameter accuracy on NOPE-SAC pose estimation.}
\vspace{-2mm}
\resizebox{0.9\linewidth}{!}{
    \begin{tabular}{c|ccc|ccc}
    \toprule
        \multicolumn{1}{c|}{\multirow{2}{*}{Setting}} & \multicolumn{3}{c|}{Trans.} & \multicolumn{3}{c}{Rot.} \\ 
                                & Med.~$\downarrow$ & Mean~$\downarrow$ & ($\le$0.5m)~$\uparrow$& Med.~$\downarrow$ & Mean~$\downarrow$ & ($\le 15^{\circ}$)~$\uparrow$\\ \midrule
        \multicolumn{7}{c}{Matterport3D~\cite{mp3d} dataset} \\ \midrule
        Initial Pose         & 0.69& 1.08& 37.0\%& 11.16& 21.49& 60.5\% \\ \midrule
        w/ GT                & 0.32& 0.70& 63.5\%&  0.34&  3.85& 96.2\% \\ \midrule
        (0.1m, $5^{\circ}$)  & 0.39& 0.76& 58.6\%&  3.27&  6.85& 95.4\% \\ 
        (0.2m, $10^{\circ}$) & 0.48& 0.83& 51.5\%&  6.75& 10.76& 83.6\% \\ 
        (0.3m, $15^{\circ}$) & 0.56& 0.89& 45.3\%& 10.40& 15.17& 66.0\% \\ 
        \midrule
        \multicolumn{7}{c}{ScanNet~\cite{dai2017scannet} dataset} \\ \midrule
        Initial Pose         & 0.48& 0.72& 51.9\%& 14.68& 26.75& 51.0\% \\ \midrule
        w/ GT                & 0.26& 0.47& 72.1\%&  3.48& 8.42 & 90.7\% \\ \midrule
        (0.1m, $5^{\circ}$)  & 0.29& 0.49& 70.4\%& 7.22& 11.63& 88.2\% \\ 
        (0.2m, $10^{\circ}$) & 0.35& 0.55& 65.4\%& 11.95& 16.35& 65.4\% \\ 
        (0.3m, $15^{\circ}$) & 0.40& 0.60& 59.9\%& 16.35& 21.11& 44.3\% \\ 
        \bottomrule
    \end{tabular}
}
\label{tab:plane_noise_ab}
\end{table}

\begin{table}[t!]
\centering
\caption{Influence of the plane matching precision (P) on NOPE-SAC pose estimation.} 
\vspace{-2mm}
\resizebox{0.9\linewidth}{!}{ 
    \begin{tabular}{c|c|ccc|ccc}
    \toprule
        \multirow{2}{*}{Threshold} & \multirow{2}{*}{P} &\multicolumn{3}{c|}{Translation} & \multicolumn{3}{c}{Rotation} \\ 
                                & & Med.~$\downarrow$ & Mean~$\downarrow$ & ($\le$0.5m)~$\uparrow$& Med.~$\downarrow$ & Mean~$\downarrow$ & ($\le 15^{\circ}$)~$\uparrow$ \\\midrule
        \multicolumn{7}{c}{Matterport3D~\cite{mp3d} dataset} \\ \midrule
        0.2  & 49.9 & 0.52 & 0.94 & 48.3\% & 2.77 & 14.37 & 86.9\% \\
        0.1  & 48.8 & 0.52 & 0.94 & 48.4\% & 2.75 & 14.40 & 86.9\% \\
        0.01 & 46.4 & 0.53 & 0.94 & 47.9\% & 2.76 & 14.63 & 86.7\% \\
        0.001& 44.7 & 0.53 & 0.94 & 47.7\% & 2.77 & 14.96 & 86.3\% \\ \midrule
        \multicolumn{7}{c}{ScanNet~\cite{dai2017scannet} dataset} \\ \midrule
        0.2  & 44.3 & 0.41 & 0.66 & 59.1\% & 8.27 & 22.12 & 73.2\% \\
        0.1  & 43.0 & 0.41 & 0.65 & 58.9\% & 8.26 & 22.11 & 73.5\% \\
        0.01 & 40.8 & 0.41 & 0.65 & 58.6\% & 8.21 & 22.36 & 73.2\% \\
        0.001& 39.8 & 0.41 & 0.65 & 58.5\% & 8.22 & 22.56 & 73.2\% \\
        \bottomrule
    \end{tabular}
}
\label{tab:outlier_ab}
\end{table}

\begin{table*}[t!]
\centering
\caption{Comparison of camera poses on the SUN3D~\cite{sun3d} dataset.}
\vspace{-2mm}
\resizebox{0.97\linewidth}{!}{ 
    \begin{tabular}{l|ccc|ccc|ccc|ccc|ccc|ccc}
    \toprule
        \multicolumn{1}{c|}{\multirow{3}{*}{Method}} & \multicolumn{6}{c|}{Overlap Ratio $\le$ 1.0 (\#758)} & \multicolumn{6}{c|}{Overlap Ratio $\le$ 0.3 (\#613)} & \multicolumn{6}{c}{Overlap Ratio $\le$ 0.1 (\#190)} \\ 
        & \multicolumn{3}{c|}{Translation} & \multicolumn{3}{c|}{Rotation}& \multicolumn{3}{c|}{Translation}& \multicolumn{3}{c|}{Rotation}& \multicolumn{3}{c|}{Translation}& \multicolumn{3}{c}{Rotation} \\
        &Med.~$\downarrow$ & Mean~$\downarrow$ & ($\le$0.5m)~$\uparrow$& Med.~$\downarrow$ &Mean~$\downarrow$& ($\le 15^{\circ}$)~$\uparrow$& Med.~$\downarrow$ &Mean~$\downarrow$& ($\le$0.5m)~$\uparrow$& Med.~$\downarrow$ &Mean~$\downarrow$& ($\le 15^{\circ}$)~$\uparrow$& Med.~$\downarrow$ &Mean~$\downarrow$& ($\le$0.5m)~$\uparrow$& Med.~$\downarrow$ &Mean~$\downarrow$& ($\le 15^{\circ}$)~$\uparrow$ \\                     
        \midrule
        SuperGlue~\cite{SuperGlue}  & - & - & - & 4.53 & \textbf{15.43} & 76.7\% & - & - &- & 5.61 & 18.49& 71.1\% & - & -&- & 31.63 & 41.29 & 32.6\% \\
        LoFTR~\cite{LoFTR}  & - & - & - & \textbf{3.98} & 17.69 & 74.4\% & - & - &- & \textbf{5.39} & 21.32& 68.7\% & - & -&- & 27.37 & 43.75 & 34.7\% \\\midrule
        SparsePlanes~\cite{jin2021planar}  & 0.74 & 0.93 & 27.6\% & 15.07 & 26.27 & 49.7\% & 0.85 & 1.03 & 20.6\% & 16.52 & 29.66 & 43.7\% & 1.09 & 1.24 & 14.7\%& 26.04 & 43.61 & 26.3\% \\
        PlaneFormers~\cite{planeformer}  & 0.69 & 0.89 & 33.9\% & 13.85 & 26.35 & 54.9\% & 0.78& 0.98&28.4\% &14.78 & 29.69& 51.2\% & 0.94& 1.15& 22.1\%& 21.13& 43.07& 37.4\% \\
        SparsePlanes-TR~\cite{jin2021planar,planeTR} & 0.71 & 0.89 & 30.1\% & 14.36 & 25.03 & 52.6\% & 0.81 & 0.98 & 23.8\% & 15.67 & 28.15 & 47.3\% & 0.99 & 1.14 & 14.2\% & 22.99 & 42.01 & 33.7\% \\
        PlaneFormers-TR~\cite{planeformer,planeTR} & 0.66 & 0.86 & 33.6\% & 13.42& 25.66 & 57.9\% & 0.74 & 0.95& 28.2\% & 13.88 & 28.70 & 55.5\% & 1.00 & 1.13 & 21.6\% & 20.69 & 41.19 & 37.4\% \\
        \midrule
        NOPE-SAC-Cls (ours)  & 0.63 &0.85& 39.5\%& 8.46 &21.43&  76.8\%&  0.73&0.94& 32.5\%& 9.29&24.84& 72.1\%&0.99&1.16&20.5\%& 13.85&39.17& 53.7\% \\
        NOPE-SAC-Reg (ours)  & \textbf{0.58} &\textbf{0.76}& \textbf{42.1\%}& 7.54 &15.70& \textbf{81.9\%}& \textbf{0.67} & \textbf{0.85} & \textbf{35.1\%}& 8.02& \textbf{17.74}& \textbf{78.8\%}&\textbf{0.93}&\textbf{1.11}&\textbf{24.7\%}& \textbf{12.47}&\textbf{30.69}& \textbf{57.9\%} \\
        \bottomrule
    \end{tabular}
}
\label{tab:cam_pose_sun3d_full}
\end{table*}
\paragraph{Influence of Plane Parameters.}
In this part, we study the influence of plane parameter accuracy on our NOPE-SAC pose estimation on the Matterport3D~\cite{mp3d} and the ScanNet~\cite{dai2017scannet} datasets. The upper bound of our method is achieved by using ground truth plane correspondences and ground truth plane parameters in our pose refinement, defined as `w/ GT' in Tab.~\ref{tab:plane_noise_ab}. Then, we add various Gaussian noises to the offset and the normal of ground truth plane parameters in both two image views. We set the mean of the Gaussian noises to zero. The standard deviation of the plane offset increases from 0.1m to 0.3m, and the standard deviation of the plane normal increases from $5^{\circ}$ to $15^{\circ}$. As shown in Tab.~\ref{tab:plane_noise_ab}, our method effectively improves the initial pose even in the challenging setting of (0.2m, $10^{\circ}$), but fails to improve rotations in the setting of (0.3m, $15^{\circ}$) on the ScanNet dataset because of too large noises on plane parameters. It demonstrates that our method is robust to the accuracy of plane parameters.

\paragraph{Influence of Matching Precision.}
We also evaluate the influence of plane-matching precision on our NOPE-SAC. We conducted the experiments by gradually reducing the threshold of plane matching from 0.2 to 0.001. 
Table~\ref{tab:outlier_ab} illustrates the results of these experiments. As the matching threshold decreases, the plane matching precision reduces from 49.9 to 44.7, while the pose metrics only change slightly on the Matterport3D dataset. Similar results can be found on the ScanNet dataset. These results demonstrate the robustness of our NOPE-SAC in handling incorrect plane correspondences.

\paragraph{Generalization Ability.}
In the last, we evaluate the generalization ability of pose estimation, which is the core of our NOPE-SAC. We randomly sample image pairs from 8 scenes\footnote{mit-76-studyroom-76-1studyroom2, harvard-c5-hv-c5-1, harvard-c6-hv-c6-1, harvard-c8-hv-c8-3, hotel-umd-maryland-hotel3, mit-32-d507-d507-2, mit-dorm-next-sj-dorm-next-sj-oct-30-2012-scan1-erika, mit-lab-hj-lab-hj-tea-nov-2-2012-scan1-erika} in the indoor SUN3D~\cite{sun3d} dataset with ground truth camera poses. We fix the frame interval within a sampled image pair to be 100 and filter samples with the image overlap ratio lower than $0.5\%$, resulting in 758 image pairs in the final. 
We use the official models of SuperGlue~\cite{SuperGlue} and LoFTR~\cite{LoFTR}. To SparsePlanes~\cite{jin2021planar}, PlaneFormes~\cite{planeformer}, and our NOPE-SAC, we use models finetuned on the ScanNet~\cite{dai2017scannet} dataset. We compare our method with these baselines under various overlap ratio thresholds. As summarized in Tab.~\ref{tab:cam_pose_sun3d_full}, our NOPE-SAC-Reg achieves state-of-the-art performance, especially under the low overlap ratio threshold, demonstrating the excellent generalization ability of our method.

\section{Conclusion} 
This paper studies the challenging two-view 3D reconstruction in a rigorous sparse-view configuration. 
At the core of estimating the camera poses from one-plane pose hypotheses generated from plane correspondences with neural networks, our proposed NOPE-SAC formulates the problem in a consensus sampling paradigm while enjoying end-to-end learning without incurring any offline optimization schemes. In the experiments, our NOPE-SAC achieves new state-of-the-art performances for sparse-view camera pose estimation and planar 3D reconstruction in indoor scenes on the challenging Matterport3D~\cite{mp3d} and ScanNet~\cite{dai2017scannet} datasets. Furthermore, our proposed method can be generalized to unseen SUN3D~\cite{sun3d} dataset without any parameter tuning with state-of-the-art performance obtained.

\section*{Acknowledgements}
We thank the anonymous reviewers and associate editors for their constructive comments. We also thank Xianpeng Liu for proofreading. This work was supported by the NSFC Grants under contracts No. 62101390, 62325111, and U22B2011. 
T. Wu was partly supported by ARO Grant W911NF1810295, NSF IIS-1909644, ARO Grant W911NF2210010, NSF IIS-1822477, NSF CMMI-2024688 and NSF IUSE-2013451. 
The views presented in this paper are those of the authors and should not be interpreted as representing any funding agencies.

\ifCLASSOPTIONcaptionsoff
  \newpage
\fi

\bibliographystyle{IEEEtran}
\bibliography{IEEEabrv,egbib}
\begin{IEEEbiography}[{\includegraphics[width=1in,height=1.25in,clip,keepaspectratio]{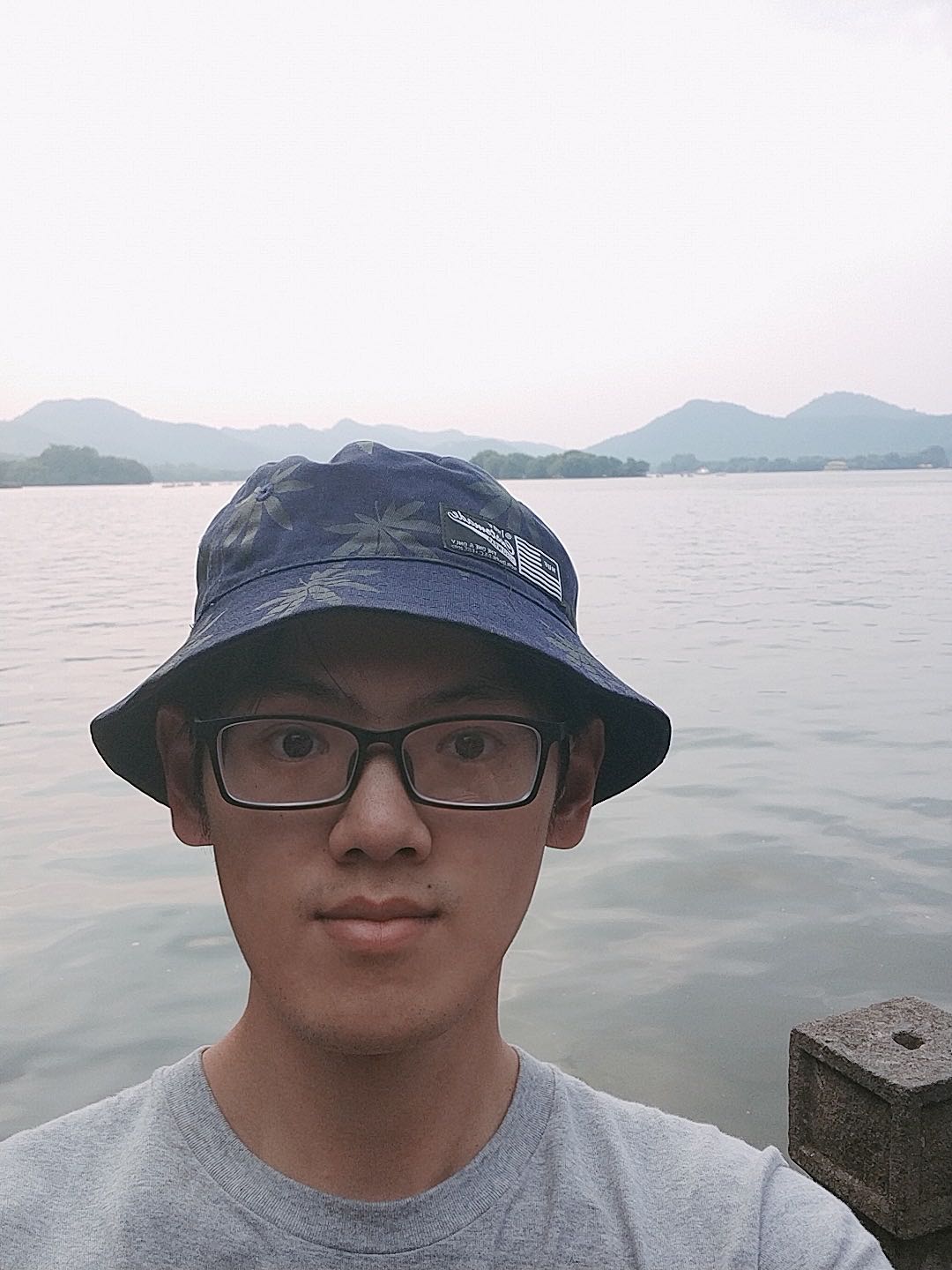}}]{Bin Tan} received the B.S. and M.S. degrees both from the School of Remote Sensing and Information Engineering, Wuhan University. He is currently working toward a Ph.D. degree in the School of Computer Science, Wuhan University, Wuhan, China. His research interest is around the perception, modeling and reconstruction of geometric structures in 3D computer vision.
\end{IEEEbiography}
 
\begin{IEEEbiography}[{\includegraphics[width=1in,height=1.25in,clip,keepaspectratio]{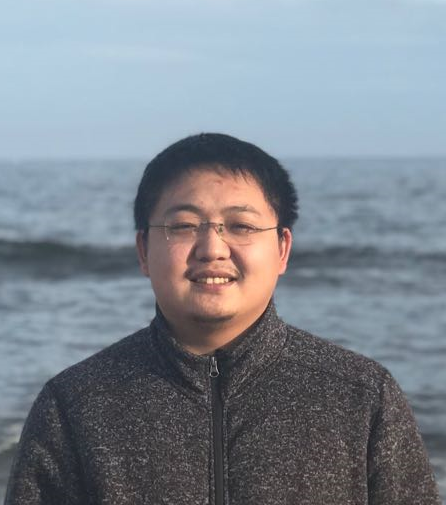}}]{Nan Xue} is currently a research scientist at Ant Research, affiliated with Ant Group. Previously, from 2020 to 2023, he served as a research associate professor in the School of Computer Science at Wuhan University. He completed his Ph.D. at the LIESMARS (State Key Laboratory of Information Engineering in Surveying, Mapping, and Remote Sensing), Wuhan University, in 2020. From 2018 to 2020, he worked as a visiting Ph.D. student with the iVMCL Group at NC State University. Prior to pursuing his Ph.D., he obtained a B.S. degree from the School of Mathematics and Statistics at Wuhan University in 2014. He is a recepient of the Outstanding Doctoral Dissertation Award, by China Society of Images and Graphics in 2022. 
His current research focus primarily revolves around the development of learning algorithms for the computation of structured visual geometry in the realm of computer vision, in particular to 3D vision problems. Please visit \url{https://xuenan.net} for the most recently updates.
\end{IEEEbiography}
 
\begin{IEEEbiography}[{\includegraphics[width=1in,height=1.25in,clip,keepaspectratio]{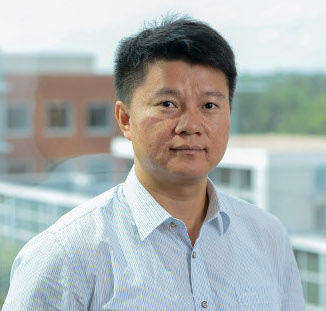}}]{Tianfu Wu}
    is an associate professor in the Department of Electrical and Computer Engineering at NC state university (NCSU), and leads the laboratory for interpretable visual modeling, computing and learning (iVMCL). He received his Ph.D. in statistics from UCLA in 2011.	His research focuses on computer vision, often motivated by the task of building explainable and improvable visual Turing test and robot autonomy through life-long communicative learning. To accomplish his research goals, he is interested in pursuing a unified framework for machines to ALTER (Ask, Learn, Test, Explain and Refine) recursively in a principled way.
\end{IEEEbiography}

\begin{IEEEbiography}[{\includegraphics[width=1in,height=1.25in,clip,keepaspectratio]{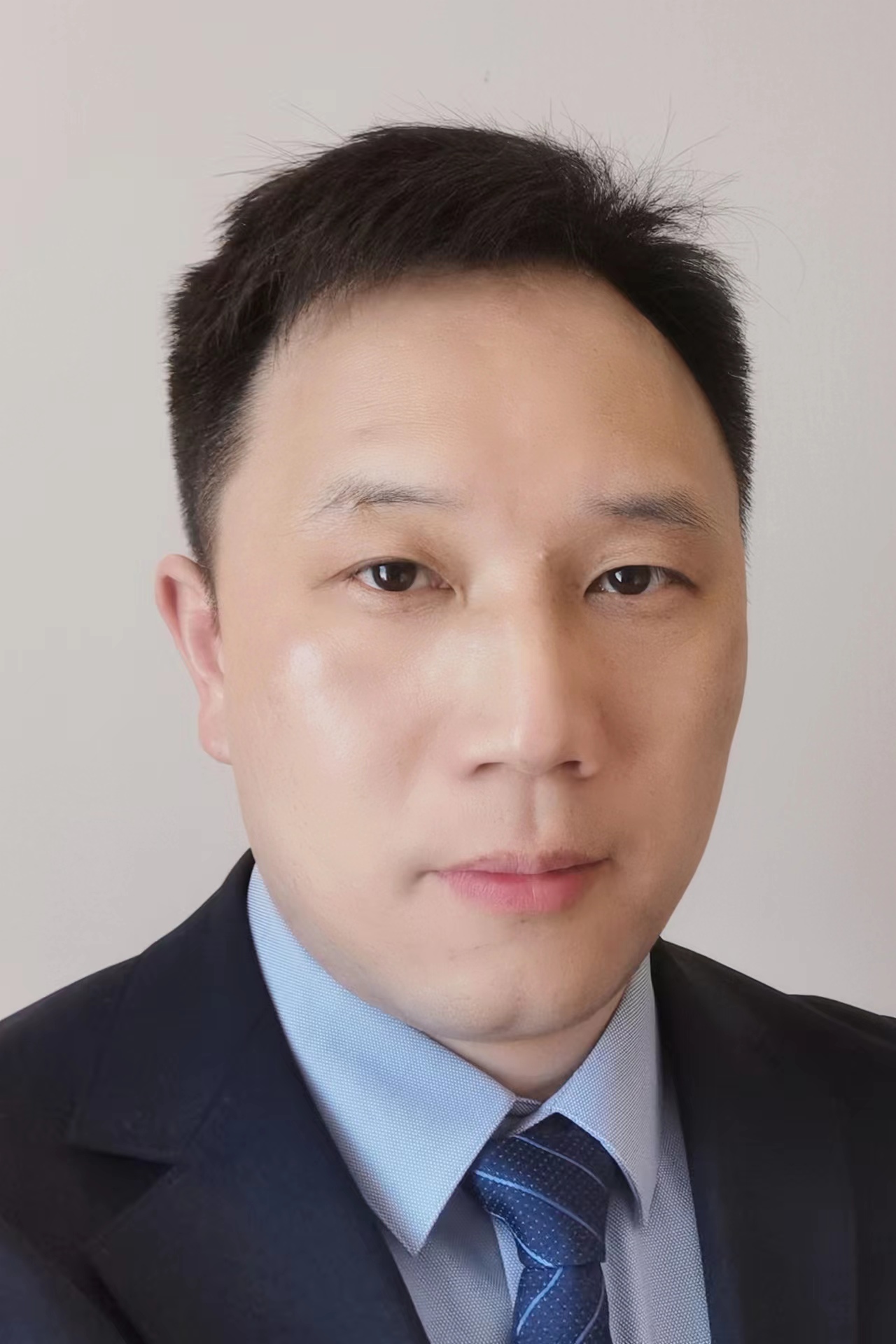}}]{Gui-Song Xia}
(M'10-SM'15) received his Ph.D. degree in image processing and computer vision from CNRS LTCI, T{\'e}l{\'e}com Paris, Paris, France, in 2011. From 2011 to 2012, he has been a Post-Doctoral Researcher with the Centre de Recherche en Math{\'e}matiques de la Decision, CNRS, Paris-Dauphine University, Paris, for one and a half years.
He is currently working as a full professor in computer vision and photogrammetry at Wuhan University. He has also been working as Visiting Scholar at DMA, {\'E}cole Normale Sup{\'e}rieure (ENS-Paris) for two months in 2018. His current research interests include computer vision, robotics and remote sensing imaging. 
\end{IEEEbiography}

\end{document}